\documentclass[10pt,twocolumn,letterpaper]{article}

\usepackage{cvpr}              

\usepackage{graphicx}
\usepackage{amsmath}
\usepackage{amssymb}
\usepackage{booktabs}

\usepackage{makecell}
\usepackage{multirow}
\usepackage{booktabs}
\usepackage{xcolor}
\usepackage{pifont}
\newcommand{\cmark}{\ding{51}}
\newcommand{\xmark}{\ding{55}}
\usepackage{setspace}
\usepackage{animate}

\usepackage[pagebackref,breaklinks,colorlinks]{hyperref}

\usepackage[capitalize]{cleveref}
\crefname{section}{Sec.}{Secs.}
\Crefname{section}{Section}{Sections}
\Crefname{table}{Table}{Tables}
\crefname{table}{Tab.}{Tabs.}

\def\cvprPaperID{3885} 
\def\confName{CVPR}
\def\confYear{2022}

\begin{document}
\newcolumntype{L}[1]{>{\raggedright\arraybackslash}p{#1}}
\newcolumntype{C}[1]{>{\centering\arraybackslash}p{#1}}
\newcolumntype{R}[1]{>{\raggedleft\arraybackslash}p{#1}}

\title{IFRNet: Intermediate Feature Refine Network for Efficient Frame Interpolation}

\author{
	Lingtong Kong\textsuperscript{\rm 1}$^{*}$,
	Boyuan Jiang\textsuperscript{\rm 2}$^{*}$,
	Donghao Luo\textsuperscript{\rm 2},
	Wenqing Chu\textsuperscript{\rm 2},
	Xiaoming Huang\textsuperscript{\rm 2}, \\
	Ying Tai\textsuperscript{\rm 2},
	Chengjie Wang\textsuperscript{\rm 2},
	Jie Yang\textsuperscript{\rm 1}$^{\dagger}$ \\
	\textsuperscript{\rm 1}Shanghai Jiao Tong University, China, \,
	\textsuperscript{\rm 2}Youtu Lab, Tencent \\
	{\tt\small \{ltkong, jieyang\}@sjtu.edu.cn } \\
	{\tt\small \{byronjiang, michaelluo, wenqingchu, skyhuang, yingtai, jasoncjwang\}@tencent.com} \\
}


\maketitle

\let\thefootnote\relax\footnotetext{$*$ Equal contribution. This work was done when Lingtong Kong was an intern at Tencent Youtu Lab.}
\let\thefootnote\relax\footnotetext{$\dagger$ Corresponding author: Jie Yang (jieyang@sjtu.edu.cn). This research is partly supported by NSFC, China (No: 61876107, U1803261).}

\begin{abstract}
	Prevailing video frame interpolation algorithms, that generate the intermediate frames from consecutive inputs, typically rely on complex model architectures with heavy parameters or large delay, hindering them from diverse real-time applications. In this work, we devise an efficient encoder-decoder based network, termed IFRNet, for fast intermediate frame synthesizing. It first extracts pyramid features from given inputs, and then refines the bilateral intermediate flow fields together with a powerful intermediate feature until generating the desired output. The gradually refined intermediate feature can not only facilitate intermediate flow estimation, but also compensate for contextual details, making IFRNet do not need additional synthesis or refinement module. To fully release its potential, we further propose a novel task-oriented optical flow distillation loss to focus on learning the useful teacher knowledge towards frame synthesizing. Meanwhile, a new geometry consistency regularization term is imposed on the gradually refined intermediate features to keep better structure layout. Experiments on various benchmarks demonstrate the excellent performance and fast inference speed of proposed approaches. Code is available at \url{https://github.com/ltkong218/IFRNet}.
\end{abstract}

\section{Introduction}
Video frame interpolation (VFI), that converts low frame rate (LFR) image sequences to high frame rate (HFR) videos is an important low-level computer vision task. 
Related techniques are widely applied to various practical applications, such as slow-motion generation~\cite{8579036}, novel view synthesis~\cite{Zhou_2016} and cartoon creation~\cite{Siyao_2021_CVPR}. 
Although it has been studied by a large number of researches, there are still great challenges when dealing with complicated dynamic scenes, which include large displacement, severe occlusion, motion blur and abrupt brightness change.

\begin{figure}[t]
	\centering
	\includegraphics[width=0.97\columnwidth]{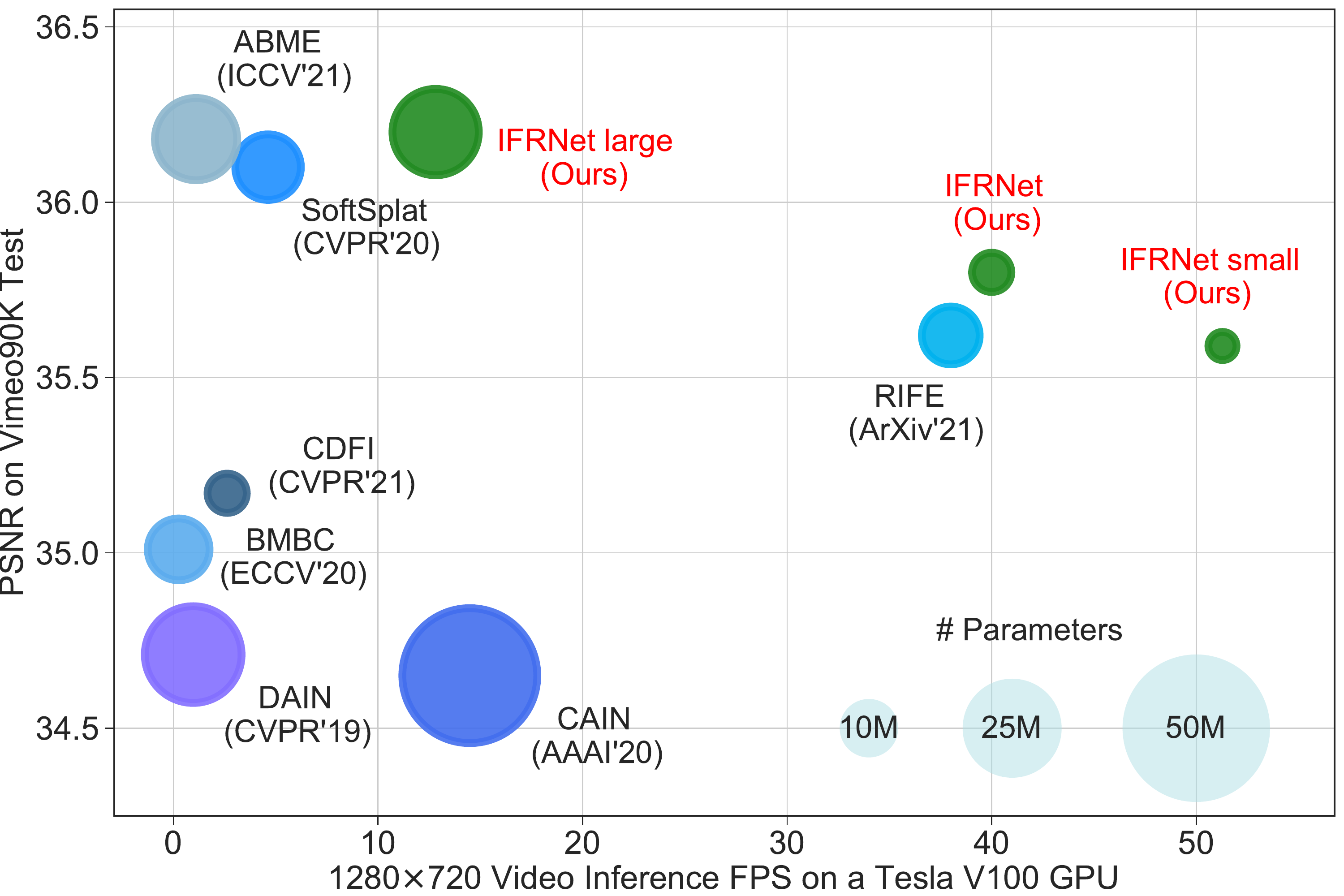}
	\vspace{-3mm}
	\caption{\textbf{Speed, accuracy and parameters comparison.} Proposed IFRNet achieves state-of-the-art frame interpolation accuracy with fast inference speed and lightweight model size.}
	\label{fig:1}
	\vspace{-4mm}
\end{figure}

\begin{figure*}[t]
	\centering
	\includegraphics[width=0.9\textwidth]{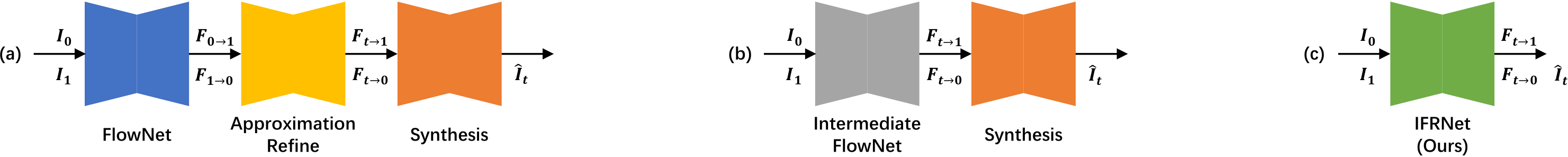}
	\vspace{-3mm}
	\caption{\textbf{Different flow-based VFI paradigms.} We roughly classify existing flow-based VFI methods based on encoder-decoders with specific function. In (a)~\cite{8954114,8578281,Niklaus_2020_CVPR,8579036,qvi_nips19,BMBC,park2021asymmetric,Sim_2021_ICCV}, FlowNet estimates conventional optical flow $F_{0\rightarrow1}, F_{1\rightarrow0}$, the middle part approximates or further refines flow fields $F_{t\rightarrow0}, F_{t\rightarrow1}$. In (b)~\cite{xue2019video,Zhang_2020,huang2021rife}, the Intermediate FlowNet directly predicts intermediate flow of $F_{t\rightarrow0}, F_{t\rightarrow1}$. Both (a) and (b) contain a separate synthesis network for target frame generation. In (c), proposed IFRNet jointly refines the intermediate flow $F_{t\rightarrow0}, F_{t\rightarrow1}$ together with a powerful intermediate feature $\hat{\phi}_{t}$ to generate the target frame in a single encoder-decoder.}
	\label{fig:2}
	\vspace{-4mm}
\end{figure*}

Recently, with the development of optical flow networks~\cite{7410673,8579029,Teed_2020,9560800}, significant progress has been made by flow-based VFI approaches~\cite{8579036,qvi_nips19,Niklaus_2020_CVPR,park2021asymmetric}, since optical flow can provide an explicit correspondence to register frames in a video sequence. 
Successful flow-based approaches usually follow a three-step pipeline: 
\textbf{1)} Estimate optical flow between target frame and input frames. 
\textbf{2)} Warp input frames or context features by predicted flow fields for spatial alignment.
\textbf{3)} Refine warped frames or features and generate the target frame by a synthesis network. 
Denoting input frames and target frame to be $I_0, I_1$ and $I_t \, (0<t<1)$, existing methods either first estimate optical flow $F_{0\rightarrow1}, F_{1\rightarrow0}$~\cite{8579036,8954114,8578281,Niklaus_2020_CVPR,BMBC}, and then approximate or refine bilateral intermediate flow $F_{t\rightarrow0}, F_{t\rightarrow1}$~\cite{8579036,qvi_nips19,chiall,Sim_2021_ICCV} as shown in Figure~\ref{fig:2} (a), or throw the intractable intermediate flow estimation sub-task to a learnable flow network for end-to-end training~\cite{xue2019video,Zhang_2020,huang2021rife} as depicted in Figure~\ref{fig:2} (b). Their common step is to further employ an image synthesis network to encode spatial aligned context feature~\cite{8578281} for target frame generation or refinement.

Although above pipeline that first estimates intermediate flow and then context feature has become the most popular paradigm for flow-based VFI approaches~\cite{8578281,Niklaus_2020_CVPR,chiall,park2021asymmetric,Sim_2021_ICCV}, it suffers from several defects. First, they divide intermediate flow and context feature refinement into separate encoder-decoders, which ignores the mutual promotion of these two crucial elements for frame interpolation. Second, their cascaded architecture based on above design concept can substantially increase the inference delay and model parameters, blocking them from mobile and real-time applications.

In this paper, we propose a novel Intermediate Feature Refine Network (IFRNet) for VFI to overcome the above limitations. For the first time, we merge above separated flow estimation and feature refinement into \textit{a single encoder-decoder based model for compactness and fast inference}, abstracted in Figure~\ref{fig:2} (c). It first extracts pyramid features from given inputs by the encoder, and then jointly refines the bilateral intermediate flow fields together with a powerful intermediate feature through coarse-to-fine decoders. The improved architecture can benefit intermediate flow and intermediate feature with each other, endowing our model with the ability to not only generate sharper moving objects but also capture better texture details.

For better supervision, we propose task-oriented flow distillation loss and feature space geometry consistency loss to effectively guide the multi-scale motion estimation and intermediate feature refinement. Specifically, our flow distillation approach adjusts the robustness of distillation loss adaptively in space and focuses on learning the useful teacher knowledge for frame synthesizing. Besides, proposed geometry consistency loss can employ the extracted intermediate features from ground truth to constrain the reconstructed intermediate features for keeping better structure layout. Figure~\ref{fig:1} gives a speed, accuracy and parameters comparison among advanced VFI methods, demonstrating the state-of-the-art performance of our approaches. In summary, our main contributions are listed as follows:
\vspace{-2mm}
\begin{itemize}
	\item We devise a novel IFRNet to jointly perform intermediate flow estimation and intermediate feature refinement for efficient video frame interpolation.
	\vspace{-2mm}
	\item Task-oriented flow distillation loss and feature space geometry consistency loss are newly proposed to promote intermediate motion estimation and intermediate feature reconstruction of IFRNet, respectively.
	\vspace{-2mm}
	\item Benchmark results demonstrate that our IFRNet not only achieves state-of-the-art VFI accuracy, but also enjoys fast inference speed and lightweight model size.
	\vspace{-2mm}
\end{itemize}

\vspace{-2mm}
\section{Related Work}
\paragraph{Video Frame Interpolation.}
The mainstream VFI methods can be classified into flow-based~\cite{8954114,8578281,Niklaus_2020_CVPR,8579036,xue2019video,qvi_nips19,8237740,Yuan_2019_CVPR,BMBC,Zhang_2020,park2021asymmetric,Sim_2021_ICCV}, kernel-based~\cite{8099727,8237299,8953614,Lee_2020_CVPR,Cheng_2020,9501506,ding2021cdfi} and hallucination-based approaches~\cite{Gui_2020_CVPR,choi2020cain,Kim_2020}. Different VFI paradigms have their own merits and flaws due to the substantial frame synthesizing manner. For example, kernel-based methods are good at handling motion blur by convolving over local patches~\cite{8099727,8237299}, successive works mainly extend it to deal with high resolution videos~\cite{8953614}, increase the degrees of freedom for convolution kernel~\cite{Lee_2020_CVPR,Cheng_2020,9501506}, or combine them with other paradigms for compensation~\cite{8840983,ding2021cdfi}. However, they are typically computationally expensive and short of dealing with occlusion. In another way, hallucination-based methods directly synthesize frames from the feature domain by blending field-of-view features generated by deformable convolution~\cite{8237351} or PixelShuffle operations~\cite{choi2020cain}. They can naturally generate complex contextual details, while the predicted frames tend to be blurry when fast-moving objects exist.

\begin{figure*}[t]
	\centering
	\includegraphics[width=0.98\textwidth]{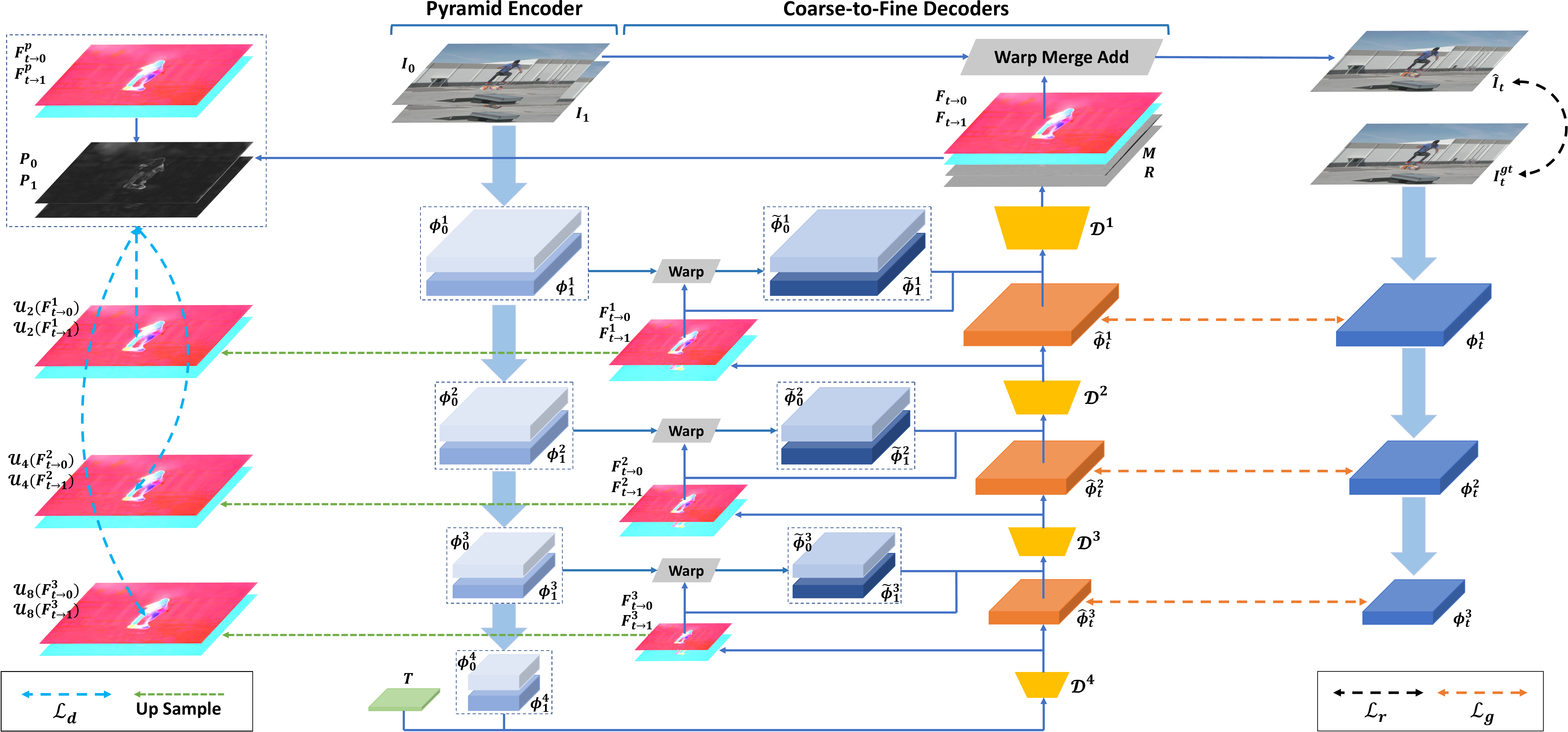}
	\vspace{-1mm}
	\caption{\textbf{Architecture overview and loss functions of IFRNet.} Our model is an efficient encoder-decoder based network, which first extracts pyramid context features from input frames with a shared encoder, and then gradually refines bilateral intermediate flow fields $F_{t\rightarrow0}, F_{t\rightarrow1}$ together with reconstructed intermediate feature $\hat{\phi}_{t}$ through coarse-to-fine decoders, until yielding the final output. Besides the common image reconstruction loss $\mathcal{L}_{r}$, task-oriented flow distillation loss $\mathcal{L}_{d}$ and feature space geometry consistency loss $\mathcal{L}_g$ are newly devised to guide the feature alignment procedure more effectively towards intermediate frame synthesizing.}
	\label{fig:3}
	\vspace{-1mm}
\end{figure*}

Recently, significant progress has been made by flow-based VFI approaches, since optical flow can provide an explicit correspondence for frame registration. These solutions either employ an off-the-shelf flow model~\cite{8578281,qvi_nips19} or estimate task-specific flow~\cite{xue2019video,8237740,8579036,park2021asymmetric,Sim_2021_ICCV} as a guidance for pixel-level motion. Common subsequent step is to forward~\cite{4056711} or backward~\cite{George_2000} warp input images to target frame, and finally refine warped frames by an image synthesis network~\cite{8578281,Niklaus_2020_CVPR,ding2021cdfi,park2021asymmetric}, often instantiated as a GridNet~\cite{fourure2017gridnet}. For achieving better image interpolation quality, more complicated deep models are devised to estimate intermediate flow fields~\cite{qvi_nips19,chiall} and refine the generated target frame~\cite{8579036,Niklaus_2020_CVPR,BMBC,park2021asymmetric}. However, the heavy computation cost and large inference delay make them unsuitable for resource limited devices. 
To take a breath from above module cascading competition, and reconsider the improvement of prior efficient flow-based VFI paradigm, \emph{e.g.} DVF~\cite{8237740}, we propose a novel single encoder-decoder based IFRNet, that can perform real-time inference with excellent accuracy.

\paragraph{Optical Flow Estimation.}
Finding dense correspondence between adjacent frames, namely optical flow estimation~\cite{HORN1981185}, has been studied for decades for its fundamental role in many downstream video processing tasks~\cite{Yuan_2020_CVPR,Chan_2021_CVPR}. FlowNet~\cite{7410673} is the first attempt to apply deep learning for optical flow estimation based on the encoder-decoder U-shape network. Inspired by traditional coarse-to-fine paradigm, SPyNet~\cite{Ranjan_2017_CVPR}, PWC-Net~\cite{8579029} and FastFlowNet~\cite{9560800} integrate pyramid feature, backward warping and achieve impressive real-time performance. Knowledge distillation~\cite{HinVin15Distilling} also plays an important role in optical flow prediction, usually embodied as generating pseudo label in unsupervised optical flow learning~\cite{DDFlow,SelFlow} or related tasks~\cite{NIPS2014_00ec53c4,aleotti2020learning}. A recent VFI method~\cite{huang2021rife} also uses a distillation strategy to promote motion prediction. Beyond the difference of architecture design, our distillation approach can focus on the useful knowledge for intermediate frame synthesizing in a task adaptative manner.

\section{Proposed Approach}
In this section, we first introduce the IFRNet architecture built on the principle of joint refinement of intermediate flow and intermediate feature, to obtain an efficient encoder-decoder based framework for VFI. Then two novel objective functions, \textit{i.e.}, task-oriented flow distillation loss and feature space geometry consistency loss are introduced to help our model achieve excellent performance.

\subsection{IFRNet}
Given two input frames $I_{0}$ and $I_{1}$ at adjacent time instances, video frame interpolation aims to synthesize an intermediate frame $I_{t}$, where $0<t<1$. To achieve this goal, proposed model performs a first extraction phase so as to retrieve a pyramid of features from each frame, then in a coarse-to-fine manner it progressively refines bilateral intermediate flow fields together with reconstructed intermediate feature until reaching the highest level of the pyramid to obtain the final output. Figure~\ref{fig:3} sketches the overall architecture of proposed IFRNet.

\noindent\textbf{Pyramid Encoder.}
To obtain contextual representation from each input frame, we design a compact encoder $\mathcal{E}$ to extract a pyramid of features. Purposely, the parameter shared encoder is built of a block of two 3$\times$3 convolutions in each pyramid level, respectively with strides 2 and 1. As shown in Figure~\ref{fig:3}, IFRNet extracts 4 levels of pyramid features, counting 8 convolution layers, each followed by a PReLU activation~\cite{7410480}. By gradually decimating the spatial size, it increases the feature channels to 32, 48, 72 and 96, generating pyramid features $\phi_{0}^{k}, \phi_{1}^{k}$ in level $k$ ($k\in\{1, 2, 3, 4\}$) for frames $I_0$ and $I_1$, respectively.

\noindent\textbf{Coarse-to-Fine Decoders.}
After extracting meaningful hierarchical representations, we then gradually refine intermediate flow fields through multiple decoders by backward warping pyramid features $\phi_{0}^{k}, \phi_{1}^{k}$ to generate $\tilde{\phi}_{0}^{k}, \tilde{\phi}_{1}^{k}$ according to $F_{t\rightarrow0}^{k}$ and $F_{t\rightarrow1}^{k}$, respectively. The main advantage of coarse-to-fine warping strategy consists of computing easier residual flow at each scale. Different from previous VFI approaches containing post-refinement~\cite{Niklaus_2020_CVPR,ding2021cdfi,park2021asymmetric,huang2021rife}, we explore to improve the bilateral flow prediction during its coarse-to-fine procedure for higher efficiency. Specifically, we make each decoder $\mathcal{D}^{k+1}$ output a higher level reconstructed intermediate feature $\hat{\phi}_{t}^{k}$ besides bilateral flow fields $F_{t\rightarrow0}^{k}, F_{t\rightarrow1}^{k}$, which can fill up the missing reference information to facilitate motion estimation. On the other hand, better predicted flow fields $F_{t\rightarrow0}^{k}, F_{t\rightarrow1}^{k}$ will align source pyramid features to the target position more precisely, thus, generating better $\tilde{\phi}_{0}^{k},\tilde{\phi}_{1}^{k}$, which can in turn improve higher level intermediate feature reconstruction. Therefore, decoders in proposed IFRNet can jointly refine bilateral intermediate flow fields together with reconstructed intermediate feature, benefitting each other until reaching desired output. Moreover, the gradually refined intermediate feature, containing bilateral occlusion and global context information, can finally generate fusion mask and compensate for motion details, that are often missing by flow-based methods, enabling IFRNet a powerful encoder-decoder VFI architecture without additional refinement~\cite{Niklaus_2020_CVPR,park2021asymmetric}.

\begin{figure}[t]
	\centering
	\includegraphics[width=0.8\columnwidth]{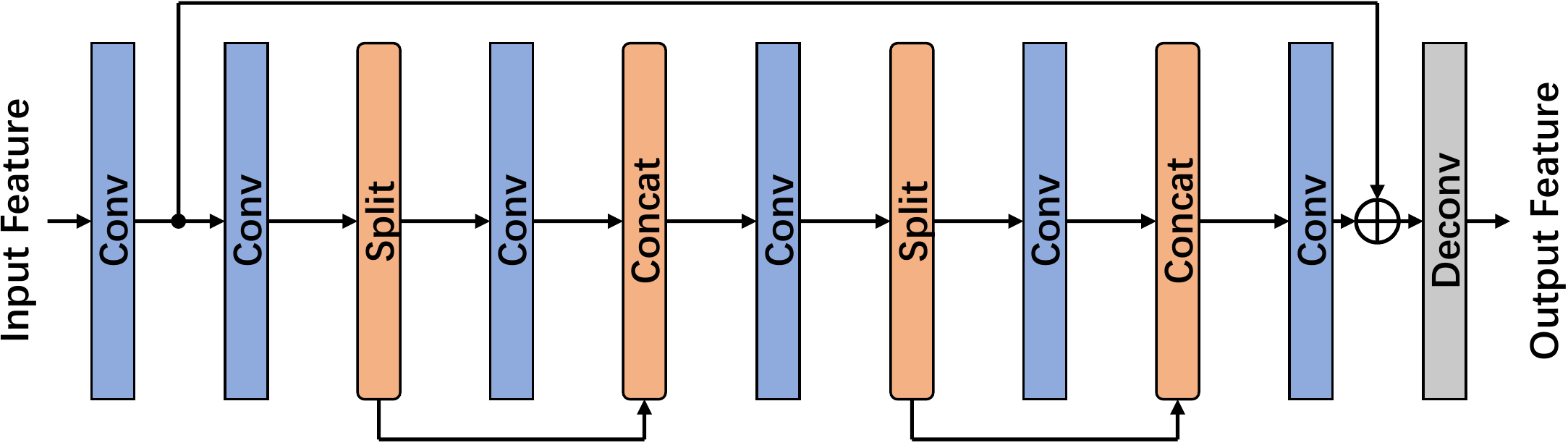}
	\vspace{-2mm}
	\caption{\textbf{Details of the decoder in each pyramid level.}}
	\label{fig:4}
	\vspace{-4mm}
\end{figure}

Concretely, in each pyramid level, we stack corresponding input features into a holistic volume that is forwarded by a compact decoder network $\mathcal{D}^{k}$, consisting of a block of six 3$\times$3 convolutions and one 4$\times$4 deconvolution, with strides 1 and $1/2$, respectively. A PReLU~\cite{7410480} follows each convolution layer. Details of each decoder is shown in Figure~\ref{fig:4}. In order to keep relative large receptive field and channel numbers for motion estimation and feature encoding while maintaining efficiency, we modify the third and the fifth convolution to update only partial channels of previous output tensor. Furthermore, residual connection and interlaced placement can promote information propagation and joint refinement. More details are shown in supplementary. Note that inputs of $\mathcal{D}^{4}$ and outputs of $\mathcal{D}^{1}$ are different from other decoders due to the task-related characteristics. In summary, features among decoders can be computed by
\begin{align}
	&[F_{t\rightarrow0}^{3}, F_{t\rightarrow1}^{3}, \hat{\phi}_{t}^{3}] = \mathcal{D}^{4}([\phi_{0}^{4}, \phi_{1}^{4}, T]), \\
	&[F_{t\rightarrow0}^{k-1}, F_{t\rightarrow1}^{k-1}, \hat{\phi}_{t}^{k-1}] = \mathcal{D}^{k}([F_{t\rightarrow0}^{k}, F_{t\rightarrow1}^{k}, \hat{\phi}_{t}^{k}, \tilde{\phi}_{0}^{k}, \tilde{\phi}_{1}^{k}]), \\
	&[F_{t\rightarrow0}, F_{t\rightarrow1}, M, R] = \mathcal{D}^{1}([F_{t\rightarrow0}^{1}, F_{t\rightarrow1}^{1}, \hat{\phi}_{t}^{1}, \tilde{\phi}_{0}^{1}, \tilde{\phi}_{1}^{1}]),
\end{align}
where $\mathcal{D}^{k} (k=2,3)$ stand for decoders at middle pyramid levels, $[\cdot]$ denotes concatenation operation. $T$ is a one-channel conditional input for arbitrary time interpolation, whose values are all the same and set to $t$. $M$ is a one-channel merge mask exported by a sigmoid layer whose elements range from 0 to 1, and $R$ is a three-channel image residual that can compensate for details. Finally, we can synthesize the desired frame $\hat{I}_{t}$ by following formulation
\begin{gather}
	\hat{I}_{t} = M \odot \tilde{I}_{0} + (1 - M) \odot \tilde{I}_{1} + R, \label{eq:4}\\
	\tilde{I}_{0} = w(I_{0}, F_{t\rightarrow0}), \, \tilde{I}_{1} = w(I_{1}, F_{t\rightarrow1}),
\end{gather}
where $w$ means backward warping, $\odot$ is element-wise multiplication. $M$ adjusts the mixing ratio according to bidirectional occlusion information, while $R$ compensates for some details when flow-based generation is unreliable, such as regions of target frame are occluded in both views.

\noindent\textbf{Discussion with Optical Flow Networks.}
Different from the coarse-to-fine pipeline in real-time optical flow~\cite{8579029,9560800} which mainly deals with large displacement matching challenge, in video interpolation, since the target frame is missing, its motion estimation becomes a ``chicken-and-egg" problem. Therefore, decoders of IFRNet reconstruct intermediate feature besides intermediate flow fields, performing spatio-temporal feature aggregation and intermediate motion refinement jointly to benefit from each other.

\noindent\textbf{Image Reconstruction Loss.}
According to above analysis, an efficient IFRNet has been designed for VFI, which is end-to-end trainable. For the purpose of generating intermediate frame, we employ the same image reconstruction loss $\mathcal{L}_{r}$ as \cite{park2021asymmetric} between network output $\hat{I}_{t}$ and ground truth frame $I_{t}^{gt}$, which is the sum of two terms and denoted by
\begin{equation}
	\mathcal{L}_{r} = \rho(\hat{I}_{t} - I_{t}^{gt}) + \mathcal{L}_{cen}(\hat{I}_{t}, I_{t}^{gt}),
	\label{eq:6}
\end{equation}
where $\rho(x) = (x^2 + \epsilon^2)^{\alpha}$ with $\alpha = 0.5, \epsilon = 10^{-3}$ is the Charbonnier loss~\cite{413553} severing as a surrogate for the $\mathcal{L}_{1}$ loss. While $\mathcal{L}_{cen}$ is the census loss, which calculates the soft Hamming distance between census-transformed~\cite{Meister_2018} image patches of size 7$\times$7.

\subsection{Task-Oriented Flow Distillation Loss}
Training IFRNet with above reconstruction loss $\mathcal{L}_{r}$ can already perform intermediate frame synthesizing. However, the simple optimization target usually drops into local minimum, since illuminance cases are often challenging, \textit{i.e.}, extreme brightness and repetitive texture regions. To deal with this problem, we try to adopt the knowledge distillation~\cite{HinVin15Distilling} strategy to guide multi-scale intermediate flow estimation of IFRNet by an off-the-shelf teacher flow network, that helps to align multi-scale pyramid features explicitly. In practice, the pre-trained teacher is only used during training, and we calculate its flow prediction as pseudo label $F_{t\rightarrow0}^{p}, F_{t\rightarrow1}^{p}$ in advance for efficiency. Note that RIFE~\cite{huang2021rife} also uses flow distillation. However, their indiscriminate distillation manner usually learns undesired noise existed in pseudo label. Even if ground truth is available, optical flow itself is often a sub-optimal representation for specific video task~\cite{xue2019video}. To overcome above limitations, we propose task-oriented flow distillation loss that can decrease the adverse impacts while focusing on the useful knowledge for better VFI.

Observing that $F_{t\rightarrow0}, F_{t\rightarrow1}$ which directly control frame synthesis are sensitive to harmful information in pseudo label. Therefore, we impose multi-scale flow distillation except for the decoder $\mathcal{D}^{1}$, and leave its flow prediction totally constrained by the reconstruction loss $\mathcal{L}_{r}$ in a task-oriented manner~\cite{xue2019video}. Furthermore, we can compare above relaxed flow prediction $F_{t\rightarrow0}, F_{t\rightarrow1}$ with pseudo label $F_{t\rightarrow0}^{p}, F_{t\rightarrow1}^{p}$ to calculate robustness masks $P_{0}, P_{1}$, and use them to adjust the robustness of distillation loss spatially in lower multiple scales for better task-oriented flow distillation, whose procedure is depicted in Figure~\ref{fig:3}. Specifically, we can obtain $P_{l} (l \in \{0, 1\})$ by the following formulation
\begin{gather}
	P_{l} = \exp (-\beta |F_{t\rightarrow l} - F_{t\rightarrow l}^{p}|_{epe}),
	\label{eq:7}
\end{gather}
where $|\cdot|_{epe}$ calculates per-pixel end-point-error, the coefficient $\beta$ controlling sensibility for robustness is set to $0.3$ according to grid search. Foundation of above operations is based on the assumption that task-oriented flow generally agrees with true optical flow but differs in some details.

\begin{figure}[t]
	\centering
	\includegraphics[width=0.94\columnwidth]{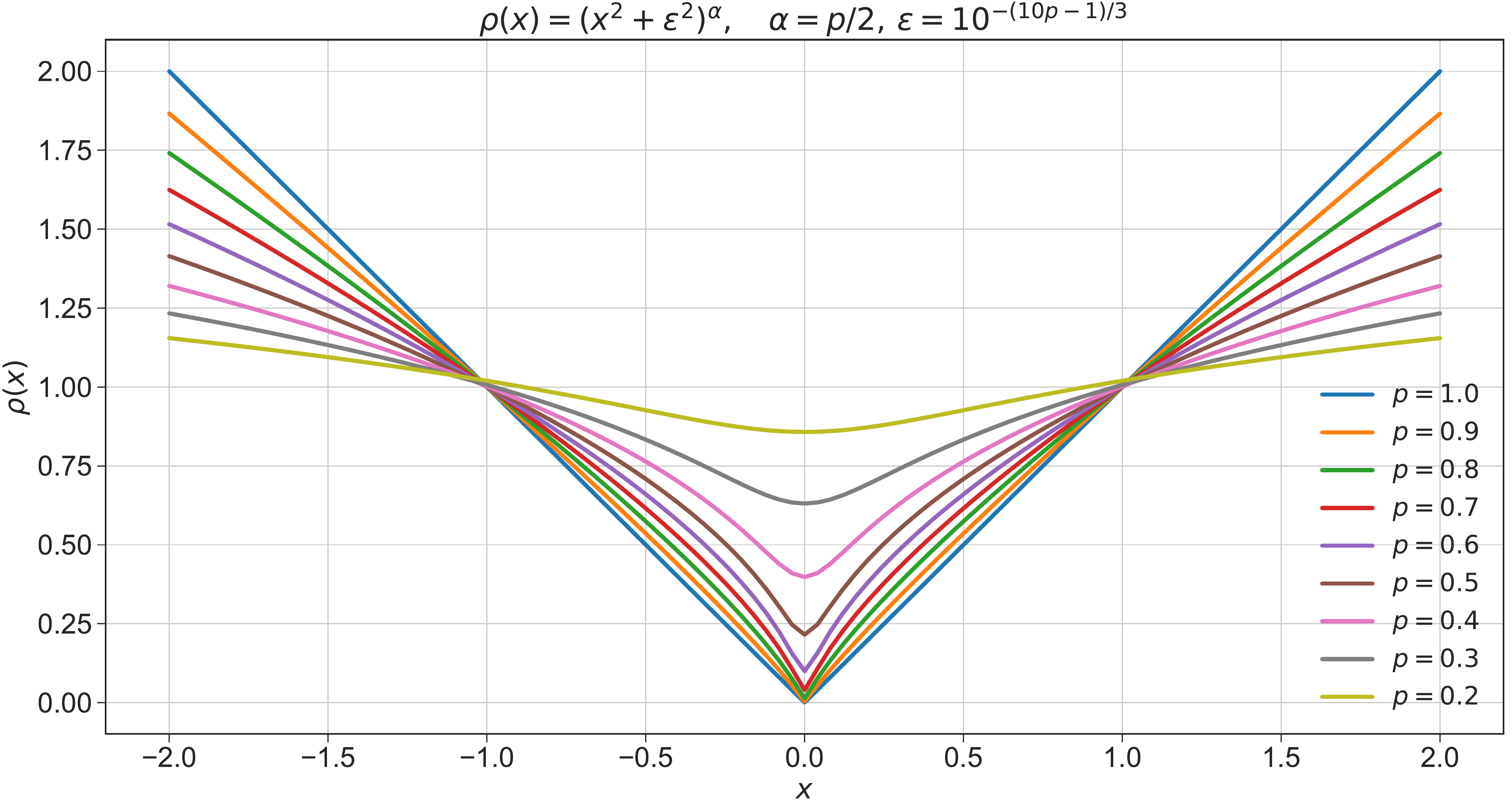}
	\vspace{-2mm}
	\caption{\textbf{Task-oriented flow distillation loss.} It takes the format of generalized Charbonnier loss, while the concrete form in each location is controlled by the corresponding robustness parameter $p$, which is determined by Eq.~\ref{eq:7} to acquire task adaptive ability.}
	\label{fig:5}
	\vspace{-3mm}
\end{figure}

Following previous experience~\cite{Sun_2014,8579034}, our task-oriented flow distillation employs the generalized Charbonnier loss $\rho(x) = (x^2 + \epsilon^2)^{\alpha}$ for better robust learning of intermediate flow, where parameters $\epsilon$ and $\alpha$ control the robustness of this loss. Formally, it can be written as
\begin{gather}
	\mathcal{L}_{d} = \sum_{k=1}^{3}\sum_{l=0}^{1} \rho(\mathcal{U}_{2^{k}} (F_{t\rightarrow l}^{k}) - F_{t\rightarrow l}^{p}),
\end{gather}
where $\mathcal{U}_{s}$ is the bilinear upsampling operation with scale factor $s$. However, different from the fixed format like previous methods~\cite{Sun_2014,8579034}, we make it adjustable about VFI task by letting $\epsilon$ and $\alpha$ be functions of the robustness parameter $p$, where $p \in (0,1]$ means the robustness value of any position in aforementioned robustness masks $P_{0}, P_{1}$. In general, we employ the linear and exponential linear functions to generate $\alpha$ and $\epsilon$ separately as follows
\vspace{-1mm}
\begin{gather}
	\alpha = p / 2, \quad \epsilon = 10^{-(10p-1)/3}.
\end{gather}
The coefficients are selected based on two typical cases. For example, when $p=1.0$, $\rho(x)$ becomes the surrogate $\mathcal{L}_1$ loss in Eq.~\ref{eq:6}. And when  $p=0.4$, it turns to be the robust loss used in LiteFlowNet~\cite{8579034}. Figure~\ref{fig:5} gives some intuitive examples of this adaptive robust loss. Comprehensively speaking, in each spatial location, if the task-oriented flow prediction of decoder $\mathcal{D}^{1}$ is consistent with that in pseudo label, the gradient of the adaptive distillation loss is relatively steep, which tends to distill this helpful information to the bottom three decoders by common gradient descent optimizer. On the other hand, the loss will become more robust to downgrade this relatively harmful flow knowledge.

\subsection{Feature Space Geometry Consistency Loss}
Besides above task-oriented flow distillation loss for facilitating multi-scale intermediate flow estimation, better supervision of intermediate feature is preferred for further improvement. Observing that extracted pyramid features $\phi_{0}^{k}, \phi_{1}^{k}$ by the encoder $\mathcal{E}$, in a sense, play an equivalent role as the reconstructed intermediate feature $\hat{\phi}_{t}^{k}$ from the decoder $\mathcal{D}^{k+1}$, we try to employ the same parameter shared encoder $\mathcal{E}$ to extract a pyramid of features $\phi_{t}^{k}$ from ground truth frame $I_{t}^{gt}$, and use $\phi_{t}^{k}$ to regularize the reconstructed intermediate feature $\hat{\phi}_{t}^{k}$ in multi-scale feature domain.

\begin{table*}[t]
	\renewcommand{\arraystretch}{0.84}
	{\small
		\centering
		\setlength\tabcolsep{4.0pt}
		\begin{tabular}{lccccccccc}
			\toprule
			\multirow{2}[2]{*}{Method} & \multirow{2}[1]{*}{Vimeo90K} & \multirow{2}[1]{*}{UCF101} & \multicolumn{4}{c}{SNU-FILM} & \multirow{2}[2]{*}{\makecell{Time \\ (s)}} & \multirow{2}[2]{*}{\makecell{Params \\ (M)}} & \multirow{2}[2]{*}{\makecell{FLOPs \\ (T)}} \\[-0.1em]
			\cmidrule(lr){4-7}
			& & & Easy & Medium & Hard & Extreme & & \\
			\midrule
			SepConv~\cite{8237299} & 33.79/0.9702 & 34.78/0.9669 & 39.41/0.9900 & 34.97/0.9762 & 29.36/0.9253 & 24.31/0.8448 & 0.065 & 21.7 & 0.36 \\
			CAIN~\cite{choi2020cain} & 34.65/0.9730 & 34.91/0.9690 & 39.89/0.9900 & 35.61/0.9776 & 29.90/0.9292 & 24.78/0.8507 & 0.069 & 42.8 & 1.29 \\
			AdaCoF~\cite{Lee_2020_CVPR} & 34.47/0.9730 & 34.90/0.9680 & 39.80/0.9900 & 35.05/0.9754 & 29.46/0.9244 & 24.31/0.8439 & 0.054 & 21.8 & 0.36 \\
			RIFE~\cite{huang2021rife} & 35.62/0.9780 & 35.28/0.9690 & 40.06/\textcolor{red}{\bf 0.9907} & 35.75/0.9789 & 30.10/0.9330 & 24.84/0.8534 & 0.026 & 9.8 & \textcolor{blue}{\underline{0.20}} \\
			IFRNet & 35.80/0.9794 & 35.29/\textcolor{blue}{\underline{0.9693}} & 40.03/0.9905 & \textcolor{blue}{\underline{35.94}}/\textcolor{blue}{\underline{0.9793}} & 30.41/0.9358 & 25.05/0.8587 & \textcolor{blue}{\underline{0.025}} & 5.0 & 0.21 \\
			IFRNet small & 35.59/0.9786 & 35.28/0.9691 & 39.96/0.9905 & 35.92/0.9792 & 30.36/0.9357 & 25.05/0.8582 & \textcolor{red}{\bf 0.019} & \textcolor{blue}{\underline{2.8}} & \textcolor{red}{\bf 0.12} \\
			\midrule
			ToFlow~\cite{xue2019video} & 33.73/0.9682 & 34.58/0.9667 & 39.08/0.9890 & 34.39/0.9740 & 28.44/0.9180 & 23.39/0.8310 & 0.152 & \textcolor{red}{\bf 1.4} & 0.62 \\
			CyclicGen~\cite{liu2019cyclicgen} & 32.09/0.9490 & 35.11/0.9684 & 37.72/0.9840 & 32.47/0.9554 & 26.95/0.8871 & 22.70/0.8083 & 0.161 & 19.8 & 1.77 \\
			DAIN~\cite{8954114} & 34.71/0.9756 & 34.99/0.9683 & 39.73/0.9902 & 35.46/0.9780 & 30.17/0.9335 & 25.09/0.8584 & 1.033 & 24.0 & 5.51 \\
			SoftSplat~\cite{Niklaus_2020_CVPR} & 36.10/0.9700 & \textcolor{blue}{\underline{35.39}}/0.9520 & - & - & - & - & 0.195 & 12.2 & 0.90 \\
			BMBC~\cite{BMBC} & 35.01/0.9764 & 35.15/0.9689 & 39.90/0.9902 & 35.31/0.9774 & 29.33/0.9270 & 23.92/0.8432 & 3.845 & 11.0 & 2.50 \\
			CDFI full~\cite{ding2021cdfi} & 35.17/0.9640 & 35.21/0.9500 & \textcolor{red}{\bf 40.12}/\textcolor{blue}{\underline{0.9906}} & 35.51/0.9778 & 29.73/0.9277 & 24.53/0.8476 & 0.380 & 5.0 & 0.82 \\
			ABME~\cite{park2021asymmetric} & \textcolor{blue}{\underline{36.18}}/\textcolor{blue}{\underline{0.9805}} & 35.38/\textcolor{red}{\bf 0.9698} & 39.59/0.9901 & 35.77/0.9789 & \textcolor{blue}{\underline{30.58}}/\textcolor{blue}{\underline{0.9364}} & \textcolor{red}{\bf 25.42}/\textcolor{red}{\bf 0.8639} & 0.905 & 18.1 & 1.30 \\
			IFRNet large & \textcolor{red}{\bf 36.20}/\textcolor{red}{\bf 0.9808} & \textcolor{red}{\bf 35.42}/\textcolor{red}{\bf 0.9698} & \textcolor{blue}{\underline{40.10}}/\textcolor{blue}{\underline{0.9906}} & \textcolor{red}{\bf 36.12}/\textcolor{red}{\bf 0.9797} & \textcolor{red}{\bf 30.63}/\textcolor{red}{\bf 0.9368} & \textcolor{blue}{\underline{25.27}}/\textcolor{blue}{\underline{0.8609}} & 0.079 & 19.7 & 0.79 \\
			\bottomrule
		\end{tabular}
		\vspace{-2mm}
		\caption{\textbf{Quantitative comparison (PSNR/SSIM) of VFI results on the Vimeo90K, UCF101 and SNU-FILM datasets.} For each item, the best result is \textcolor{red}{\textbf{boldfaced}}, and the second best is \textcolor{blue}{\underline{underlined}}. Top and bottom parts are divided by running time.}
		\label{tab:1}}
	\vspace{-3mm}
\end{table*}

Intuitively, we can adopt the commonly used $\mathcal{L}_{1}$ loss to restrict $\hat{\phi}_{t}^{k}$ to be close to $\phi_{t}^{k}$. However, the overtighten constraint will harm the global context and occlusion information contained in reconstructed intermediate feature $\hat{\phi}_{t}^{k}$. To relax it and inspired by the local geometry alignment property of census transform~\cite{Zabih_1994}, we extend the census loss $\mathcal{L}_{cen}$~\cite{Meister_2018} into multi-scale feature space for progressive supervision, where the soft Hamming distance is calculated between census-transformed corresponding feature maps with 3$\times$3 patches in a channel-by-channel manner. Formally, this loss can be written as
\vspace{-1.5mm}
\begin{equation}
	\mathcal{L}_{g} = \sum_{k=1}^{3} \mathcal{L}_{cen}(\hat{\phi}_{t}^{k}, \phi_{t}^{k}).
\end{equation}
Our motivation is that the extracted pyramid feature, containing useful low-level structure information for frame synthesizing, can regularize the reconstructed intermediate feature to keep better geometry layout. For each spatial location, $\mathcal{L}_{g}$ only constrain the geometry of its neighbor local patch in every feature map. Consequently, there is no restriction on the channel-wise representation for $\hat{\phi}_{t}^{k}$ to encode bilateral occlusion and residual information.

Based on above analysis, our final loss function, containing three parts for joint optimization, is formulated as
\begin{equation}
	\mathcal{L} = \mathcal{L}_{r} + \lambda \mathcal{L}_{d} + \eta \mathcal{L}_{g},
\end{equation}
where weighting parameters are set to $\lambda=0.01, \eta=0.01$.

\section{Experiments}
In this section, we first introduce implementation details and datasets used in this paper. Then, we quantitatively and qualitatively compare IFRNet with recent state-of-the-arts on various benchmarks. Finally, ablation studies are carried out to analyze the contribution of proposed approaches. Experiments in the main paper follow a common practice of $t=0.5$, that is synthesizing the single middle frame. IFRNet also supports multi-frame interpolation with temporal encoding $T$, whose results are presented in supplementary.
\subsection{Implementation Details}
We implement proposed algorithm in PyTorch, and use Vimeo90K~\cite{xue2019video} training set to train IFRNet from scratch. Our model is optimized by AdamW~\cite{Loshchilov_2019} algorithm for 300 epochs with total batch size 24 on four NVIDIA Tesla V100 GPUs. The learning rate is initially set to $1 \times 10^{-4}$, and gradually decays to $1 \times 10^{-5}$ following a cosine attenuation schedule. During training, we augment the samples by random flipping, rotating, reversing sequence order and random cropping patches with size 224 $\times$ 224. For optical flow distillation, we extract pseudo label of bilateral intermediate flow fields with the pre-trained LiteFlowNet~\cite{8579034} in advance, and perform consistent augmentation operations with frame triplets during the whole training process.
\subsection{Evaluation Metrics and Datasets}
We evaluate our method on various datasets covering diverse motion scenes for comprehensive comparison. Common metrics, such as PSNR and SSIM~\cite{1284395} are adopted for quantitative evaluation. For Middlebury, we use the official IE and NIE indices. Now, we briefly introduce the used test datasets to assess our approaches.

\noindent\textbf{Vimeo90K~\cite{xue2019video}:} It contains frame triplets of 448$\times$256 resolution. There are 3,782 triplets consisted in the test part.

\noindent\textbf{UCF101~\cite{Soomro_2012}:}
We adopt the test set selected in DVF~\cite{8237740}, which includes 379 triplets of 256$\times$256 frame size.

\noindent\textbf{SNU-FILM~\cite{choi2020cain}:}
SNU-FILM contains 1,240 frame triplets of approximate 1280$\times$720 resolution. According to motion magnitude, it is divided into four different parts, namely, Easy, Medium, Hard, and Extreme for detailed comparison.

\noindent\textbf{Middlebury~\cite{Baker_2011}:}
The Middlebury benchmark is a widely used dataset to evaluate optical flow and VFI methods. Image resolution in this dataset is around 640$\times$480. In this paper, we test on the Evaluation set without using Other set.

\subsection{Comparison with the State-of-the-Arts}
We compare IFRNet with state-of-the-art VFI methods, including kernel-based SepConv~\cite{8237299}, AdaCoF~\cite{Lee_2020_CVPR} and CDFI~\cite{ding2021cdfi}, flow-based ToFlow~\cite{xue2019video}, DAIN~\cite{8954114}, SoftSplat~\cite{Niklaus_2020_CVPR}, BMBC~\cite{BMBC}, RIFE~\cite{huang2021rife} and ABME~\cite{park2021asymmetric}, and hallucination-based CAIN~\cite{choi2020cain} and FeFlow~\cite{Gui_2020_CVPR}. For results on SNU-FILM, we execute the released codes of CDFI and RIFE and refer to the other results tested in ABME. For Middlebury, we directly test on the Evaluation part and submit interpolation results to the online benchmark. To measure the inference speed and computation complexity, we run all methods on one Tesla V100 GPU under 1280$\times$720 resolution and average the running time with 100 iterations. For fair comparison, we further build a large and a small version of IFRNet by scaling feature channels with 2.0 and 0.75, respectively, and separate above methods into two classes, \textit{i.e.}, \textit{fast} and \textit{slow}, according to their inference time.

\begin{figure*}[t]
	\centering
	{\includegraphics[width=0.108\linewidth]{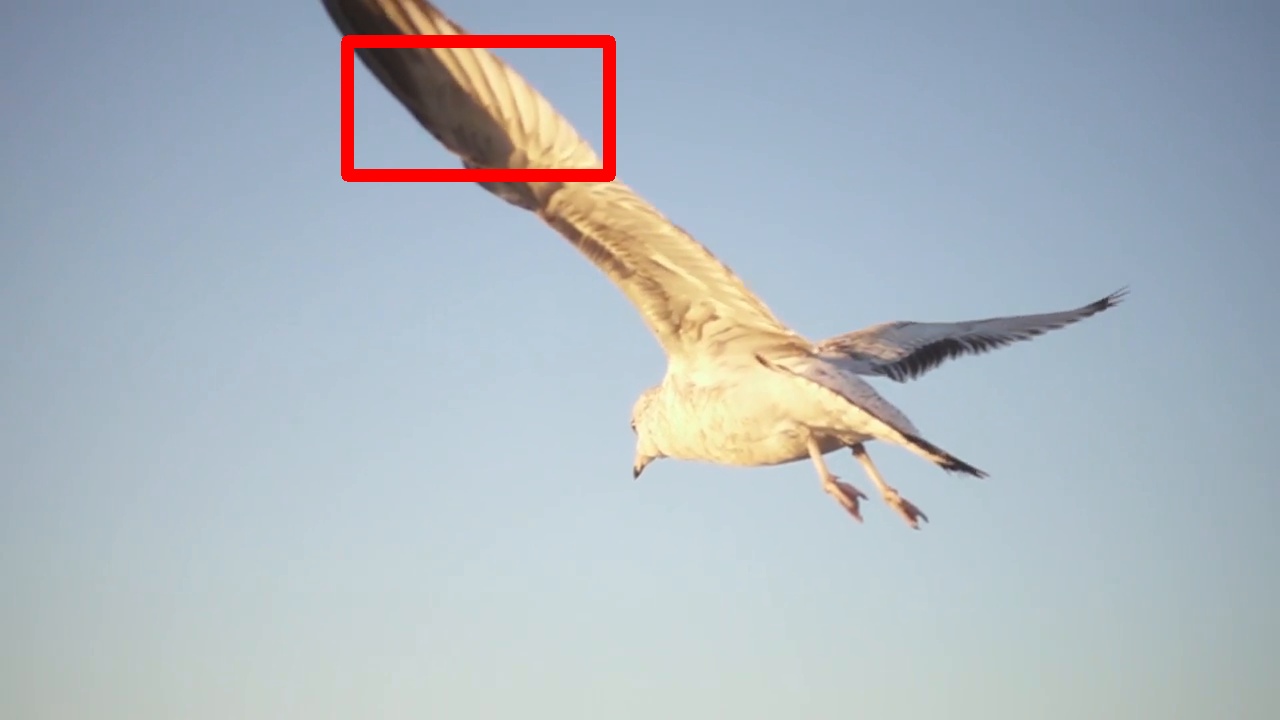}}\hspace{-1pt}
	{\includegraphics[width=0.108\linewidth]{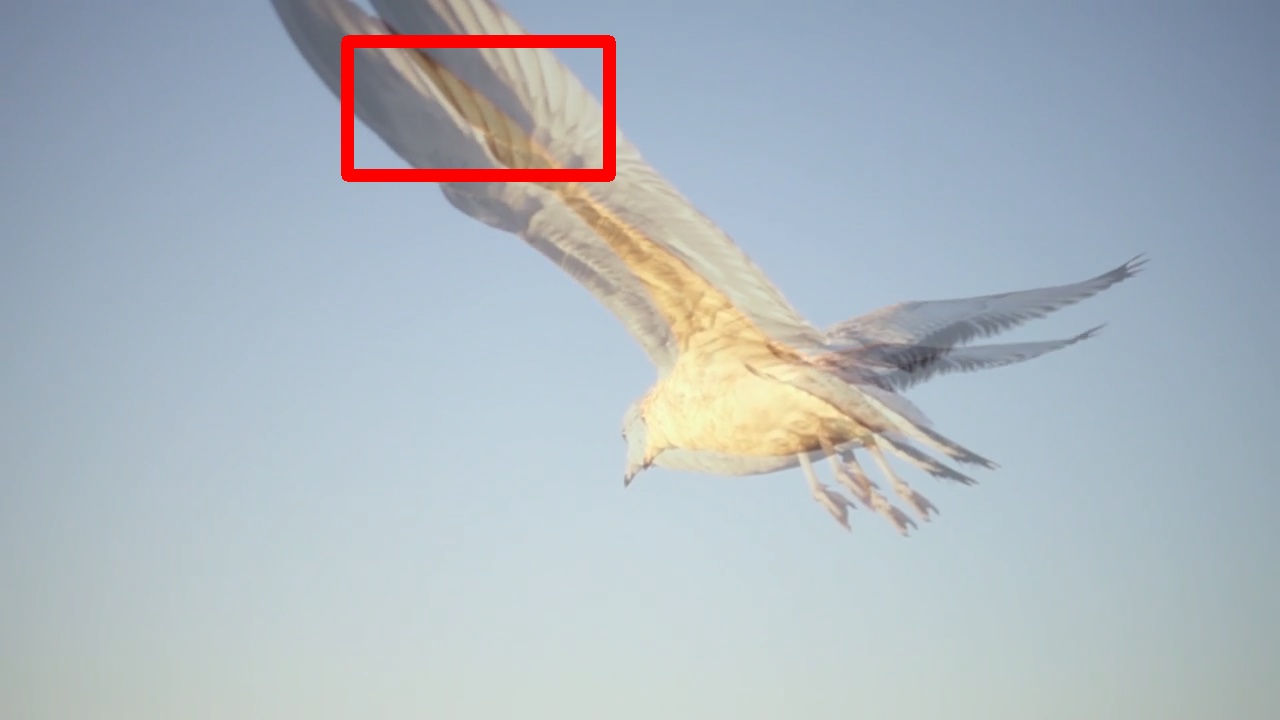}}\hspace{-1pt}
	{\includegraphics[width=0.108\linewidth]{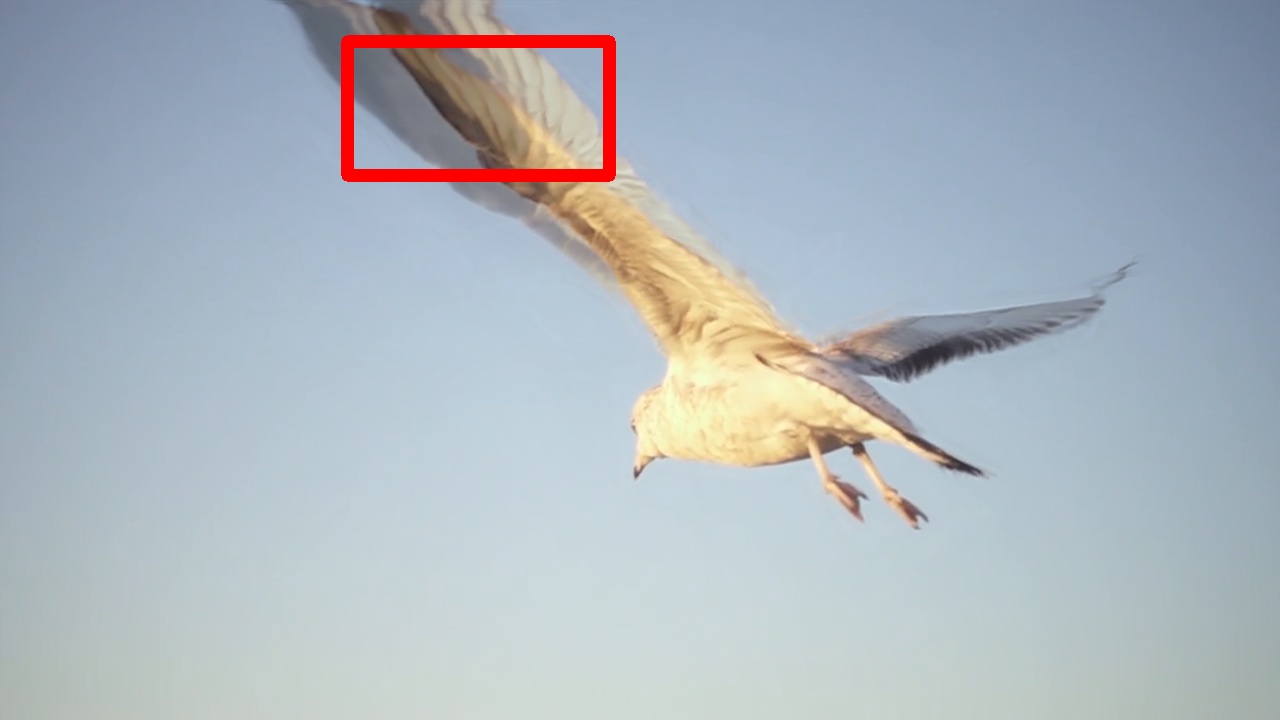}}\hspace{-1pt}
	{\includegraphics[width=0.108\linewidth]{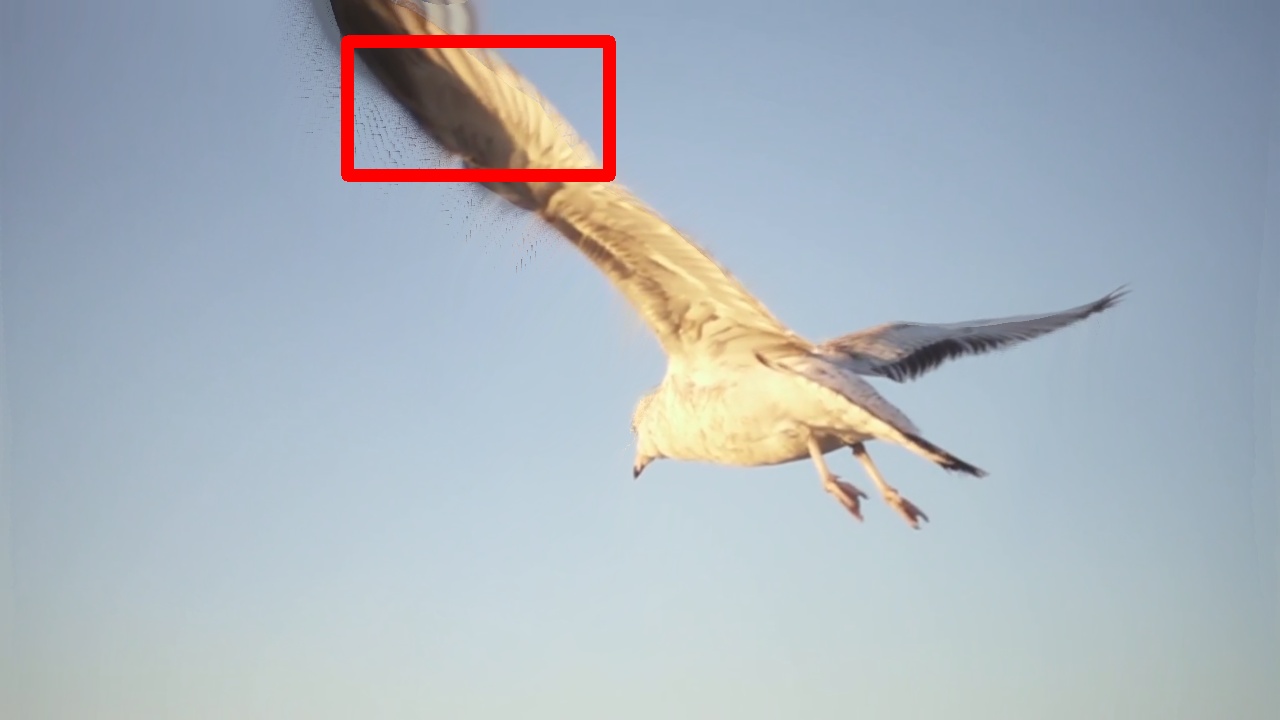}}\hspace{-1pt}
	{\includegraphics[width=0.108\linewidth]{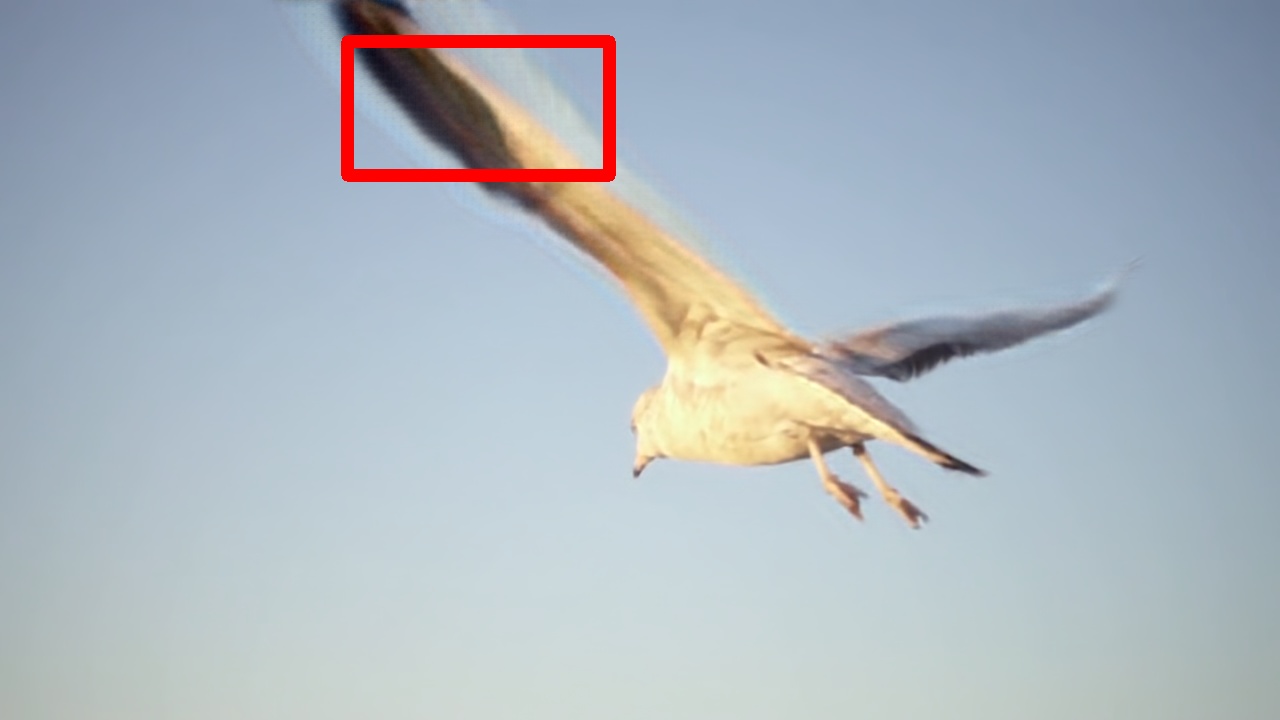}}\hspace{-1pt}
	{\includegraphics[width=0.108\linewidth]{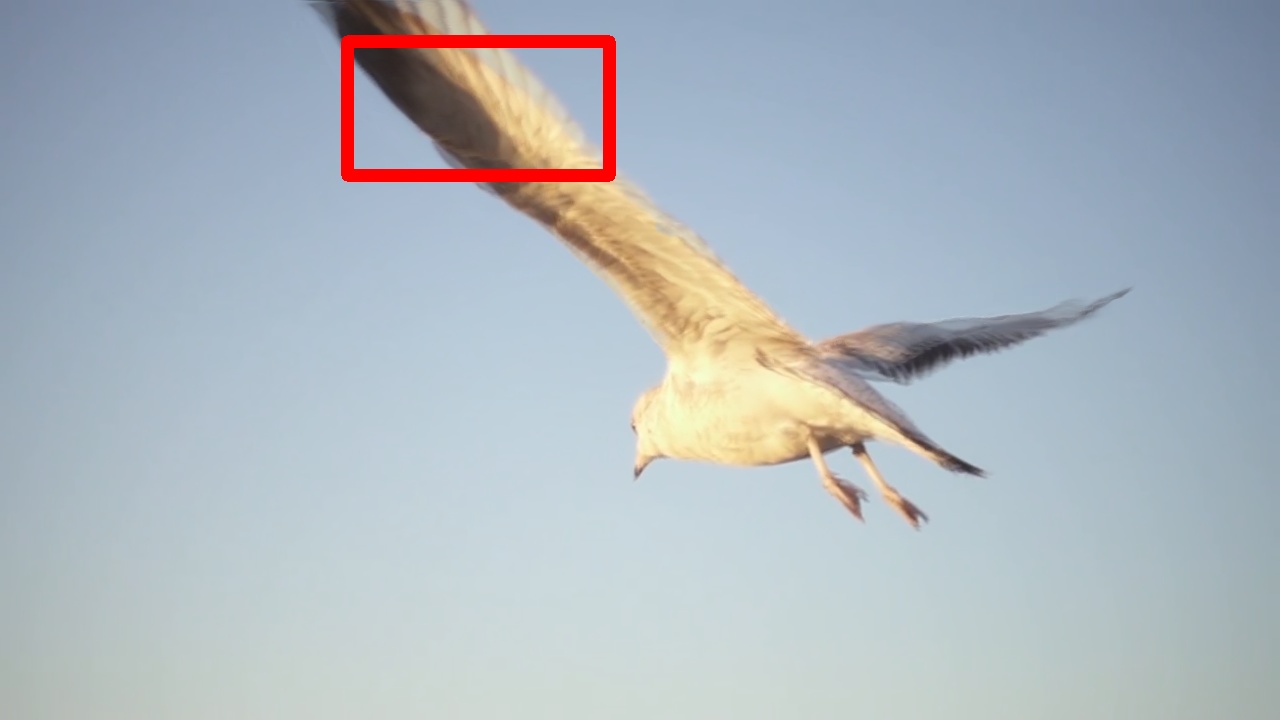}}\hspace{-1pt}
	{\includegraphics[width=0.108\linewidth]{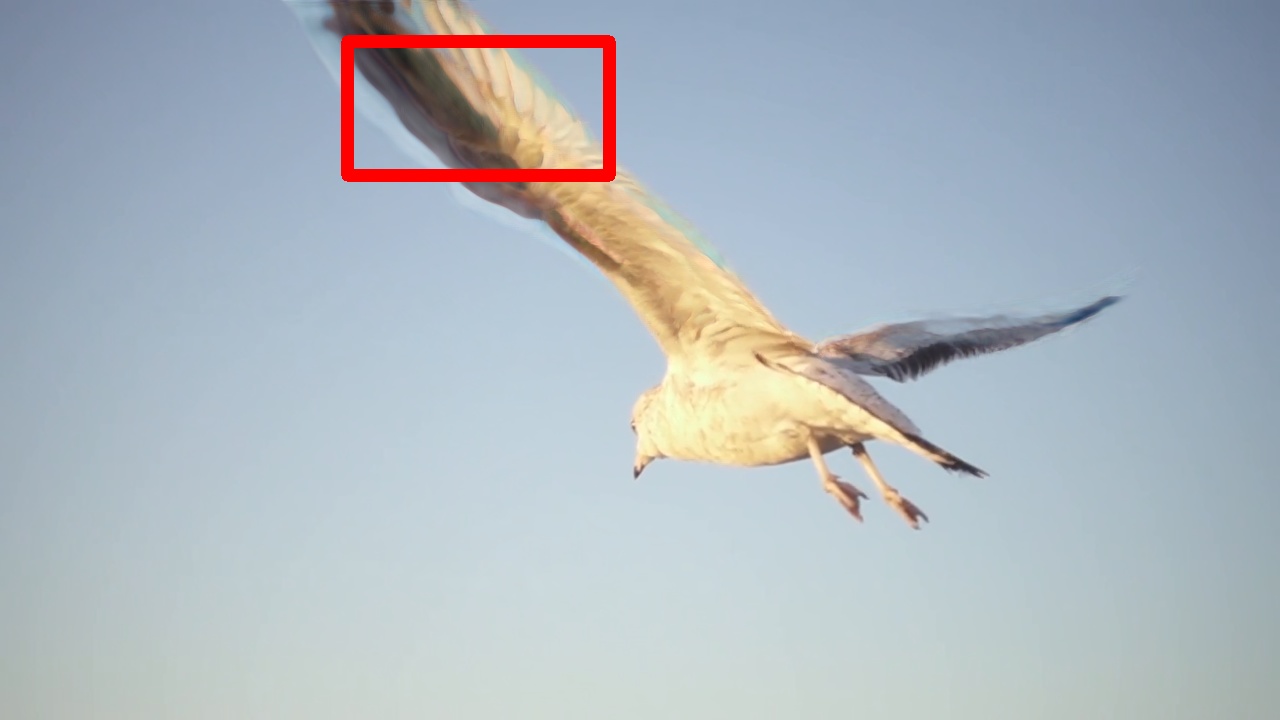}}\hspace{-1pt}
	{\includegraphics[width=0.108\linewidth]{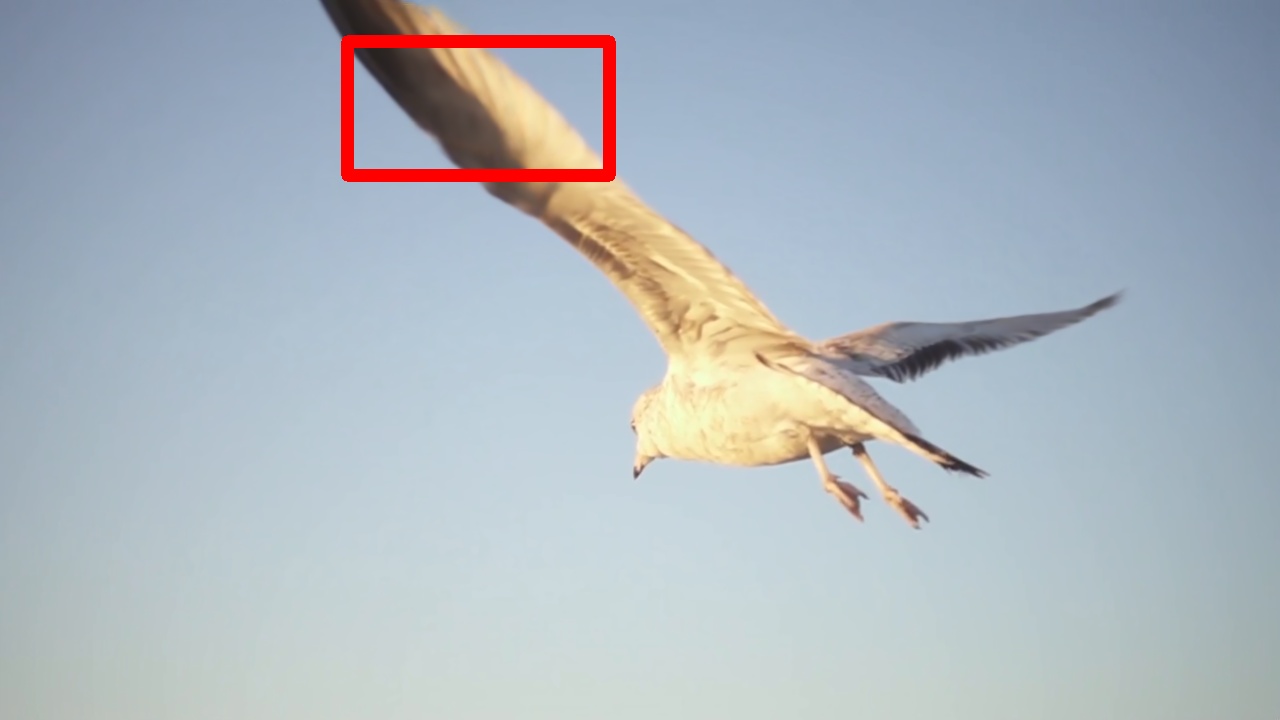}}\hspace{-1pt}
	{\includegraphics[width=0.108\linewidth]{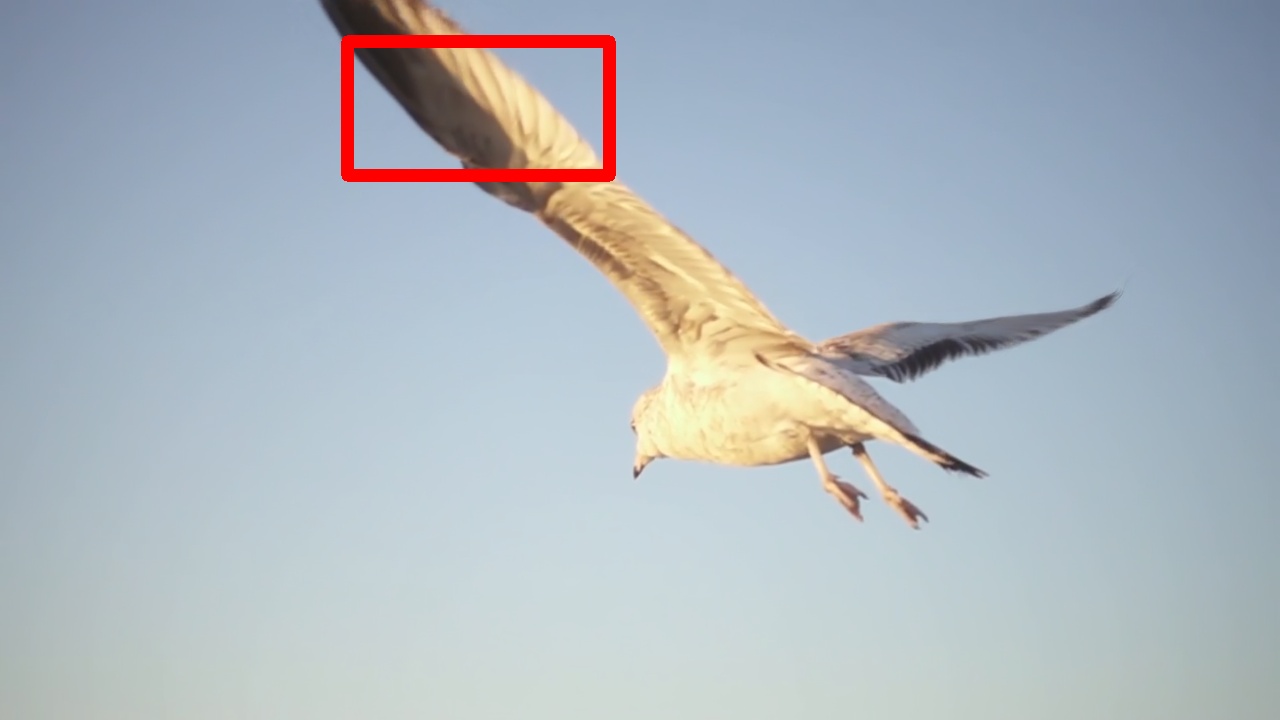}}\newline
	{\includegraphics[width=0.108\linewidth]{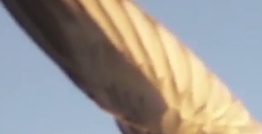}}\hspace{-1pt}
	{\includegraphics[width=0.108\linewidth]{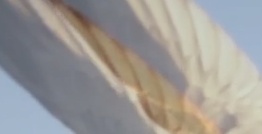}}\hspace{-1pt}
	{\includegraphics[width=0.108\linewidth]{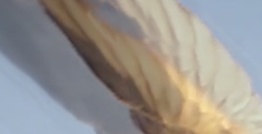}}\hspace{-1pt}
	{\includegraphics[width=0.108\linewidth]{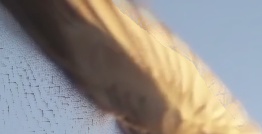}}\hspace{-1pt}
	{\includegraphics[width=0.108\linewidth]{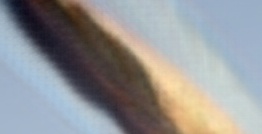}}\hspace{-1pt}
	{\includegraphics[width=0.108\linewidth]{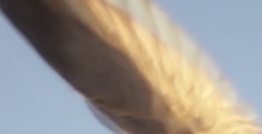}}\hspace{-1pt}
	{\includegraphics[width=0.108\linewidth]{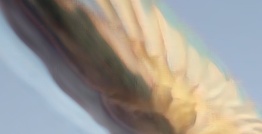}}\hspace{-1pt}
	{\includegraphics[width=0.108\linewidth]{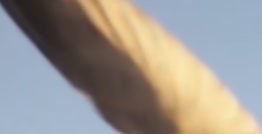}}\hspace{-1pt}
	{\includegraphics[width=0.108\linewidth]{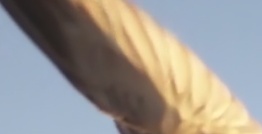}}\newline
	{\includegraphics[width=0.108\linewidth]{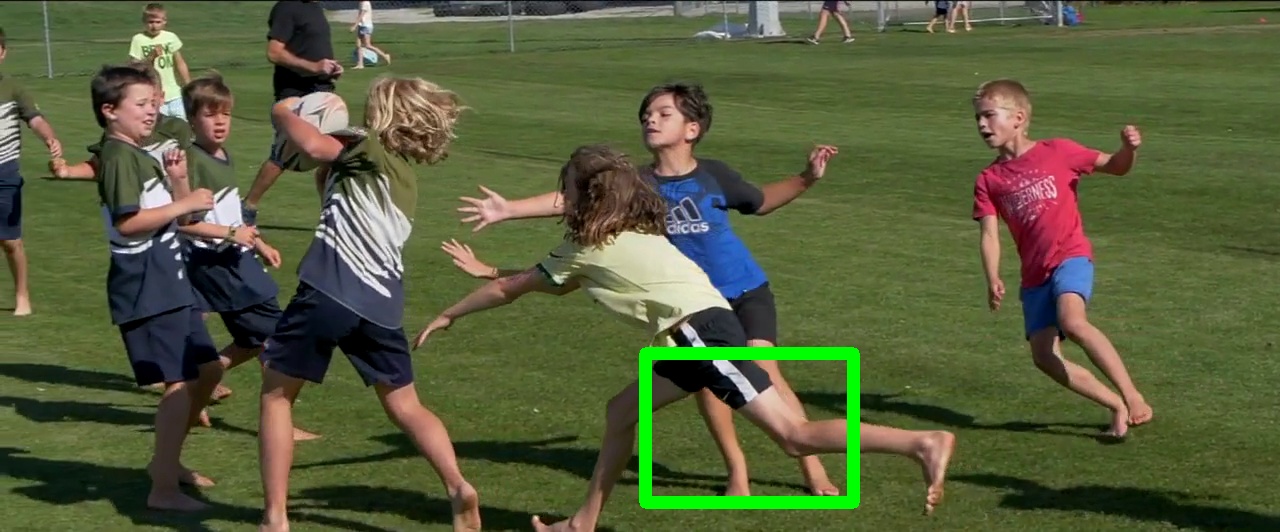}}\hspace{-1pt}
	{\includegraphics[width=0.108\linewidth]{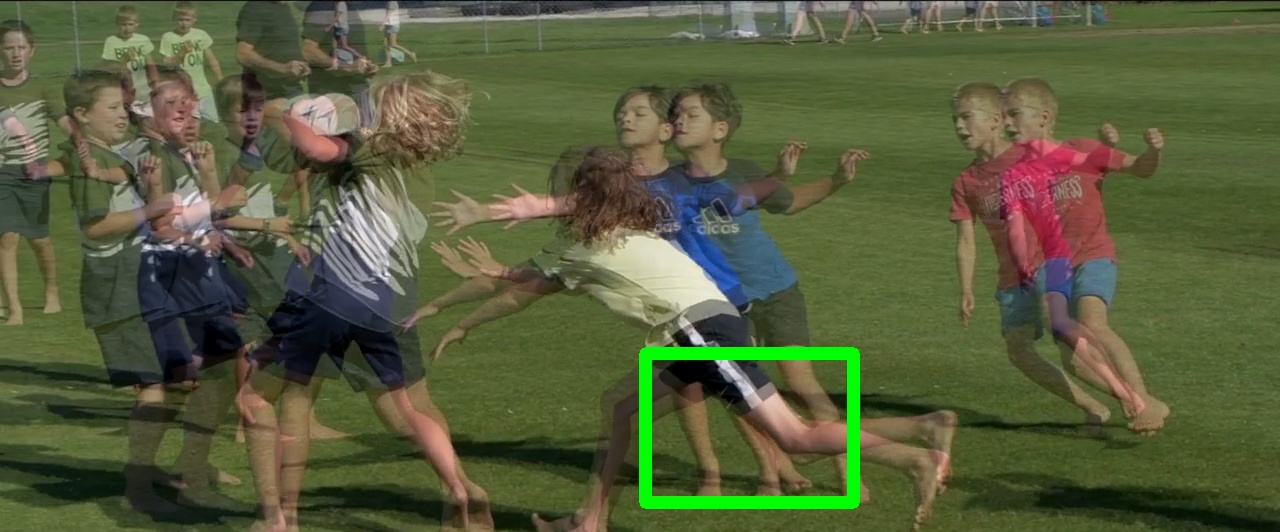}}\hspace{-1pt}
	{\includegraphics[width=0.108\linewidth]{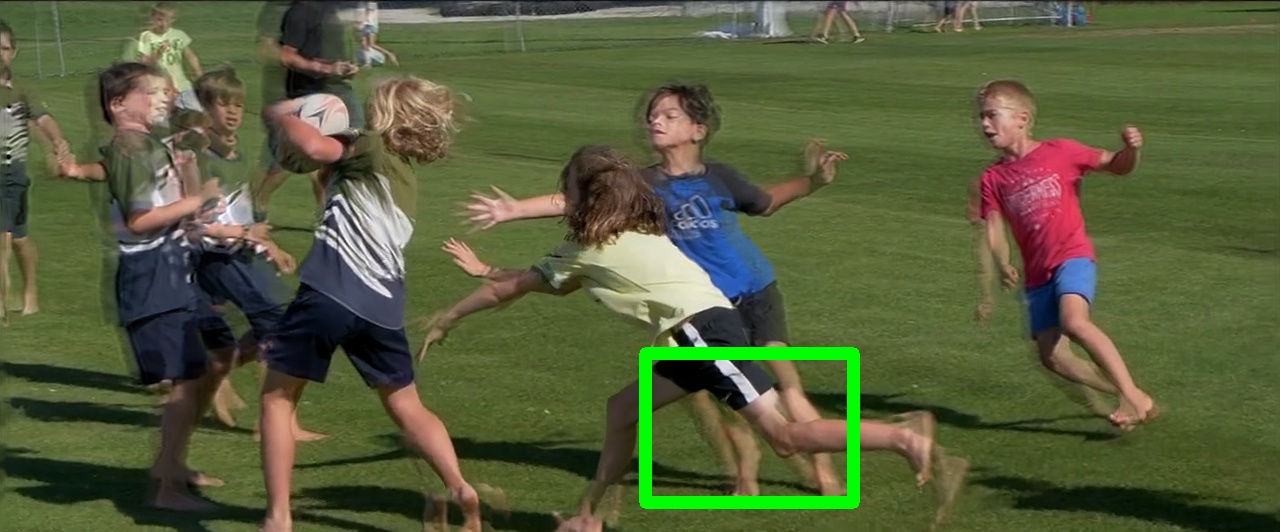}}\hspace{-1pt}
	{\includegraphics[width=0.108\linewidth]{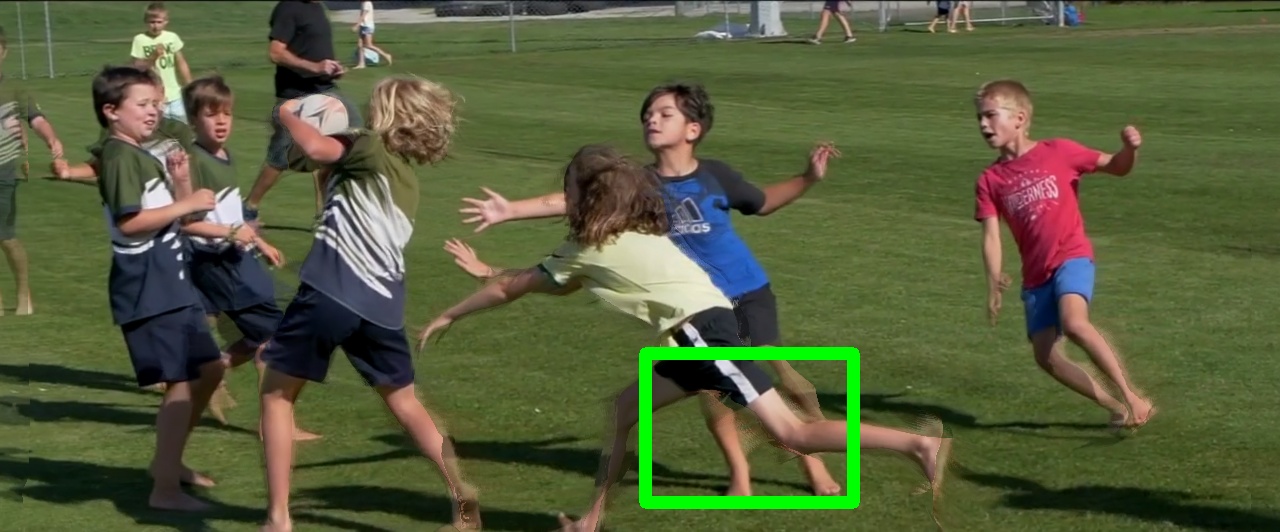}}\hspace{-1pt}
	{\includegraphics[width=0.108\linewidth]{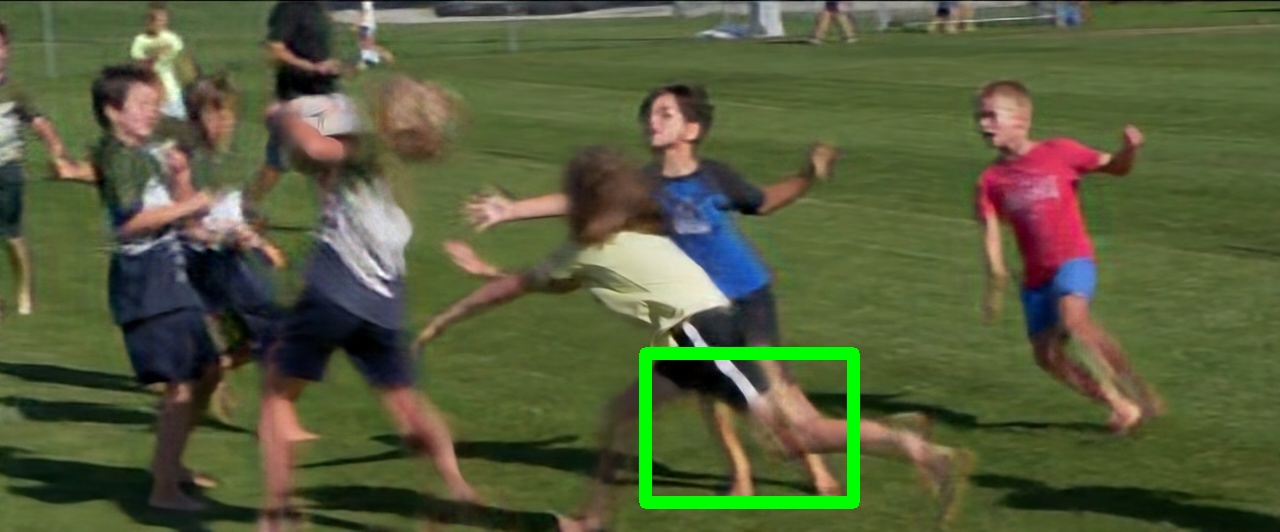}}\hspace{-1pt}
	{\includegraphics[width=0.108\linewidth]{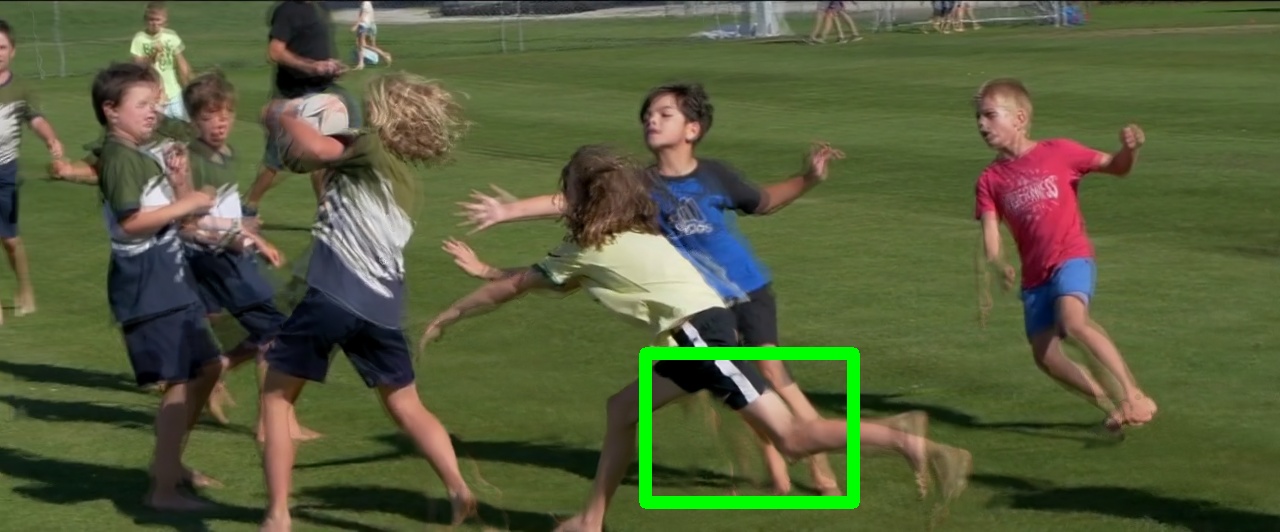}}\hspace{-1pt}
	{\includegraphics[width=0.108\linewidth]{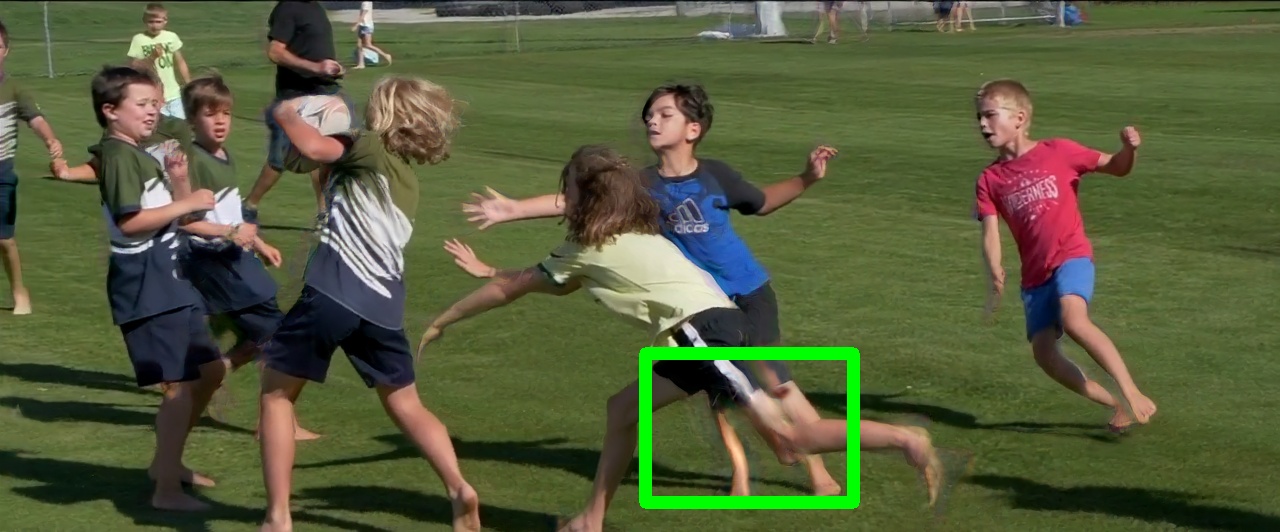}}\hspace{-1pt}
	{\includegraphics[width=0.108\linewidth]{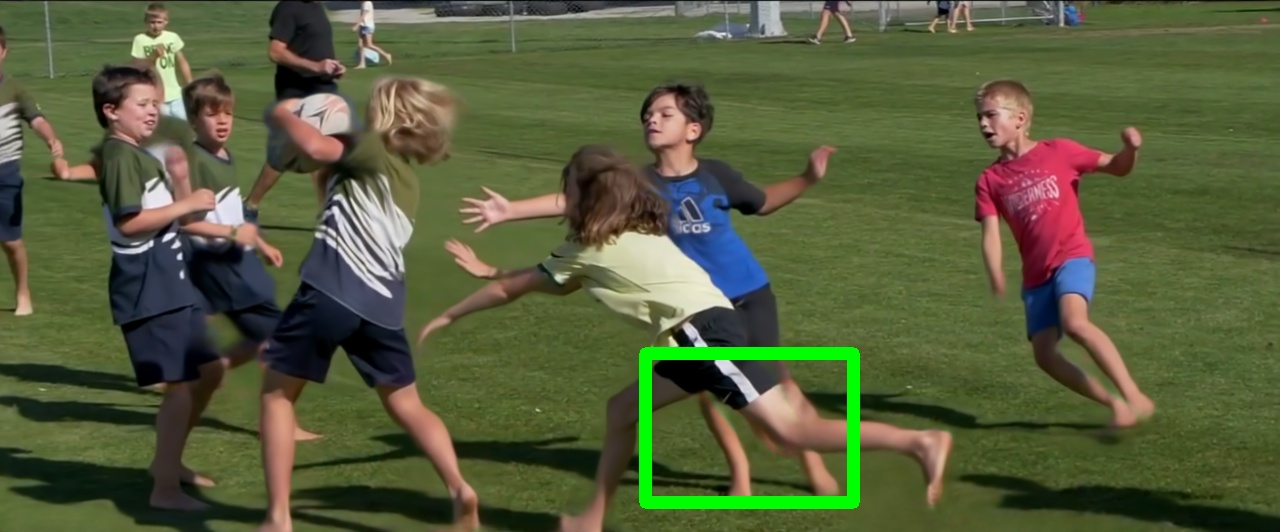}}\hspace{-1pt}
	{\includegraphics[width=0.108\linewidth]{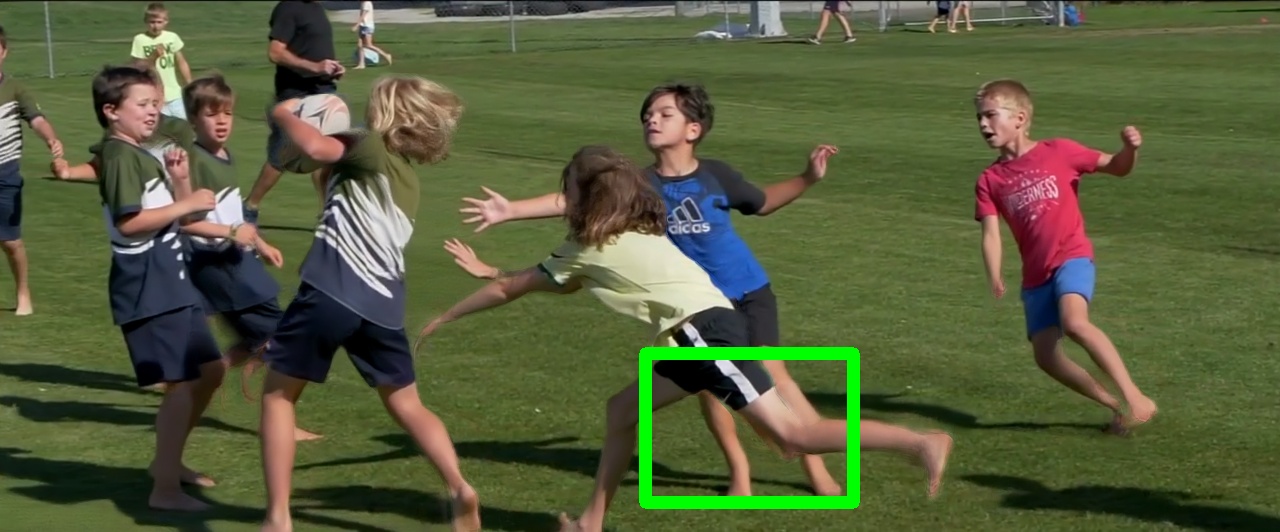}}\newline
	{\includegraphics[width=0.108\linewidth]{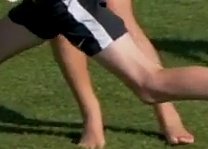}}\hspace{-1pt}
	{\includegraphics[width=0.108\linewidth]{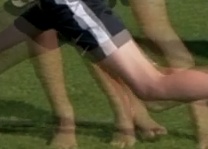}}\hspace{-1pt}
	{\includegraphics[width=0.108\linewidth]{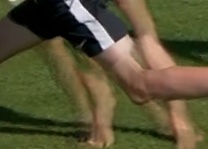}}\hspace{-1pt}
	{\includegraphics[width=0.108\linewidth]{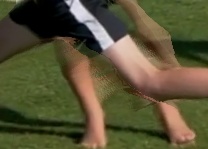}}\hspace{-1pt}
	{\includegraphics[width=0.108\linewidth]{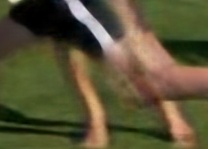}}\hspace{-1pt}
	{\includegraphics[width=0.108\linewidth]{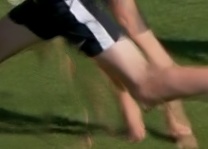}}\hspace{-1pt}
	{\includegraphics[width=0.108\linewidth]{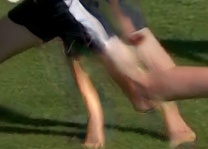}}\hspace{-1pt}
	{\includegraphics[width=0.108\linewidth]{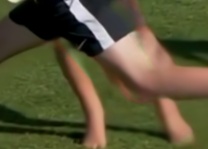}}\hspace{-1pt}
	{\includegraphics[width=0.108\linewidth]{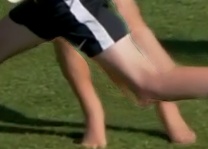}}\newline
	{\includegraphics[width=0.108\linewidth]{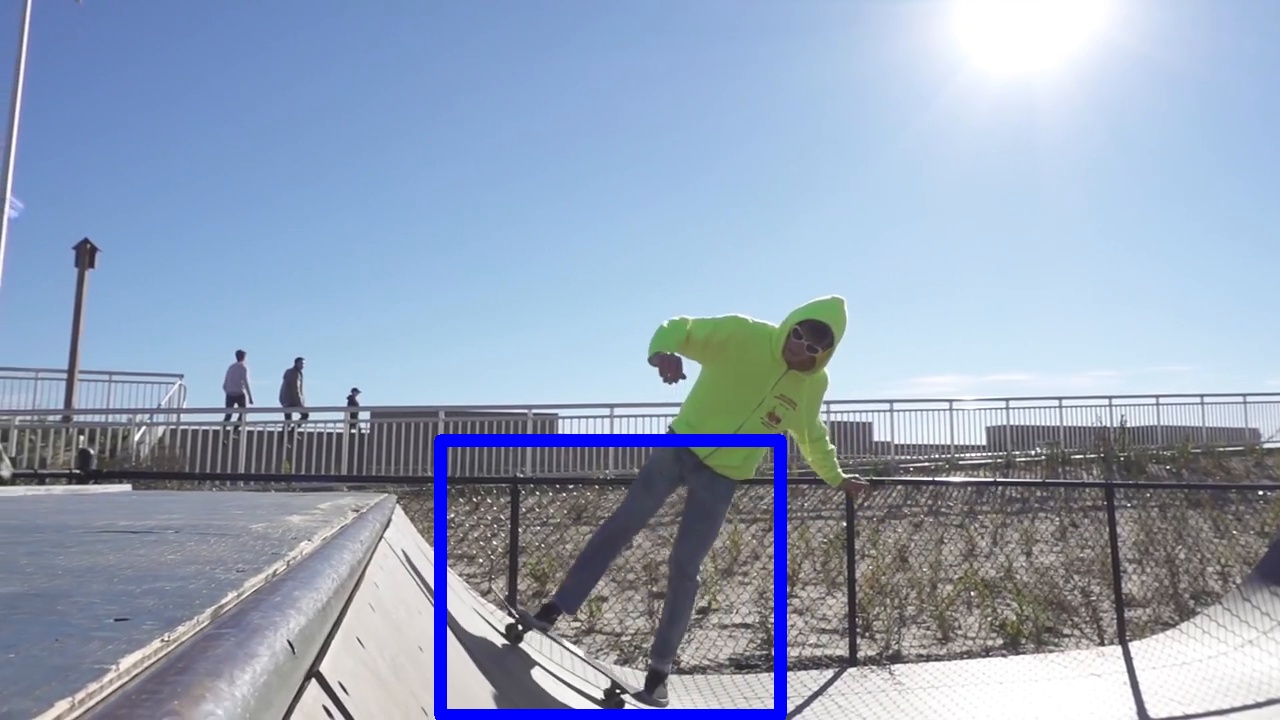}}\hspace{-1pt}
	{\includegraphics[width=0.108\linewidth]{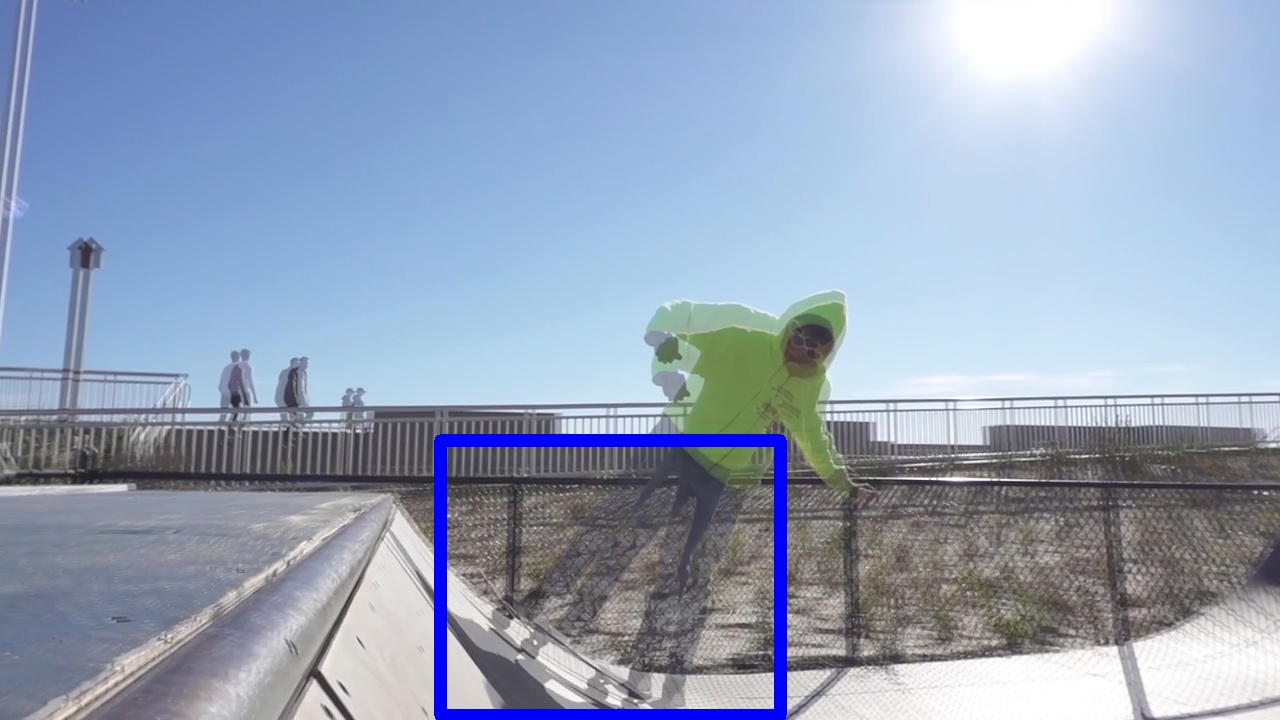}}\hspace{-1pt}
	{\includegraphics[width=0.108\linewidth]{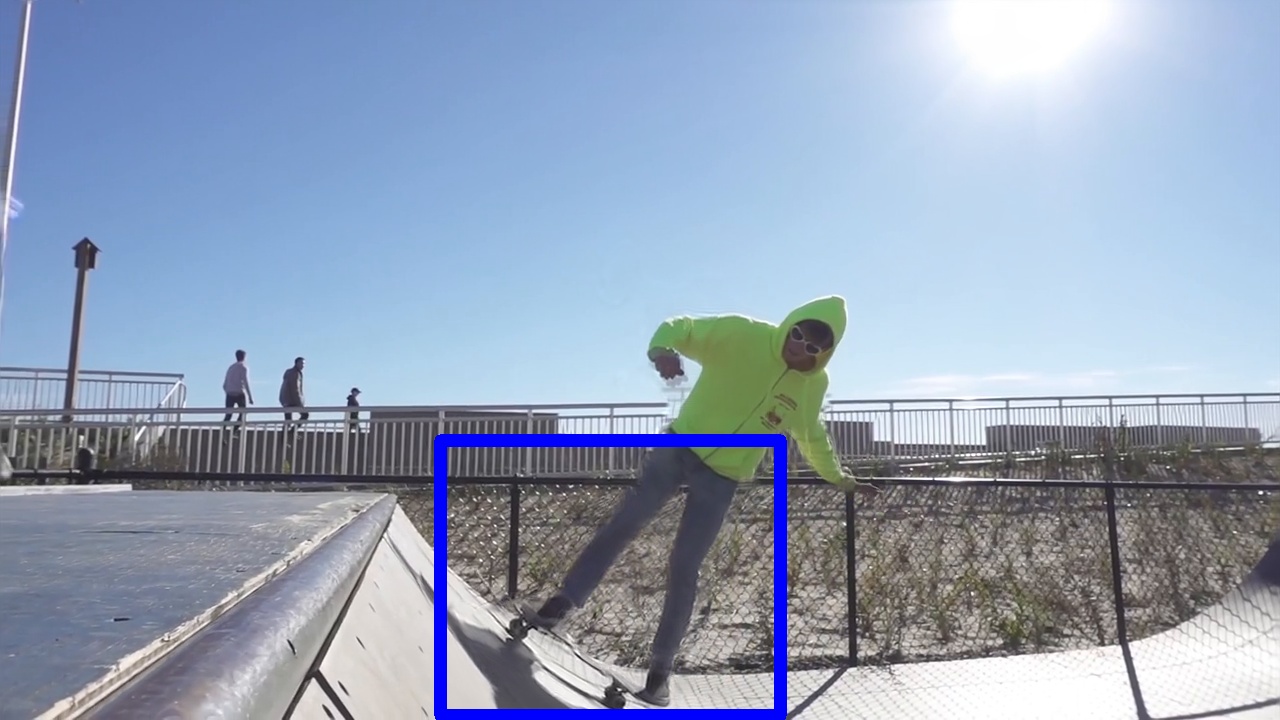}}\hspace{-1pt}
	{\includegraphics[width=0.108\linewidth]{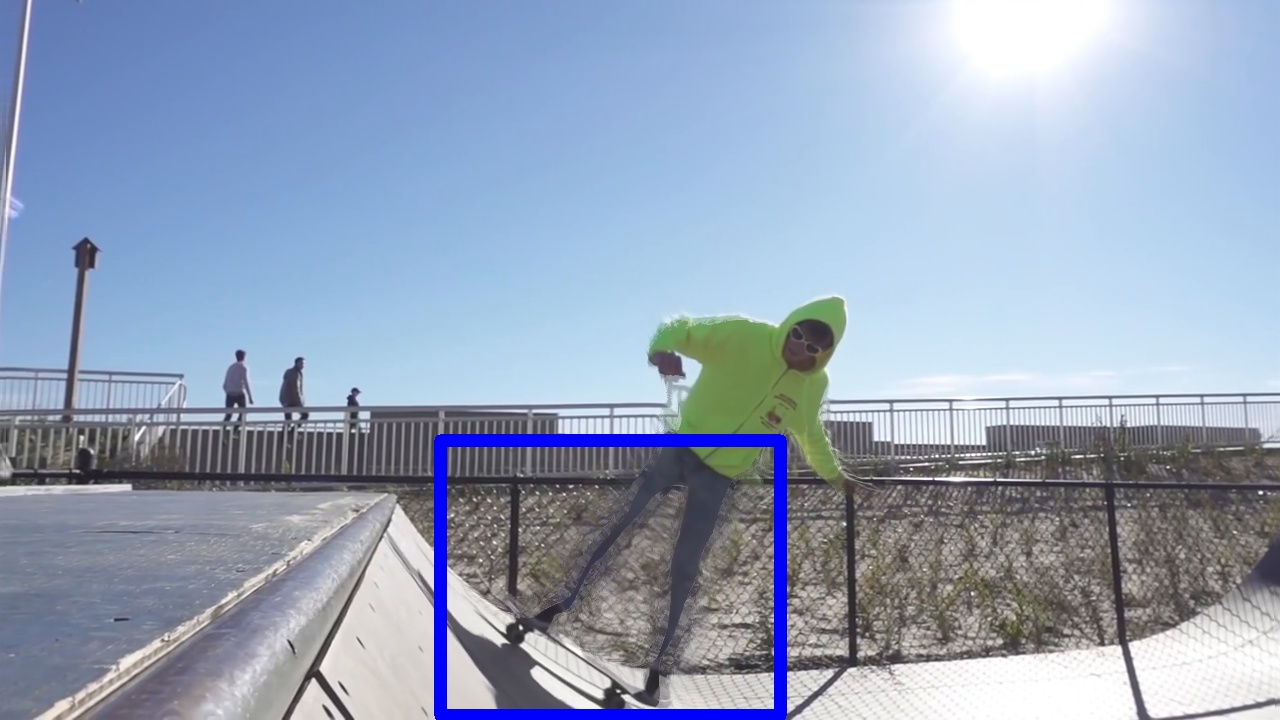}}\hspace{-1pt}
	{\includegraphics[width=0.108\linewidth]{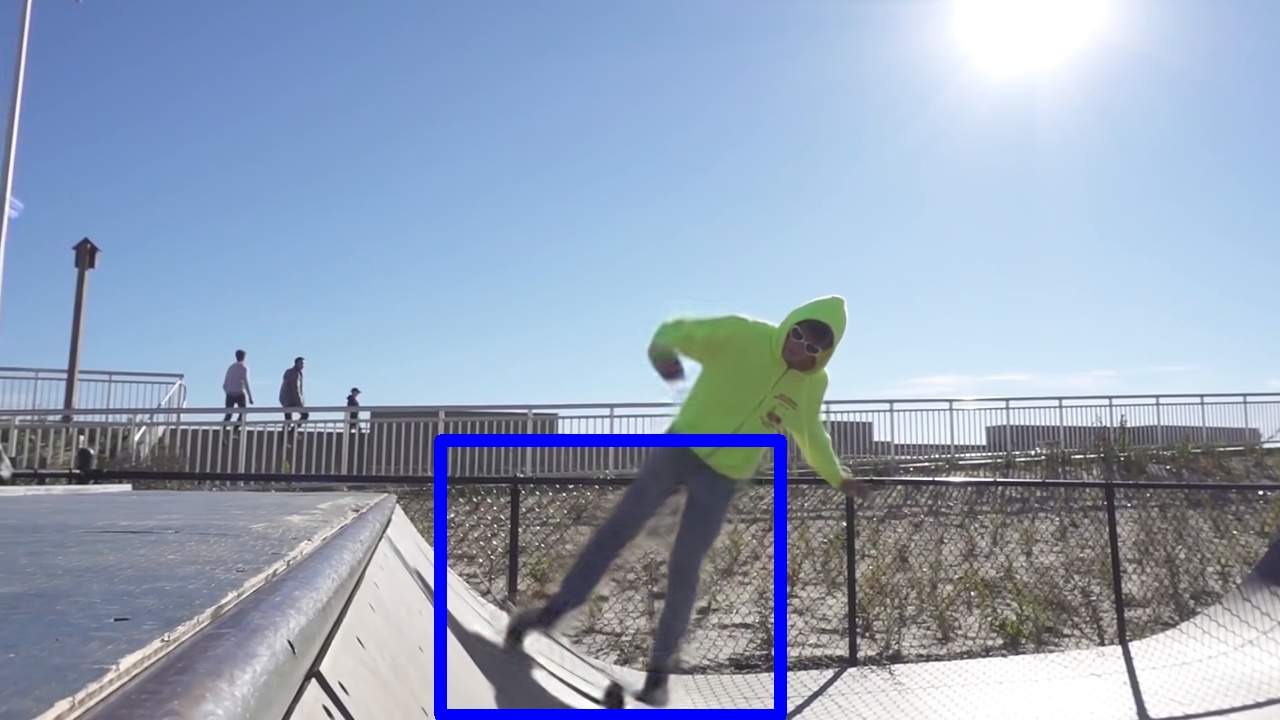}}\hspace{-1pt}
	{\includegraphics[width=0.108\linewidth]{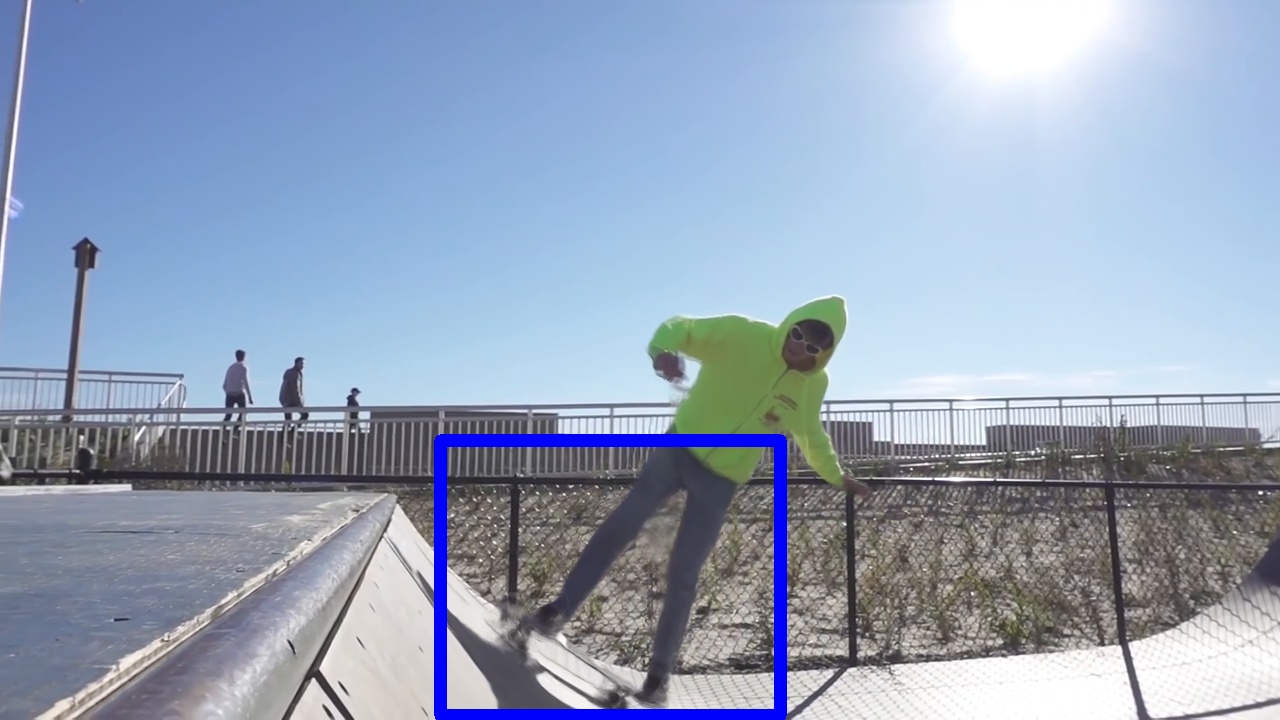}}\hspace{-1pt}
	{\includegraphics[width=0.108\linewidth]{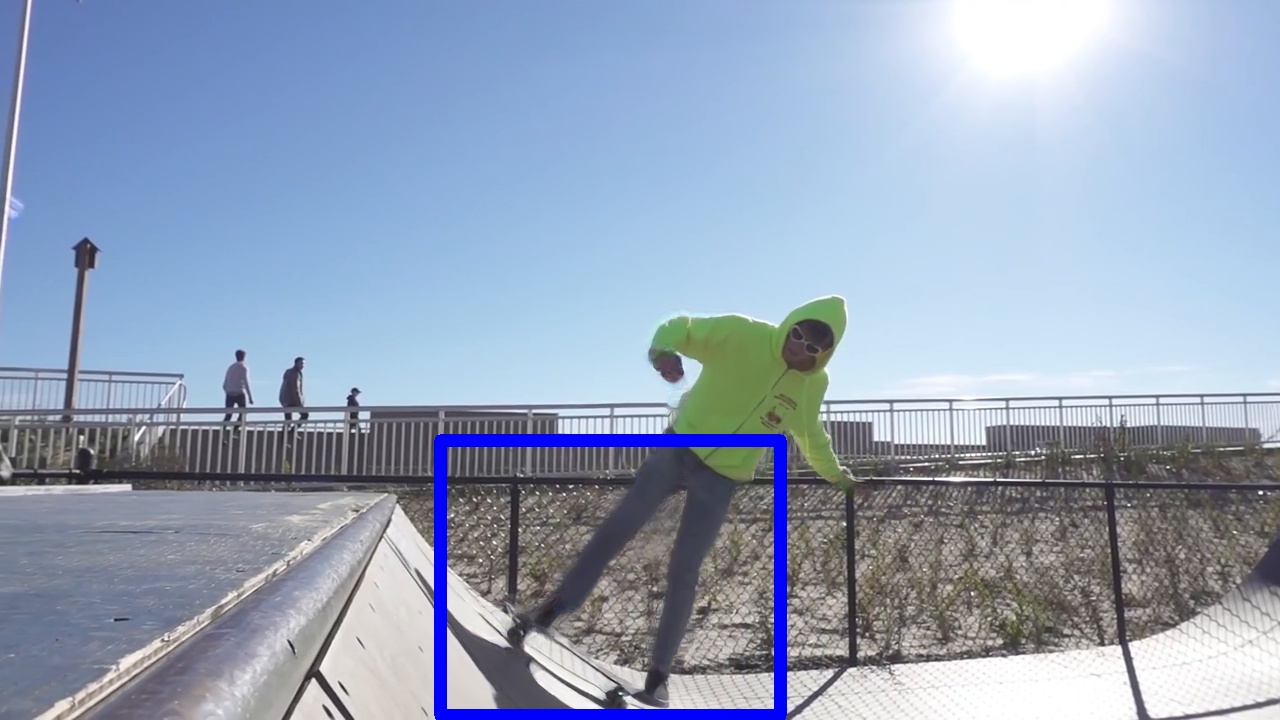}}\hspace{-1pt}
	{\includegraphics[width=0.108\linewidth]{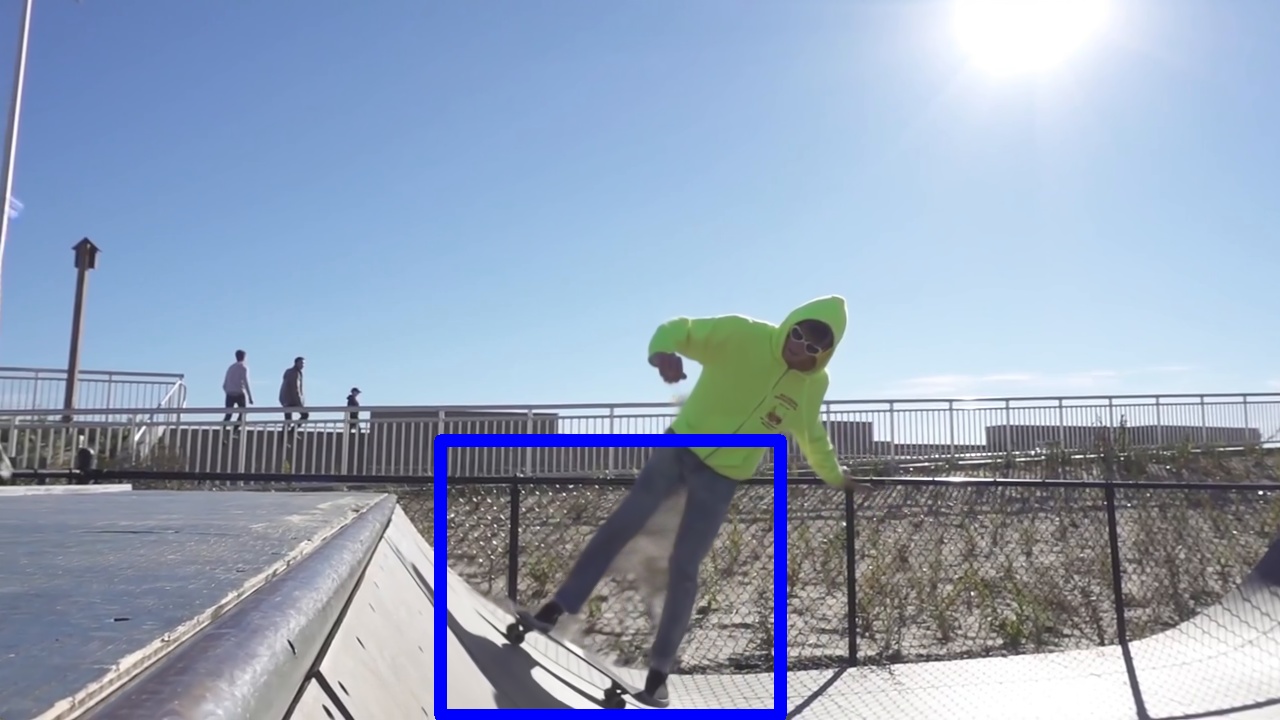}}\hspace{-1pt}
	{\includegraphics[width=0.108\linewidth]{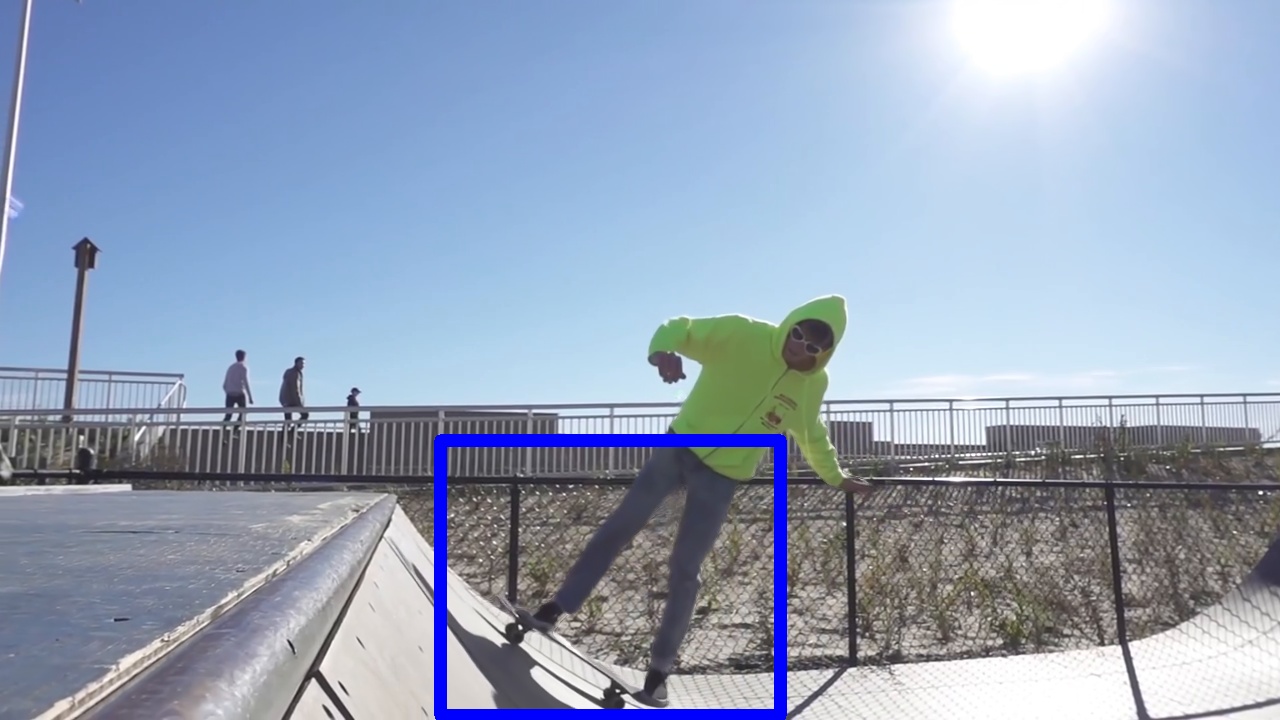}}\newline
	\subfloat[\scriptsize Ground Truth] {\includegraphics[width=0.108\linewidth]{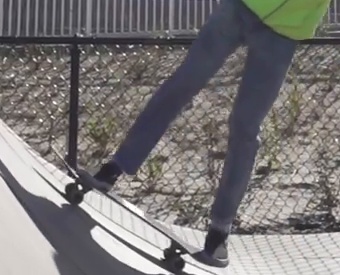}}\hspace{-1pt}
	\subfloat[\scriptsize Overlaid]{\includegraphics[width=0.108\linewidth]{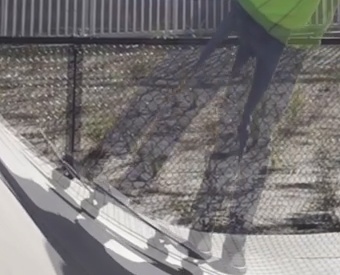}}\hspace{-1pt}
	\subfloat[\scriptsize SepConv~\cite{8237299}] {\includegraphics[width=0.108\linewidth]{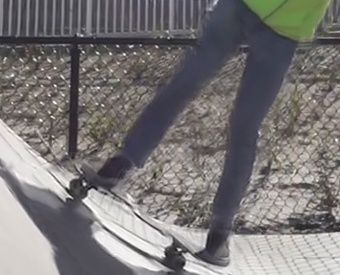}}\hspace{-1pt}
	\subfloat[\scriptsize DAIN~\cite{8954114}] {\includegraphics[width=0.108\linewidth]{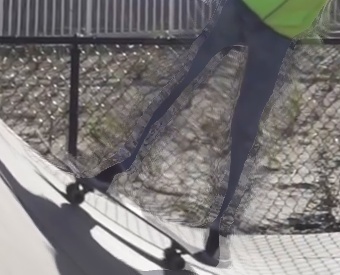}}\hspace{-1pt}
	\subfloat[\scriptsize CAIN~\cite{choi2020cain}] {\includegraphics[width=0.108\linewidth]{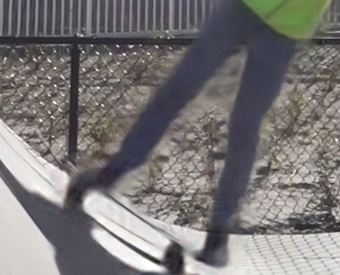}}\hspace{-1pt}
	\subfloat[\scriptsize AdaCoF~\cite{Lee_2020_CVPR}] {\includegraphics[width=0.108\linewidth]{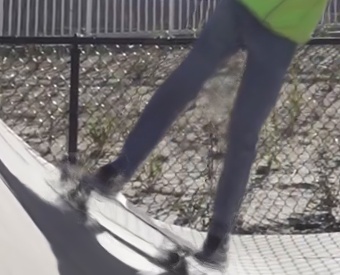}}\hspace{-1pt}
	\subfloat[\scriptsize CDFI~\cite{ding2021cdfi}] {\includegraphics[width=0.108\linewidth]{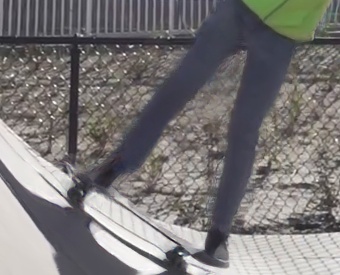}}\hspace{-1pt}
	\subfloat[\scriptsize ABME~\cite{park2021asymmetric}] {\includegraphics[width=0.108\linewidth]{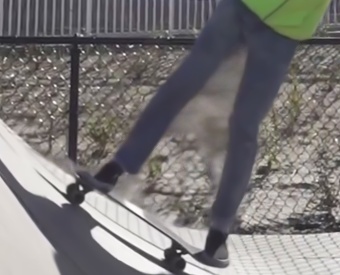}}\hspace{-1pt}
	\subfloat[\scriptsize Ours] {\includegraphics[width=0.108\linewidth]{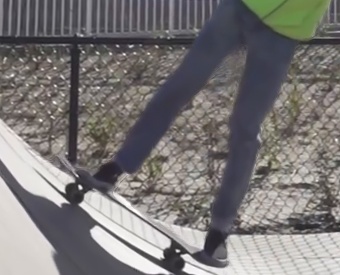}}\newline
	\vspace{-3mm}
	\caption{\textbf{Qualitative comparison of different VFI methods on SNU-FILM (Hard) dataset.} Proposed IFRNet algorithm can synthesize fast moving objects with sharp boundary while maintaining distinct contextual details. Zoom in for best view.}
	\label{fig:6}
	\vspace{-2mm}
\end{figure*}

\begin{table*}
	\renewcommand{\arraystretch}{0.84}
	{\small
		\centering
		\setlength\tabcolsep{3.4pt}
		\begin{tabular}{lcc|ccccccccccccccccccccccccc}
			\toprule
			\multirow{2}[2]{*}{Method} & \multicolumn{2}{c}{Average} & \multicolumn{2}{c}{Mequon} & \multicolumn{2}{c}{Schefflera} & \multicolumn{2}{c}{Urban} & \multicolumn{2}{c}{Teddy} & \multicolumn{2}{c}{Backyard} & \multicolumn{2}{c}{Basketball} & \multicolumn{2}{c}{Dumptruck} & \multicolumn{2}{c}{Evergreen}\\
			\cmidrule(l{5pt}r{5pt}){2-3} \cmidrule(l{5pt}r{5pt}){4-5} \cmidrule(l{5pt}r{5pt}){6-7} \cmidrule(l{5pt}r{5pt}){8-9} \cmidrule(l{5pt}r{5pt}){10-11} \cmidrule(l{5pt}r{5pt}){12-13} \cmidrule(l{5pt}r{5pt}){14-15} \cmidrule(l{5pt}r{5pt}){16-17} \cmidrule(l{5pt}r{5pt}){18-19}
			& IE & NIE & IE & NIE & IE & NIE & IE & NIE & IE & NIE & IE & NIE & IE & NIE & IE & NIE & IE & NIE \\
			\midrule
			SuperSlomo~\cite{8579036} & 5.310 & 0.778 & 2.51 & 0.59 & 3.66 & 0.72 & 2.91 & 0.74 & 5.05 & 0.98 & 9.56 & 0.94 & 5.37 & 0.96 & 6.69 & 0.60 & 6.73 & 0.69  \\
			ToFlow~\cite{xue2019video} & 5.490 & 0.840 & 2.54 & 0.55 & 3.70 & 0.72 & 3.43 & 0.92 & 5.05 & 0.96 & 9.84 & 0.97 & 5.34 & 0.98 & 6.88 & 0.72 & 7.14 & 0.90  \\
			DAIN~\cite{8954114} & 4.856 & 0.713 & 2.38 & 0.58 & 3.28 & 0.60 & 3.32 & 0.69 & 4.65 & 0.86 & 7.88 & 0.87 & 4.73 & 0.85 & 6.36 & 0.59 & 6.25 & 0.66  \\
			FeFlow~\cite{Gui_2020_CVPR} & 4.820 & 0.719 & 2.28 & \textcolor{red}{\bf 0.51} & 3.50 & 0.66 & 2.82 & 0.70 & 4.75 & 0.87 & \textcolor{blue}{\underline{7.62}} & \textcolor{red}{\bf0.84} & 4.74 & 0.86 & 6.07 & 0.64 & 6.78 & 0.67 \\
			AdaCoF~\cite{Lee_2020_CVPR} & 4.751 & 0.730 & 2.41 & 0.60 & 3.10 & 0.59 & 3.48 & 0.84 & 4.84 & 0.92 & 8.68 & 0.90 & 4.13 & 0.84 & 5.77 & \textcolor{blue}{\underline{0.58}} & 5.60 & \textcolor{blue}{\underline{0.57}} \\
			BMBC~\cite{BMBC} & 4.479 & 0.696 & 2.30 & 0.57 & 3.07 & 0.58 & 3.17 & 0.77 & 4.24 & 0.84 & 7.79 & \textcolor{blue}{\underline{0.85}} & \textcolor{red}{\bf 4.08} & \textcolor{blue}{\underline{0.82}} & 5.63 & \textcolor{blue}{\underline{0.58}} & \textcolor{blue}{\underline{5.55}} & \textcolor{red}{\bf 0.56} \\
			SoftSplat~\cite{Niklaus_2020_CVPR} & \textcolor{blue}{\underline{4.223}} & \textcolor{blue}{\underline{0.645}} & \textcolor{red}{\bf 2.06} & \textcolor{blue}{\underline{0.53}} & \textcolor{blue}{\underline{2.80}} & \textcolor{blue}{\underline{0.52}} & \textcolor{blue}{\underline{1.99}} & \textcolor{blue}{\underline{0.52}} & \textcolor{red}{\bf 3.84} & \textcolor{red}{\bf 0.80} & 8.10 & \textcolor{blue}{\underline{0.85}} & \textcolor{blue}{\underline{4.10}} & \textcolor{red}{\bf 0.81} & \textcolor{red}{\bf 5.49} & \textcolor{red}{\bf 0.56} & \textcolor{red}{\bf 5.40} & \textcolor{blue}{\underline{0.57}} \\
			IFRNet large & \textcolor{red}{\bf 4.216} &  \textcolor{red}{\bf 0.644} & \textcolor{blue}{\underline{2.08}} & \textcolor{blue}{\underline{0.53}} & \textcolor{red}{\bf 2.78} & \textcolor{red}{\bf 0.51} & \textcolor{red}{\bf 1.74} & \textcolor{red}{\bf 0.43} & \textcolor{blue}{\underline{3.96}} & \textcolor{blue}{\underline{0.83}} & \textcolor{red}{\bf 7.55} & 0.87 & 4.42 & 0.84 & \textcolor{blue}{\underline{5.56}} & \textcolor{red}{\bf 0.56} & 5.64 & 0.58 \\
			\bottomrule
		\end{tabular}
		\vspace{-2mm}
		\caption{\textbf{Evaluation results on the Middlebury benchmark.} For each item, the best result is \textcolor{red}{\textbf{boldfaced}}, and the second best is \textcolor{blue}{\underline{underlined}}.}
		\label{tab:2}}
	\vspace{-3mm}
\end{table*}

\noindent\textbf{Quantitative Evaluation.}
Table~\ref{tab:1} and Table~\ref{tab:2} summarize quantitative results on diverse benchmarks. On Vimeo90K and UCF101 test datasets, IFRNet large achieves the best results on both PSNR and SSIM metrics. A recent method ABME~\cite{park2021asymmetric} also gets similar accuracy. However, our model runs \textbf{11.5} $\times$ faster with similar amount of parameters due to the efficiency of single encoder-decoder based architecture. Our large model also obtains leading results on the Easy, Medium and Hard parts of SNU-FILM datasets, while only falls behind ABME on the Extreme part. We attribute the reason to be that the bilateral cost volume constructed by ABME is good at estimating large displacement motion. In Table~\ref{tab:2}, IFRNet large achieves top-performing VFI accuracy in most of the eight Middlebury test sequences, and outperforms the previous state-of-the-art SoftSplat~\cite{Niklaus_2020_CVPR} on both average IE and NIE metrics. Although the improvement is limited, our approach runs \textbf{2.5} $\times$ faster than SoftSplat which takes cascaded VFI architecture. For FLOPs in convolution layers, IFRNet large also consumes significantly less computation than other VFI architectures.

In regard to real-time and lightweight VFI approaches, IFRNet yields about 0.2 dB better result than RIFE~\cite{huang2021rife} on Vimeo90K, and the margin is more distinct on large motion cases in SNU-FILM dataset. It is worth noting that IFRNet only contains \textbf{half} parameters to achieve better results than RIFE thanks to the superiority of joint refinement of intermediate flow and context feature. Compared with CDFI full~\cite{ding2021cdfi}, IFRNet has the same 5M parameters, while achieving \textbf{0.63} dB higher PSNR on Vimeo90K with \textbf{15.2} $\times$ faster inference speed. Moreover, IFRNet small can further improve speed by \textbf{31}\% and reduce parameters and computation complexity by \textbf{44}\% than IFRNet while with only slight frame interpolation accuracy decrease.

\vspace{1mm}
\noindent\textbf{Qualitative Evaluation.}
Figure~\ref{fig:6} visually compares well-behaved VFI methods on SNU-FILM (Hard) dataset which contains large and complex motion scenes. It can be seen that kernel-based~\cite{8237299,Lee_2020_CVPR,ding2021cdfi} and hallucination-based~\cite{choi2020cain} methods fail to synthesize sharp motion boundary, containing ghost and blur artifacts. Compared with flow-based algorithms~\cite{8954114,park2021asymmetric}, our approach can generate texture details faithfully thanks to the powerfulness of gradually refined intermediate feature. In short, IFRNet can synthesize pleasing target frame with more comfortable visual experience. More qualitative results can be found in our supplementary.

\subsection{Ablation Study}
To verify the effectiveness of proposed approaches, we carry out ablation study in terms of network architecture and loss function on Vimeo90K and SNU-FILM Hard datasets.

\begin{table}[t]
	\renewcommand{\arraystretch}{0.8}
	{\small
		\centering
		\begin{tabular}{C{1.5cm}C{1.5cm}C{1.5cm}C{1.5cm}}
			\toprule
			\multicolumn{2}{c}{Architecture} & Vimeo90K & Hard \\
			\cmidrule(lr){1-2} \cmidrule(lr){3-3} \cmidrule(lr){4-4}
			\; IF & R & PSNR & PSNR \\
			\midrule
			\; \xmark & \xmark & 34.83 & 29.96 \\
			\; \cmark & \xmark & 35.22 & 30.22 \\
			\; \xmark & \cmark & 35.11 & 30.06 \\
			\; \cmark & \cmark & \textbf{35.51} & \textbf{30.27} \\
			\bottomrule
		\end{tabular}
		\vspace{-2mm}
		\caption{\textbf{Ablation study on different architecture variants.} `IF' means intermediate feature $\hat{\phi}_{t}^{k}$ and `R' stands for residual $R$.}
		\label{tab:3}}
	\vspace{-3mm}
\end{table}

\begin{figure}[t]
	\centering
	\includegraphics[width=0.98\columnwidth]{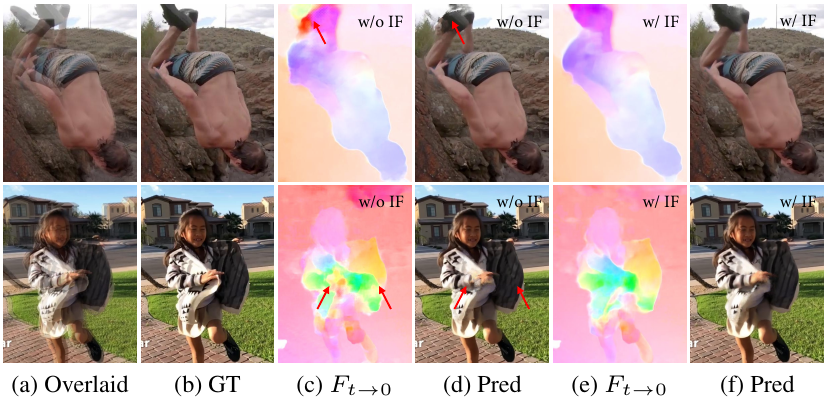}
	\vspace{-2mm}
	\caption{\textbf{Visual comparison of intermediate flow and predicted frame of IFRNet w/o and w/ intermediate feature.}}
	\label{fig:7}
	\vspace{-4mm}
\end{figure}

\noindent\textbf{Intermediate Feature.}
To ablate the effectiveness of intermediate feature $\hat{\phi}_{t}^{k}$ in IFRNet, we build a model by removing $\hat{\phi}_{t}^{k}$ from the input and output of multiple decoders, while keeping feature channels of middle parts of decoders unchanged. Also, we selectively remove residual R in Eq.~\ref{eq:4} to isolate the improvement from intermediate flow and residual. We train them with only the reconstruction loss $\mathcal{L}_{r}$ under the same learning schedule as before. As listed in Table~\ref{tab:3}, from the first two rows, we can observe that intermediate feature can provide reference anchor information to promote intermediate flow estimation. Figure~\ref{fig:7} also presents some visual examples to confirm the conclusion. Compared with the last and the second rows in Table~\ref{tab:3}, it demonstrates that gradually refined intermediate feature, containing global context information, can compensate better scene details. Conclusively, residual compensation from the intermediate context feature is necessary for IFRNet to achieve advanced VFI performance, since intermediate flow prediction is substantively unreliable. Overall, the two-fold benefits from intermediate feature greatly improves VFI accuracy of IFRNet with relatively small additional cost.

\begin{figure}[t]
	\centering
	\includegraphics[width=0.94\columnwidth]{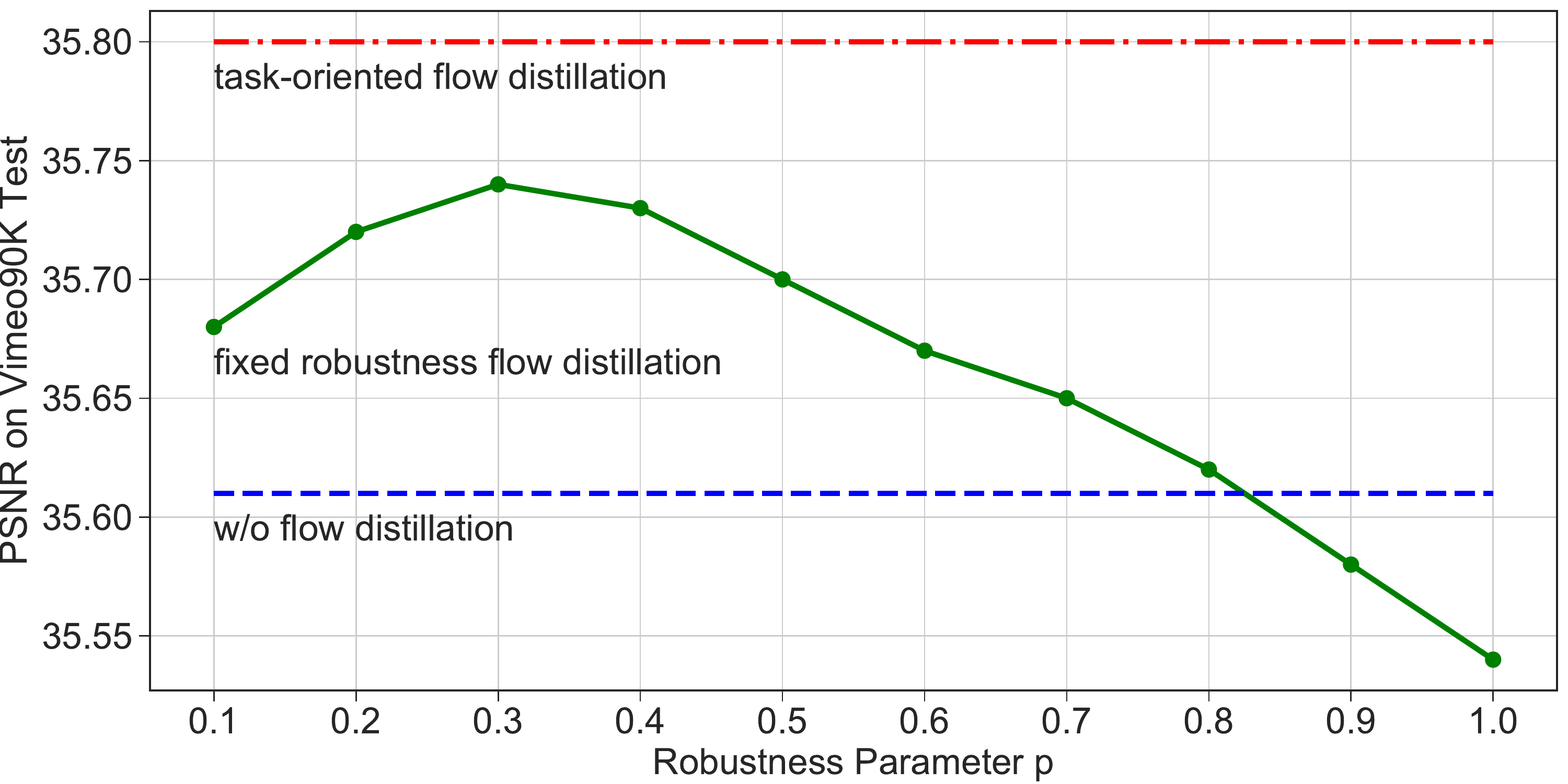}
	\vspace{-2mm}
	\caption{\textbf{Ablation study on different flow distillation losses.}}
	\label{fig:8}
	\vspace{-4mm}
\end{figure}

\vspace{1mm}
\noindent\textbf{Task-Oriented Flow Distillation.}
Table~\ref{tab:4} compares VFI accuracy under different combinations of proposed loss functions quantitatively. It can be seen that adding task-oriented flow distillation loss $\mathcal{L}_{d}$ consistently improves PSNR of 0.2 dB on Vimeo90K. To verify the superiority of its task adaptive ability, we also perform flow distillation with generalized Charbonnier loss under different robustness shown in Figure~\ref{fig:5}, whose results are summarized in Figure~\ref{fig:8}. It turns out that robustness parameter $p=0.3$ achieves best VFI accuracy in the fixed robustness setting. On the other hand, flow distillation can damage frame quality when $p$ approaches to 1.0 due to the harmful knowledge in pseudo label. In a word, proposed task-oriented approach achieves the best accuracy thanks to its spatial adaptive ability for adjusting robustness loss during flow distillation.

\begin{table}
	\renewcommand{\arraystretch}{0.8}
	{\small
		\centering
		\begin{tabular}{C{0.9cm}C{0.9cm}C{0.9cm}C{1.5cm}C{1.5cm}}
			\toprule
			\multicolumn{3}{c}{Loss Function} & Vimeo90K & Hard \\
			\cmidrule(lr){1-3} \cmidrule(lr){4-4} \cmidrule(lr){5-5}
			\; $\mathcal{L}_{r}$ & $\mathcal{L}_{d}$ & $\mathcal{L}_{g}$ & PSNR & PSNR \\
			\midrule
			\; \cmark & \xmark & \xmark & 35.51 & 30.27 \\
			\; \cmark & \cmark & \xmark & 35.72 & 30.38 \\
			\; \cmark & \xmark & \cmark & 35.61 & 30.30 \\
			\; \cmark & \cmark & \cmark & \textbf{35.80} & \textbf{30.41} \\
			\bottomrule
		\end{tabular}
		\vspace{-2mm}
		\caption{\textbf{Ablation study on different loss functions.}}
		\label{tab:4}}
		\vspace{-3mm}
\end{table}

\begin{figure}[t]
	\centering
	\includegraphics[width=0.98\columnwidth]{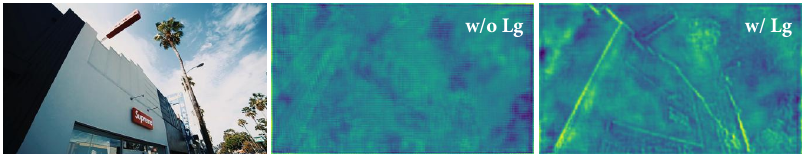}
	\vspace{-2mm}
	\caption{\textbf{Visual comparison of mean feature map of intermediate feature $\hat{\phi}_{t}^{1}$ w/o and w/ $\mathcal{L}_{g}$.} Leftmost is the ground truth.}
	\label{fig:9}
	\vspace{-4mm}
\end{figure}

\vspace{1mm}
\noindent\textbf{Feature Space Geometry Consistency.}
As shown in Table~\ref{tab:4}, adding proposed feature space geometry consistency loss $\mathcal{L}_{g}$ based on above contributions, we can obtain a further improvement, that confirms the complementary effect of $\mathcal{L}_{g}$ in regard to $\mathcal{L}_{d}$. Figure~\ref{fig:9} visually compares mean feature maps of intermediate feature $\hat{\phi}_{t}^{1}$ w/o and w/ $\mathcal{L}_{g}$. It shows that $\mathcal{L}_{g}$ can regularize the reconstructed intermediate feature to keep better geometry layout in multi-scale feature space, resulting in better VFI performance.

\section{Conclusion}
In this paper, we have devised an efficient deep architecture, termed IFRNet, for video frame interpolation, without any cascaded synthesis or refinement module. It gradually refines intermediate flow together with a powerful intermediate feature, that can not only boost intermediate flow estimation to synthesize sharp motion boundary but also provide global context representation to generate vivid motion details. Moreover, we have presented task-oriented flow distillation loss and feature space geometry consistency loss to fully release its potential. Experiments on various benchmarks demonstrate the state-of-the-art performance and fast inference speed of proposed approaches. We expect proposed single encoder-decoder joint refinement based IFRNet to be a useful component for many frame rate up-conversion and intermediate view synthesis systems.

\newpage
\twocolumn[{
	\renewcommand\twocolumn[1][]{#1}%
	\begin{center}
		\textbf{\Large IFRNet: Intermediate Feature Refine Network for Efficient Frame Interpolation Supplementary Material\\}
		\vspace{8mm}
		{\large Lingtong Kong\textsuperscript{\rm 1}$^{*}$,
			Boyuan Jiang\textsuperscript{\rm 2}$^{*}$,
			Donghao Luo\textsuperscript{\rm 2},
			Wenqing Chu\textsuperscript{\rm 2},
			Xiaoming Huang\textsuperscript{\rm 2}, \\
			Ying Tai\textsuperscript{\rm 2},
			Chengjie Wang\textsuperscript{\rm 2},
			Jie Yang\textsuperscript{\rm 1}$^{\dagger}$ \\
			\textsuperscript{\rm 1}Institute of Image Processing and Pattern Recognition, Shanghai Jiao Tong University, \\
			\textsuperscript{\rm 2}Youtu Lab, Tencent \\}
		{\tt\small \{ltkong, jieyang\}@sjtu.edu.cn } \\
		{\tt\small \{byronjiang, michaelluo, wenqingchu, skyhuang, yingtai, jasoncjwang\}@tencent.com} \\
	\end{center}
	\begin{center}
		\centering
		\setlength{\tabcolsep}{0.4mm}
		\captionsetup{type=figure}
		\begin{tabular}{cccc}
			\animategraphics[width=0.245\linewidth, autoplay, poster=0, palindrome, final, nomouse, method=widget]{8}{figures_supp/fig1/fig1_1/}{00}{08}
			&
			\animategraphics[width=0.245\linewidth, autoplay, poster=0, palindrome, final, nomouse, method=widget]{8}{figures_supp/fig1/fig1_2/}{00}{08}
			&
			\animategraphics[width=0.245\linewidth, autoplay, poster=0, palindrome, final, nomouse, method=widget]{8}{figures_supp/fig1/fig1_3/}{00}{08}
			&
			\animategraphics[width=0.245\linewidth, autoplay, poster=0, palindrome, final, nomouse, method=widget]{8}{figures_supp/fig1/fig1_4/}{00}{08} \\
			\animategraphics[width=0.245\linewidth, autoplay, poster=0, palindrome, final, nomouse, method=widget]{8}{figures_supp/fig1/fig1_5/}{00}{08}
			&
			\animategraphics[width=0.245\linewidth, autoplay, poster=0, palindrome, final, nomouse, method=widget]{8}{figures_supp/fig1/fig1_6/}{00}{08}
			&
			\animategraphics[width=0.245\linewidth, autoplay, poster=0, palindrome, final, nomouse, method=widget]{8}{figures_supp/fig1/fig1_7/}{00}{08}
			&
			\animategraphics[width=0.245\linewidth, autoplay, poster=0, palindrome, final, nomouse, method=widget]{8}{figures_supp/fig1/fig1_8/}{00}{08} \\
		\end{tabular}
		\vspace{-4mm}
		\caption{\textbf{Qualitative results of IFRNet for 8$\times$ interpolation on GoPro~\cite{Nah_2017_CVPR} and Adobe240~\cite{8099516} test datasets.} Please watch the video with Adobe Reader. Each video has 9 frames where the first and the last frames are input, and the middle 7 frames are predicted by IFRNet. }
		\label{fig:10}
	\end{center}
}]

\let\thefootnote\relax\footnotetext{$*$ Equal contribution. This work was done when Lingtong Kong was an intern at Tencent Youtu Lab. Code is available at \url{https://github.com/ltkong218/IFRNet}.}
\let\thefootnote\relax\footnotetext{$\dagger$ Corresponding author: Jie Yang (jieyang@sjtu.edu.cn). This research is partly supported by NSFC, China (No: 61876107, U1803261).}

\thispagestyle{empty}

In the supplementary, we first present multi-frame interpolation experiments of IFRNet. Second, qualitative video comparisions with other advanced VFI approaches are displayed. Third, we depict structure details of IFRNet and its variants. Fourth, we provide more visual examples and analysis of middle components for better understanding the workflow of IFRNet. Finally, we show the screenshot of VFI results on the Middlebury benchmark. Please note that the numbering within this supplementary has manually been adjusted to continue the ones in our main paper.

\section{Multi-Frame Interpolation}
Different from other multi-frame interpolation methods which scales optical flow~\cite{8579036,8954114} or interpolates middle frames recursively~\cite{choi2020cain,Lee_2020_CVPR}, IFRNet can predict multiple intermediate frames by proposed one-channel temporal encoding mask $T$, which is one of the input of the coarsest decoder $\mathcal{D}^{4}$. The temporal encoding is a conditional input signal whose values are all the same and set to $t$, where $t\in\{1/8, 2/8, \ldots, 7/8\}$ in 8$\times$ interpolation setting. Also, proposed task-oriented flow distillation loss and feature space geometry consistency loss still work for any intermediate time instance $t$. To evaluate IFRNet for 8$\times$ interpolation, we use the train/test split of FLAVR~\cite{kalluri2021flavr}, where we train IFRNet on GoPro~\cite{Nah_2017_CVPR} training set with the same learning schedule and loss functions as our main paper. Then we test the pre-trained model on GoPro testing and Adobe240~\cite{8099516} datasets whose results are listed in Table~\ref{tab:A}.

\begin{table}[t]
	\renewcommand{\arraystretch}{0.8}
	\vspace{2mm}
	{\small
		\centering
		\setlength\tabcolsep{2.0mm}
		\begin{tabular}{cccccc}
			\toprule
			\multirow{2}[2]{*}{Method} & \multicolumn{2}{c}{GoPro~\cite{Nah_2017_CVPR}} & \multicolumn{2}{c}{Adobe240~\cite{8099516}} & Time \\
			\cmidrule(lr){2-3} \cmidrule(lr){4-5} & PSNR & SSIM & PSNR & SSIM & (s) \\
			\midrule
			DVF~\cite{8237740} & 21.94 & 0.776 & 28.23 & 0.896 & 0.87 \\
			SuperSloMo~\cite{8579036} & 28.52 & 0.891 & 30.66 & 0.931  & 0.44 \\
			DAIN~\cite{8954114} & 29.00 & 0.910 & 29.50 & 0.910 & 4.10 \\
			IFRNet (Ours) & \textbf{29.84} & \textbf{0.920} & \textbf{31.93} & \textbf{0.943} & \textbf{0.16} \\
			\bottomrule
		\end{tabular}
		\vspace{-2mm}
		\caption{\textbf{Quantitative comparison for 8$\times$ interpolation.}}
		\label{tab:A}}
	\vspace{-6mm}
\end{table}

\begin{figure*}[t]
	{\small
		\centering
		\setlength{\tabcolsep}{0.4mm}
		\begin{tabular}{ccc}
			\animategraphics[width=0.33\linewidth, autoplay, poster=0, palindrome, final, nomouse, method=widget]{2}{figures_supp/fig2/GT/video_1/}{00}{04}
			&
			\animategraphics[width=0.33\linewidth, autoplay, poster=0, palindrome, final, nomouse, method=widget]{2}{figures_supp/fig2/DAIN/video_1/}{00}{04}
			&
			\animategraphics[width=0.33\linewidth, autoplay, poster=0, palindrome, final, nomouse, method=widget]{2}{figures_supp/fig2/CAIN/video_1/}{00}{04} \vspace{-0.8mm} \\
			Ground Truth & DAIN~\cite{8954114} & CAIN~\cite{choi2020cain} \\
			\animategraphics[width=0.33\linewidth, autoplay, poster=0, palindrome, final, nomouse, method=widget]{2}{figures_supp/fig2/AdaCoF/video_1/}{00}{04}
			&
			\animategraphics[width=0.33\linewidth, autoplay, poster=0, palindrome, final, nomouse, method=widget]{2}{figures_supp/fig2/ABME/video_1/}{00}{04}
			&
			\animategraphics[width=0.33\linewidth, autoplay, poster=0, palindrome, final, nomouse, method=widget]{2}{figures_supp/fig2/IFRNet/video_1/}{00}{04} \vspace{-0.8mm} \\
			AdaCoF~\cite{Lee_2020_CVPR} & ABME~\cite{park2021asymmetric} & IFRNet (Ours) \\
			
			\animategraphics[width=0.33\linewidth, autoplay, poster=0, palindrome, final, nomouse, method=widget]{2}{figures_supp/fig2/GT/video_2/}{00}{04}
			&
			\animategraphics[width=0.33\linewidth, autoplay, poster=0, palindrome, final, nomouse, method=widget]{2}{figures_supp/fig2/DAIN/video_2/}{00}{04}
			&
			\animategraphics[width=0.33\linewidth, autoplay, poster=0, palindrome, final, nomouse, method=widget]{2}{figures_supp/fig2/CAIN/video_2/}{00}{04} \vspace{-0.8mm} \\
			Ground Truth & DAIN~\cite{8954114} & CAIN~\cite{choi2020cain} \\
			\animategraphics[width=0.33\linewidth, autoplay, poster=0, palindrome, final, nomouse, method=widget]{2}{figures_supp/fig2/AdaCoF/video_2/}{00}{04}
			&
			\animategraphics[width=0.33\linewidth, autoplay, poster=0, palindrome, final, nomouse, method=widget]{2}{figures_supp/fig2/ABME/video_2/}{00}{04}
			&
			\animategraphics[width=0.33\linewidth, autoplay, poster=0, palindrome, final, nomouse, method=widget]{2}{figures_supp/fig2/IFRNet/video_2/}{00}{04} \vspace{-0.8mm} \\
			AdaCoF~\cite{Lee_2020_CVPR} & ABME~\cite{park2021asymmetric} & IFRNet (Ours) \\
			
			\animategraphics[width=0.33\linewidth, autoplay, poster=0, palindrome, final, nomouse, method=widget]{2}{figures_supp/fig2/GT/video_3/}{00}{04}
			&
			\animategraphics[width=0.33\linewidth, autoplay, poster=0, palindrome, final, nomouse, method=widget]{2}{figures_supp/fig2/DAIN/video_3/}{00}{04}
			&
			\animategraphics[width=0.33\linewidth, autoplay, poster=0, palindrome, final, nomouse, method=widget]{2}{figures_supp/fig2/CAIN/video_3/}{00}{04} \vspace{-0.8mm} \\
			Ground Truth & DAIN~\cite{8954114} & CAIN~\cite{choi2020cain} \\
			\animategraphics[width=0.33\linewidth, autoplay, poster=0, palindrome, final, nomouse, method=widget]{2}{figures_supp/fig2/AdaCoF/video_3/}{00}{04}
			&
			\animategraphics[width=0.33\linewidth, autoplay, poster=0, palindrome, final, nomouse, method=widget]{2}{figures_supp/fig2/ABME/video_3/}{00}{04}
			&
			\animategraphics[width=0.33\linewidth, autoplay, poster=0, palindrome, final, nomouse, method=widget]{2}{figures_supp/fig2/IFRNet/video_3/}{00}{04} \vspace{-0.8mm} \\
			AdaCoF~\cite{Lee_2020_CVPR} & ABME~\cite{park2021asymmetric} & IFRNet (Ours) \\
		\end{tabular}
	}
	\vspace{-2mm}
	\caption{\textbf{Video comparison on SNU-FILM~\cite{choi2020cain} dataset}. Please watch the video with Adobe Reader and zoom in for best view.}
	\label{fig:11}
	\vspace{-4mm}
\end{figure*}

IFRNet outperforms all of the other SOTA methods with $2$ input frames on both GoPro and Adobe240 datasets in both PSNR and SSIM metrics. For example, IFRNet achieves \textbf{0.84} dB better results than DAIN~\cite{8954114} on GoPro and exceeds SuperSloMo~\cite{8579036} by \textbf{1.27} dB on Adobe240. Thanks to the modularity character of IFRNet, the encoder only needs a single forward pass, while the decoders infer $7$ times with different temporal embedding to convert videos from $30$ fps into $240$ fps. Therefore, the speed advantage of IFRNet is still or even more obvious than other approaches. Figure~\ref{fig:10} gives some qualitative results of IFRNet for 8$\times$ interpolation, demonstrating its superior ability for frame rate up-conversion and slow motion generation.

\section{Video Comparison}
In this part, we qualitatively compare interpolated videos by proposed IFRNet against other open source VFI methods on SNU-FILM~\cite{choi2020cain} dataset, whose results are shown in Figure~\ref{fig:11}. As can be seen, our approach can generate motion boundary and texture details faithfully thanks to the powerfulness of gradually refined intermediate feature.

\section{Network Architecture}
In this section, we present the structure details of five sub-networks of IFRNet, \textit{i.e.}, pyramid encoder $\mathcal{E}$ and coarse-to-fine decoders $\mathcal{D}^{4}, \mathcal{D}^{3}, \mathcal{D}^{2}, \mathcal{D}^{1}$. In each following figure, arguments of `Conv' and `Deconv' from left to right are input channels, output channels, kernel size, stride and padding, respectively. Dimensions of input and output tensors from left to right stand for feature channels, height and width, separately. A PReLU~\cite{7410480} follows each `Conv' layer, while there is no activation after each `Deconv' layer. In practice, the intermediate flow fields are estimated in a residual manner, which is not reflected in the figures to emphasize the primary network structure. We take input frames with spatial size of 640$\times$480 as example.

\begin{figure}[h]
	\centering
	\includegraphics[width=0.54\columnwidth]{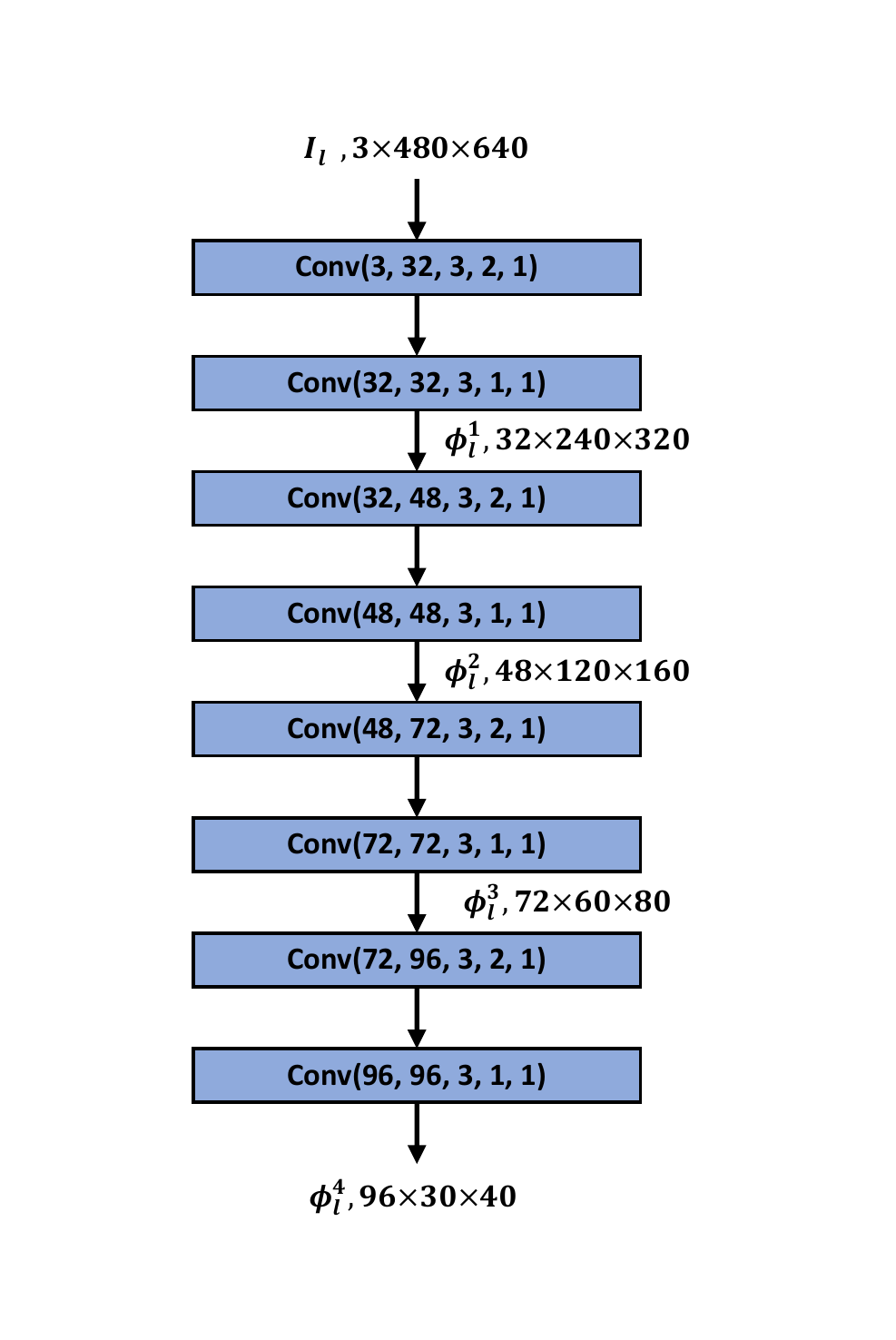}
	\vspace{-5mm}
	\caption{\textbf{Details of the pyramid encoder $\mathcal{E}$.} The two input frames $I_{l}, l \in \{0, 1\}$ are encoded by the same
		Siamese network.}
	\label{fig:12}
	\vspace{-2mm}
\end{figure}

\begin{figure}[h]
	\centering
	\includegraphics[width=0.54\columnwidth]{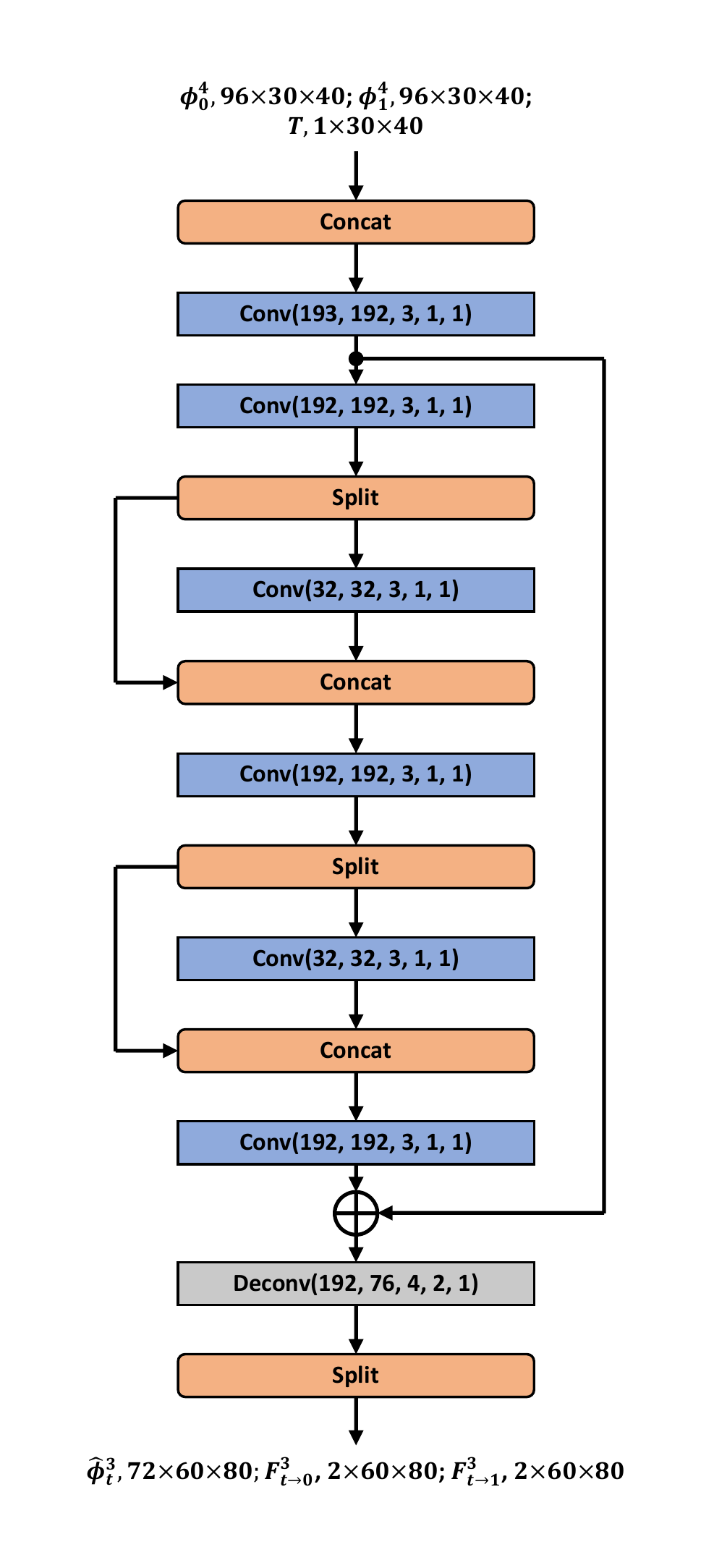}
	\vspace{-5mm}
	\caption{\textbf{Details of the bottom decoder $\mathcal{D}^{4}$.}}
	\label{fig:13}
	\vspace{-2mm}
\end{figure}

\begin{figure}[h]
	\centering
	\includegraphics[width=0.54\columnwidth]{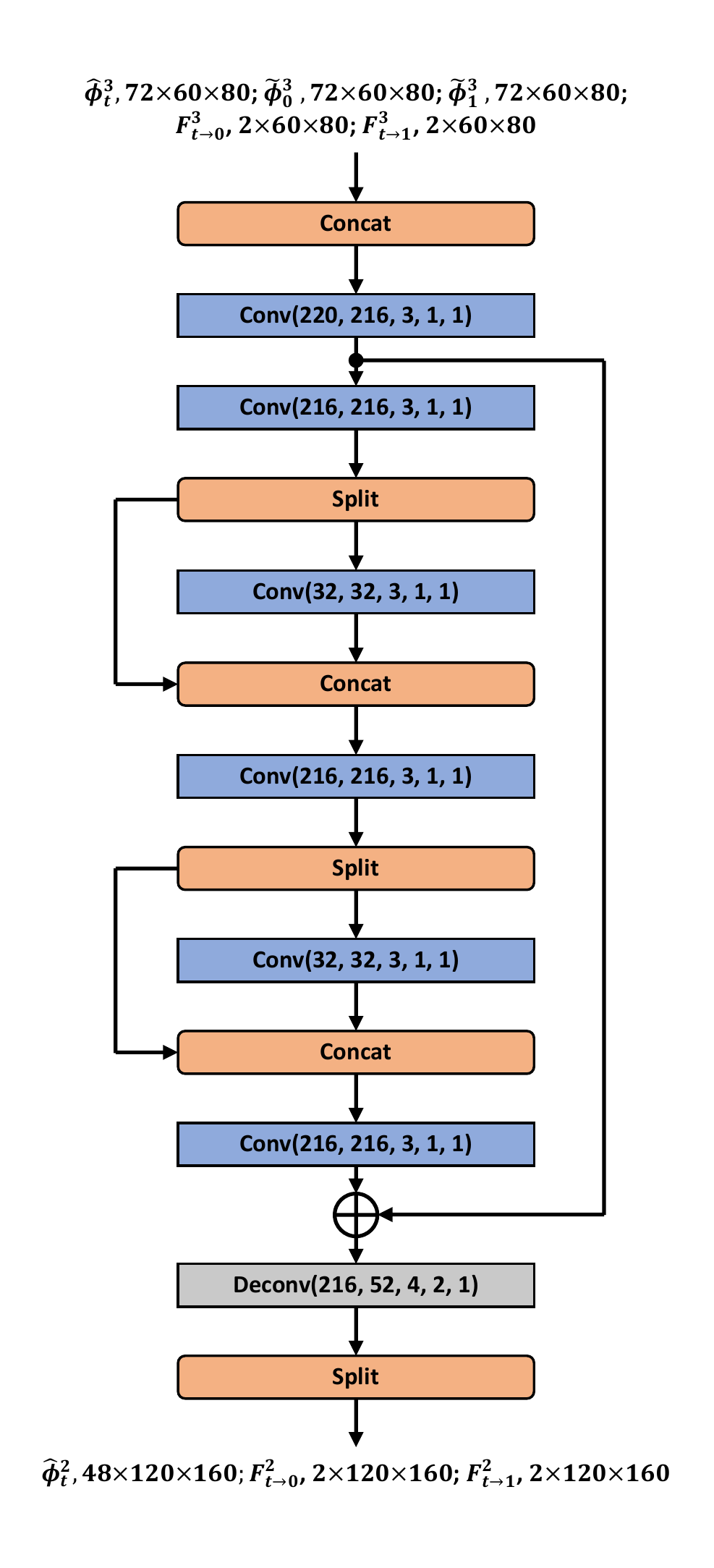}
	\vspace{-5mm}
	\caption{\textbf{Details of the middle decoder $\mathcal{D}^{3}$.}}
	\label{fig:14}
	\vspace{-2mm}
\end{figure}

\begin{figure}[h]
	\centering
	\includegraphics[width=0.54\columnwidth]{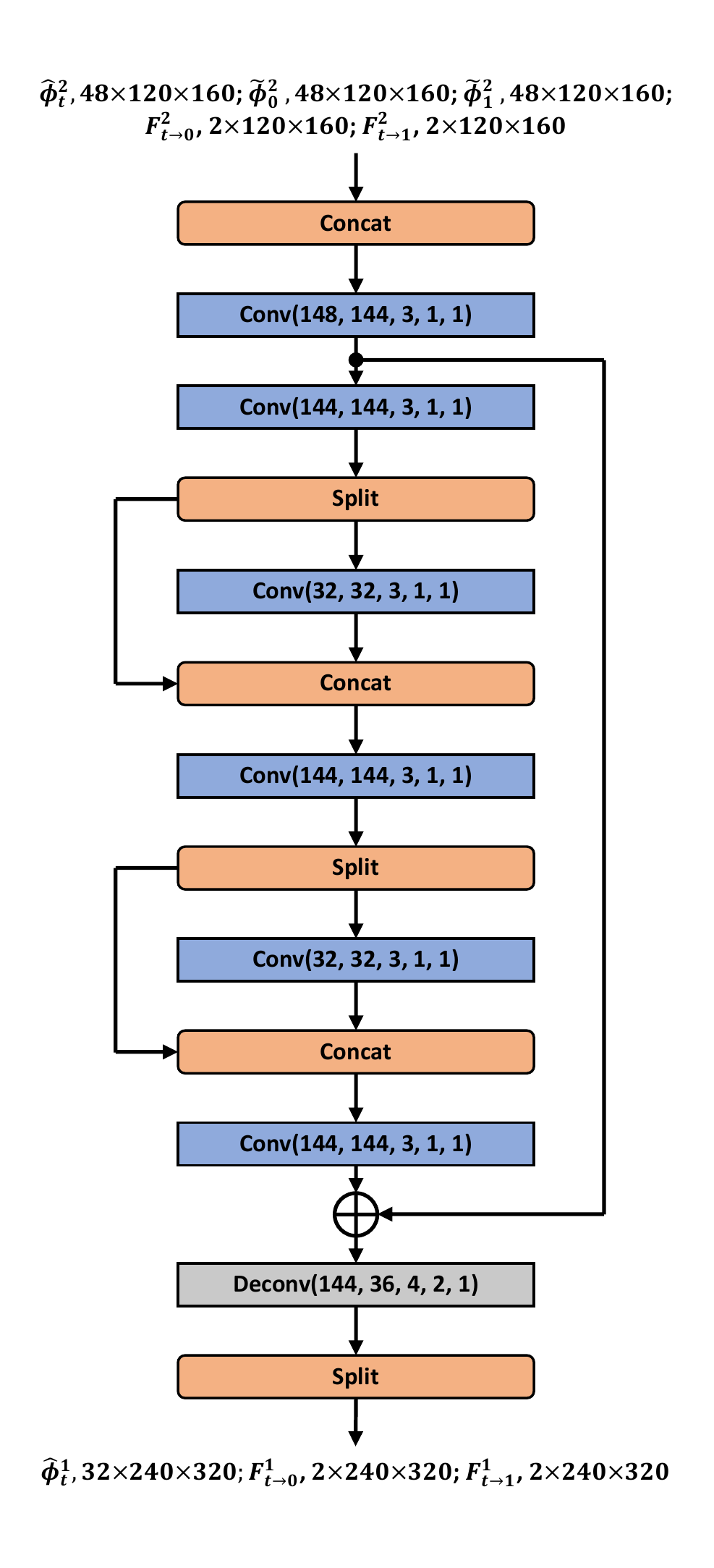}
	\vspace{-5mm}
	\caption{\textbf{Details of the middle decoder $\mathcal{D}^{2}$.}}
	\label{fig:15}
	\vspace{-2mm}
\end{figure}

\begin{figure}[h]
	\centering
	\includegraphics[width=0.54\columnwidth]{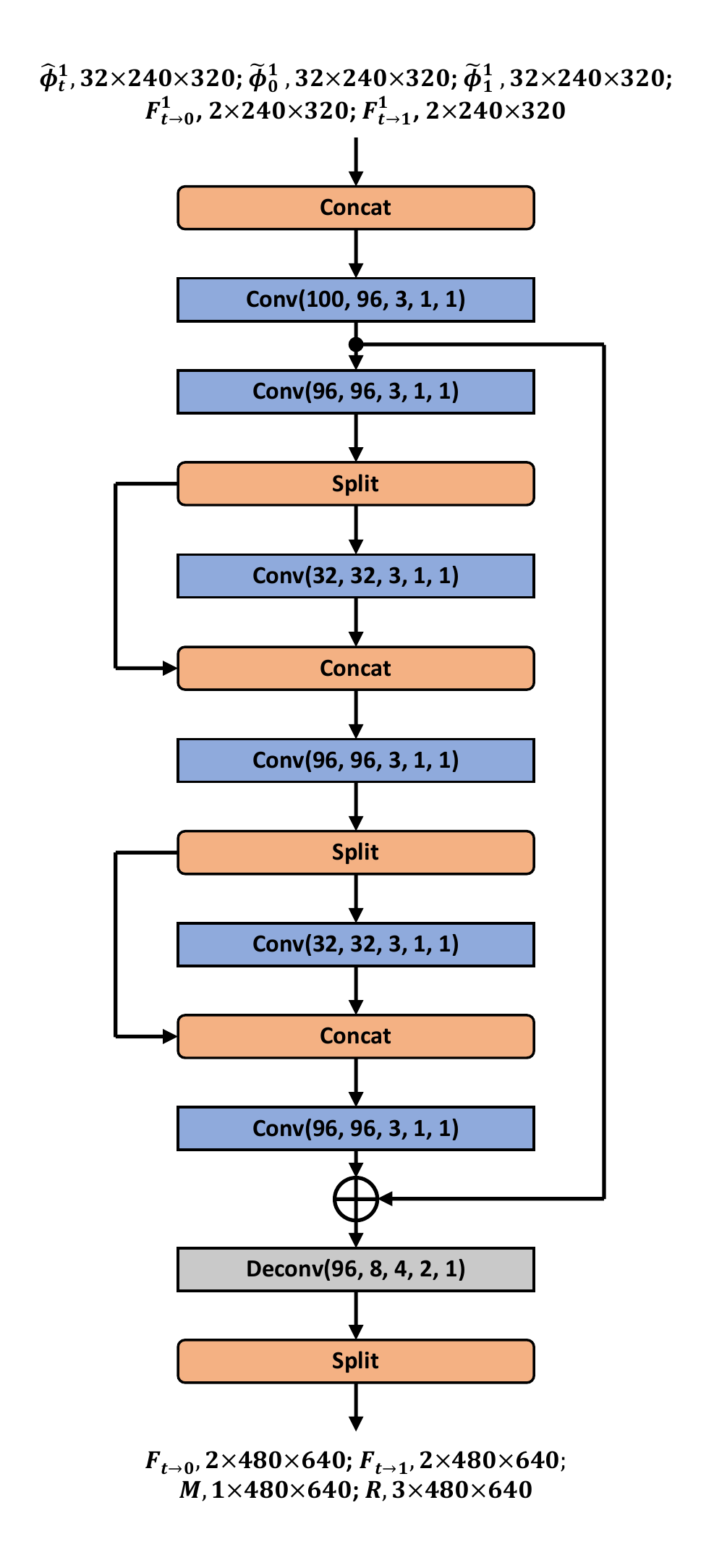}
	\vspace{-5mm}
	\caption{\textbf{Details of the top decoder $\mathcal{D}^{1}$.}}
	\label{fig:16}
	\vspace{-2mm}
\end{figure}

As for IFRNet large and IFRNet small, feature channels from the first to the fourth pyramid levels are set to 64, 96, 144, 192 and 24, 36, 54, 72, respectively. Correspondingly, channel numbers in multiple decoders are adjusted. Also, feature channels of the third and the fifth convolution layers in coarse-to-fine decoders of IFRNet large and IFRNet small are set to 64 and 24, separately.

\section{Visualization and Discussion}
\begin{figure}[h]
	\centering
	{\includegraphics[width=0.242\linewidth]{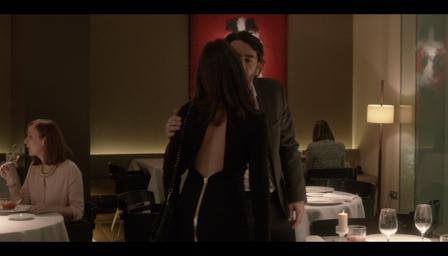}}\hspace{-1pt}
	{\includegraphics[width=0.242\linewidth]{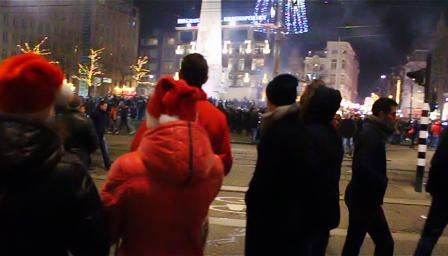}}\hspace{-1pt}
	{\includegraphics[width=0.242\linewidth]{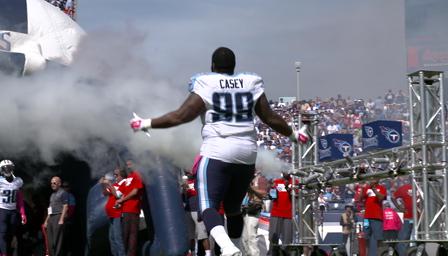}}\hspace{-1pt}
	{\includegraphics[width=0.242\linewidth]{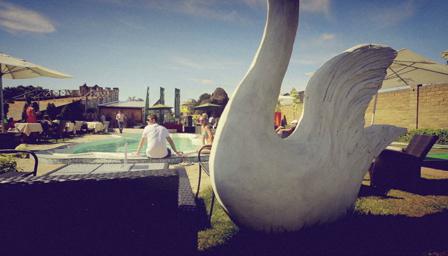}}\newline
	{\includegraphics[width=0.242\linewidth]{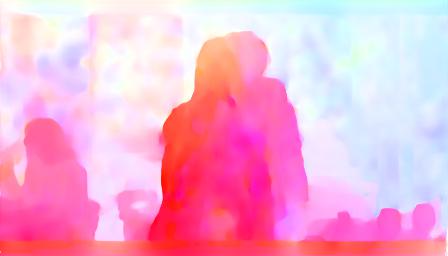}}\hspace{-1pt}
	{\includegraphics[width=0.242\linewidth]{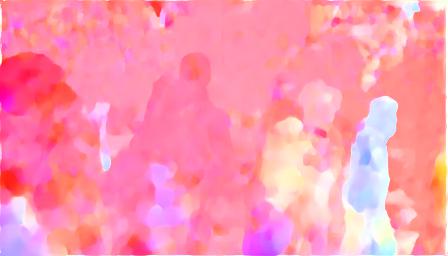}}\hspace{-1pt}
	{\includegraphics[width=0.242\linewidth]{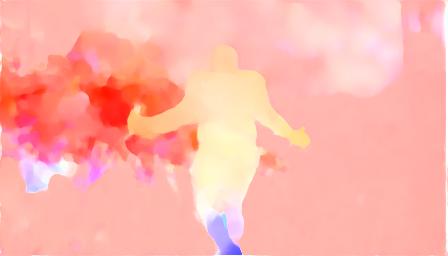}}\hspace{-1pt}
	{\includegraphics[width=0.242\linewidth]{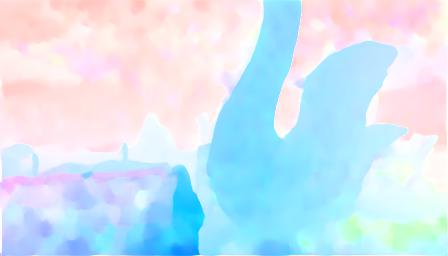}}\newline
	{\includegraphics[width=0.242\linewidth]{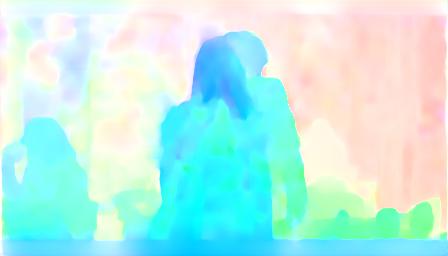}}\hspace{-1pt}
	{\includegraphics[width=0.242\linewidth]{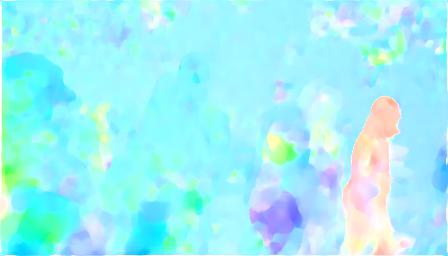}}\hspace{-1pt}
	{\includegraphics[width=0.242\linewidth]{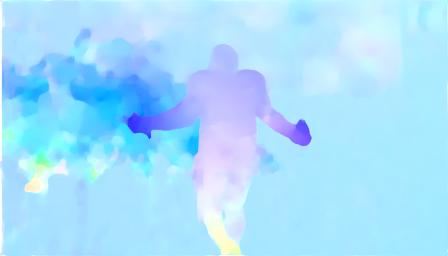}}\hspace{-1pt}
	{\includegraphics[width=0.242\linewidth]{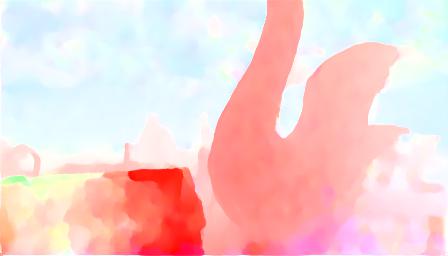}}\newline
	{\includegraphics[width=0.242\linewidth]{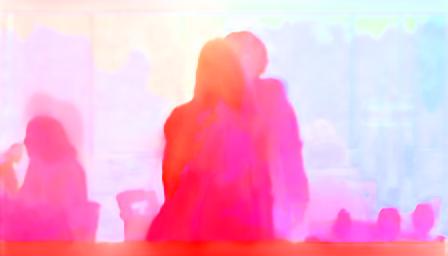}}\hspace{-1pt}
	{\includegraphics[width=0.242\linewidth]{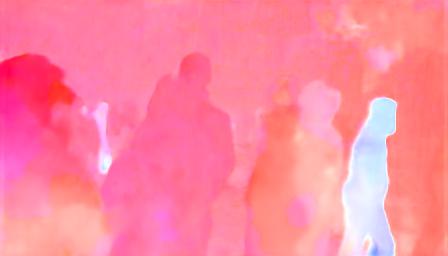}}\hspace{-1pt}
	{\includegraphics[width=0.242\linewidth]{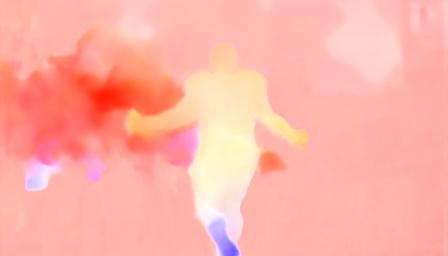}}\hspace{-1pt}
	{\includegraphics[width=0.242\linewidth]{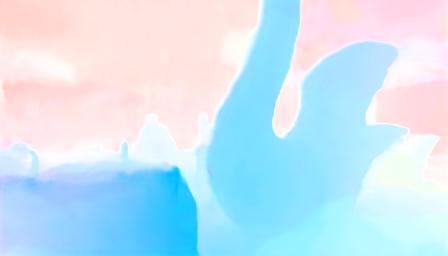}}\newline
	{\includegraphics[width=0.242\linewidth]{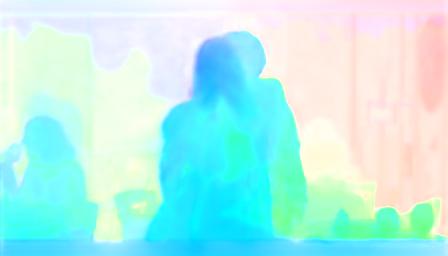}}\hspace{-1pt}
	{\includegraphics[width=0.242\linewidth]{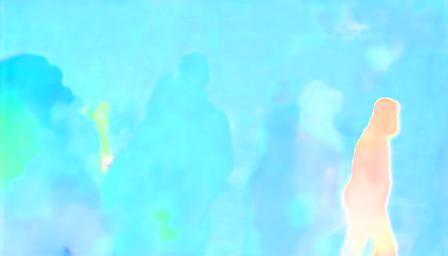}}\hspace{-1pt}
	{\includegraphics[width=0.242\linewidth]{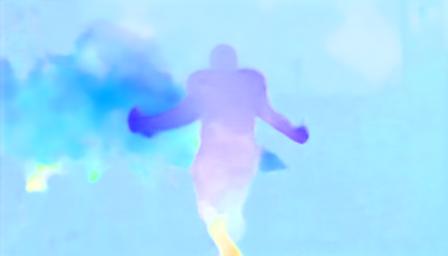}}\hspace{-1pt}
	{\includegraphics[width=0.242\linewidth]{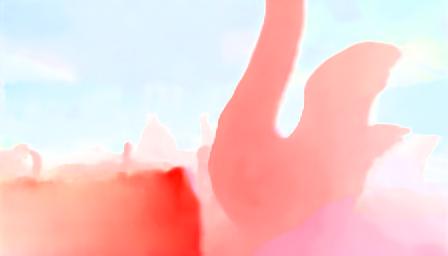}}\newline
	{\includegraphics[width=0.242\linewidth]{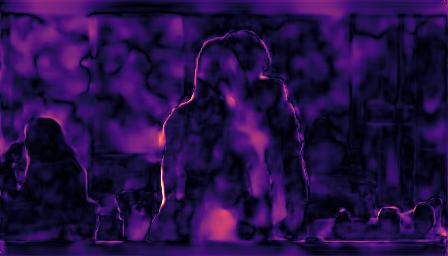}}\hspace{-1pt}
	{\includegraphics[width=0.242\linewidth]{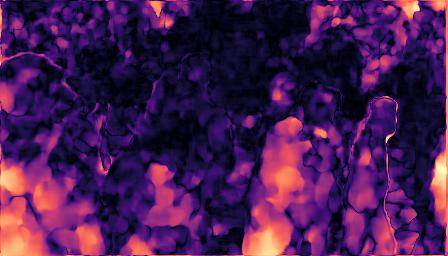}}\hspace{-1pt}
	{\includegraphics[width=0.242\linewidth]{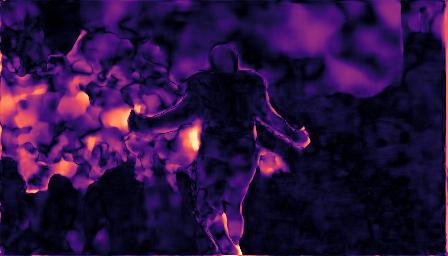}}\hspace{-1pt}
	{\includegraphics[width=0.242\linewidth]{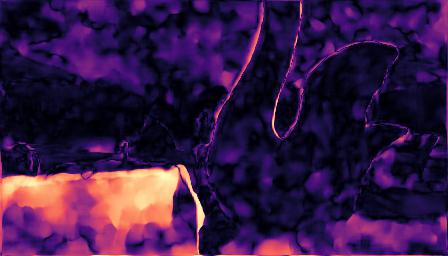}}\newline
	{\includegraphics[width=0.242\linewidth]{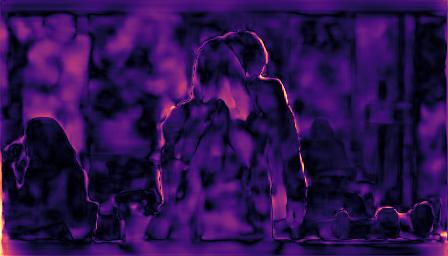}}\hspace{-1pt}
	{\includegraphics[width=0.242\linewidth]{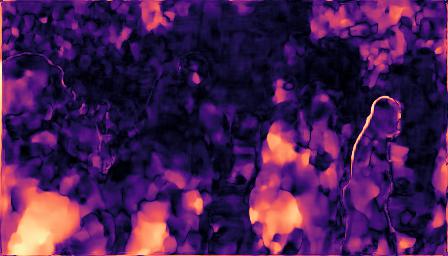}}\hspace{-1pt}
	{\includegraphics[width=0.242\linewidth]{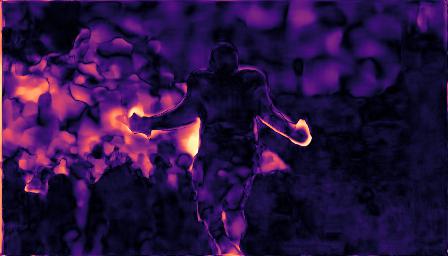}}\hspace{-1pt}
	{\includegraphics[width=0.242\linewidth]{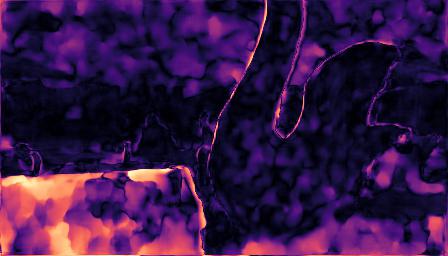}}\newline
	\vspace{-5mm}
	\caption{\textbf{Illustration of task-oriented flow distillation.} From top to bottom rows are ground truth frame $I_{t}^{gt}$, pseudo label of intermediate flow fields $F_{t\rightarrow0}^{p}, F_{t\rightarrow1}^{p}$, predicted intermediate flow fields $F_{t\rightarrow0}, F_{t\rightarrow1}$, task-oriented robustness masks $P_{0}, P_{1}$. Darker color in $P_{0}, P_{1}$ approaches to 1, while brighter color tends to 0. Each column represents a separate example on Vimeo90K~\cite{xue2019video} dataset. Zoom in for best view.}
	\label{fig:17}
	\vspace{-2mm}
\end{figure}
Figure~\ref{fig:17} presents some visual examples to show the robustness masks in proposed task-oriented flow distillation loss, which can decrease the adverse impacts while focusing on the useful knowledge for better frame interpolation. It seems that intermediate flow prediction of IFRNet behaves smoother and contains less artifacts than flow prediction of pseudo label, that helps to achieve better VFI accuracy.

\begin{figure}[h]
	\centering
	{\includegraphics[width=0.242\linewidth]{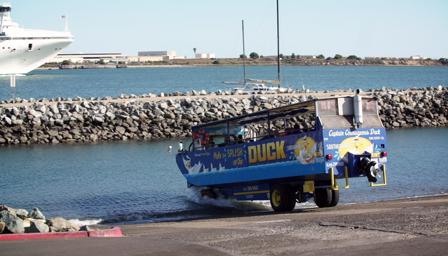}}\hspace{-1pt}
	{\includegraphics[width=0.242\linewidth]{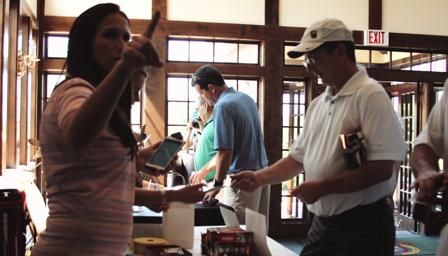}}\hspace{-1pt}
	{\includegraphics[width=0.242\linewidth]{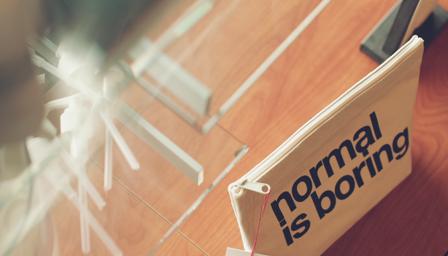}}\hspace{-1pt}
	{\includegraphics[width=0.242\linewidth]{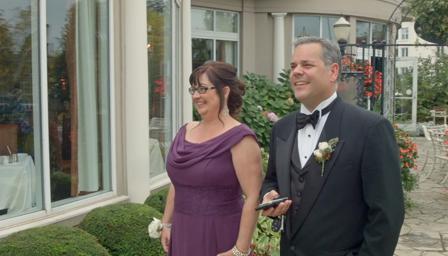}}\newline
	{\includegraphics[width=0.242\linewidth]{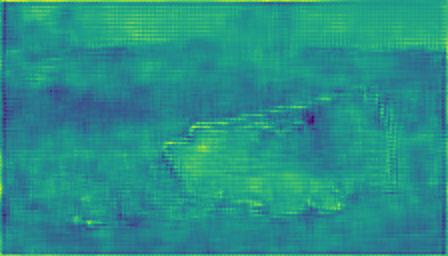}}\hspace{-1pt}
	{\includegraphics[width=0.242\linewidth]{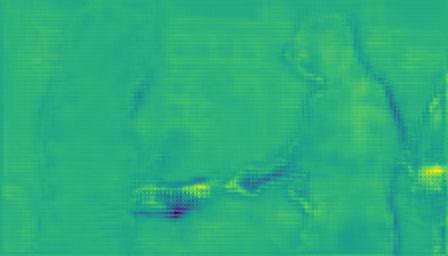}}\hspace{-1pt}
	{\includegraphics[width=0.242\linewidth]{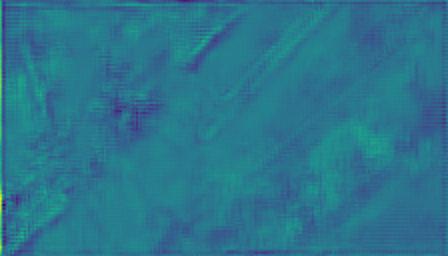}}\hspace{-1pt}
	{\includegraphics[width=0.242\linewidth]{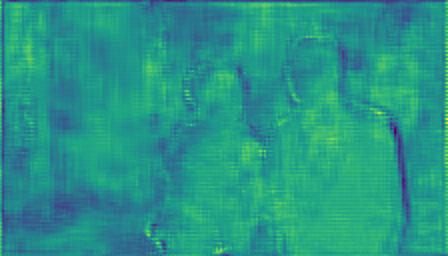}}\newline
	{\includegraphics[width=0.242\linewidth]{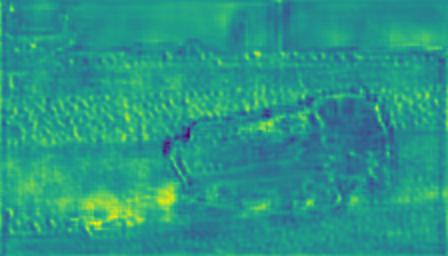}}\hspace{-1pt}
	{\includegraphics[width=0.242\linewidth]{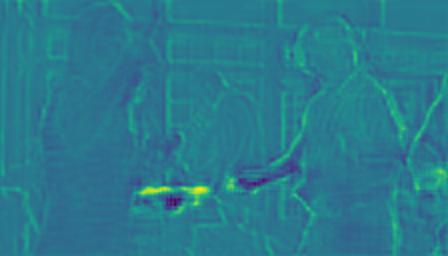}}\hspace{-1pt}
	{\includegraphics[width=0.242\linewidth]{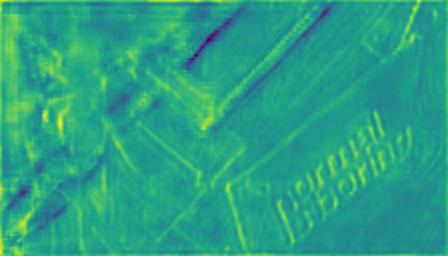}}\hspace{-1pt}
	{\includegraphics[width=0.242\linewidth]{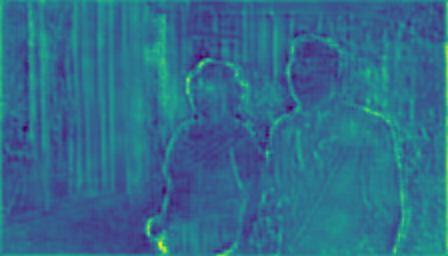}}\newline
	\vspace{-5mm}
	\caption{\textbf{Illustration of mean feature map of intermediate feature $\hat{\phi}_{t}^{1}$ w/o and w/ $\mathcal{L}_{g}$.} From top to bottom rows are ground truth frame $I_{t}^{gt}$, mean feature map of $\hat{\phi}_{t}^{1}$ w/o $\mathcal{L}_{g}$, mean feature map of $\hat{\phi}_{t}^{1}$ w/ $\mathcal{L}_{g}$. Each column represents a separate example on Vimeo90K~\cite{xue2019video} dataset. Zoom in for best view.}
	\label{fig:18}
	\vspace{-2mm}
\end{figure}
Figure~\ref{fig:18} depicts more visual results of mean feature maps of intermediate feature w/o and w/ proposed geometry consistency loss, demonstrating its effect on regularizing refined intermediate feature to keep better structure layout.

\begin{figure*}[t]
	\centering
	{\includegraphics[width=0.122\linewidth]{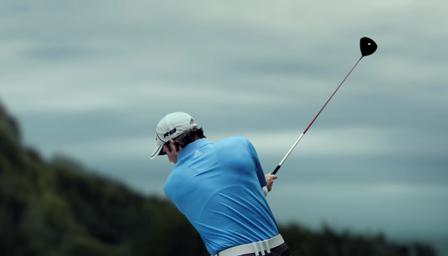}}\hspace{-1pt}
	{\includegraphics[width=0.122\linewidth]{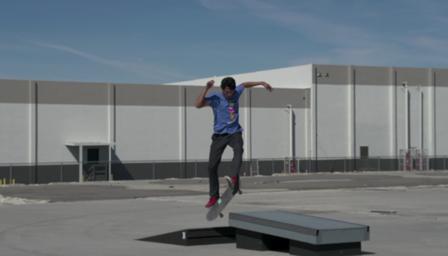}}\hspace{-1pt}
	{\includegraphics[width=0.122\linewidth]{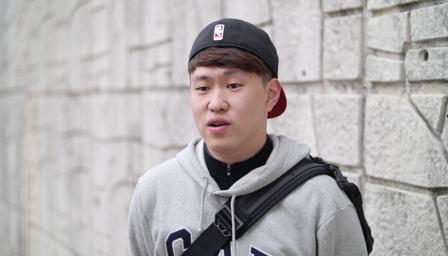}}\hspace{-1pt}
	{\includegraphics[width=0.122\linewidth]{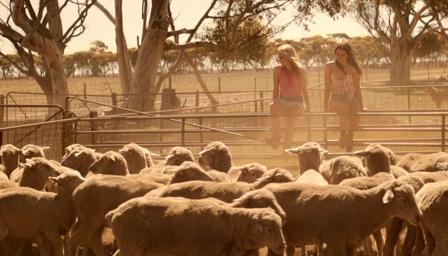}}\hspace{-1pt}
	{\includegraphics[width=0.122\linewidth]{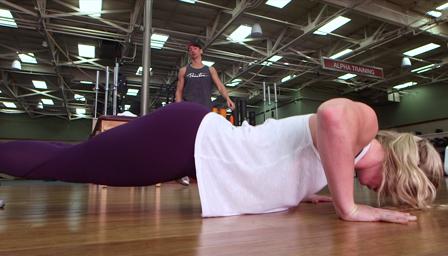}}\hspace{-1pt}
	{\includegraphics[width=0.122\linewidth]{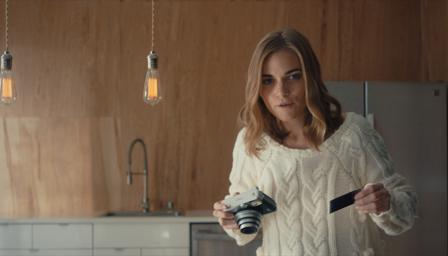}}\hspace{-1pt}
	{\includegraphics[width=0.122\linewidth]{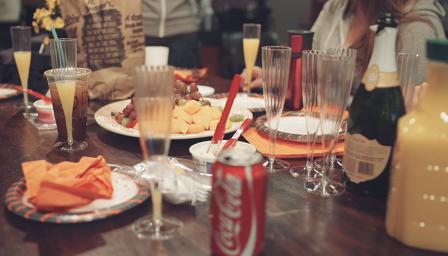}}\hspace{-1pt}
	{\includegraphics[width=0.122\linewidth]{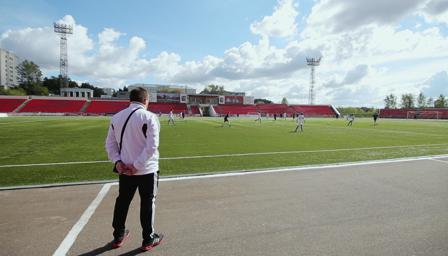}}\newline
	{\includegraphics[width=0.122\linewidth]{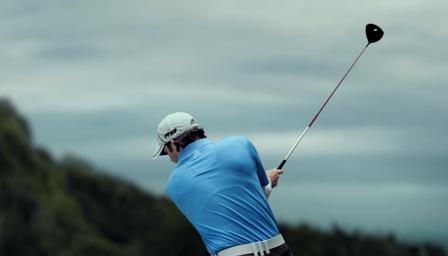}}\hspace{-1pt}
	{\includegraphics[width=0.122\linewidth]{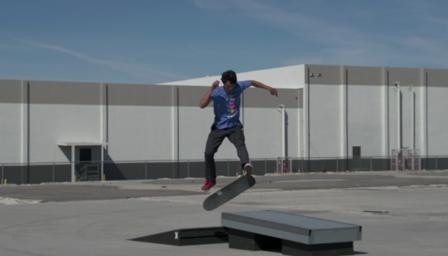}}\hspace{-1pt}
	{\includegraphics[width=0.122\linewidth]{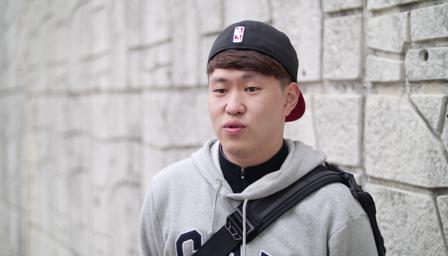}}\hspace{-1pt}
	{\includegraphics[width=0.122\linewidth]{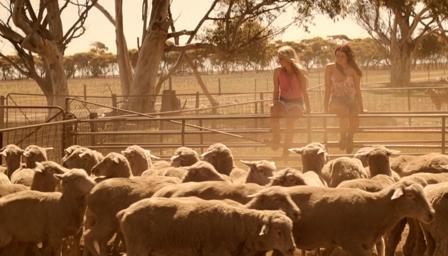}}\hspace{-1pt}
	{\includegraphics[width=0.122\linewidth]{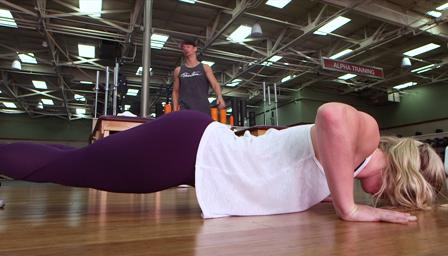}}\hspace{-1pt}
	{\includegraphics[width=0.122\linewidth]{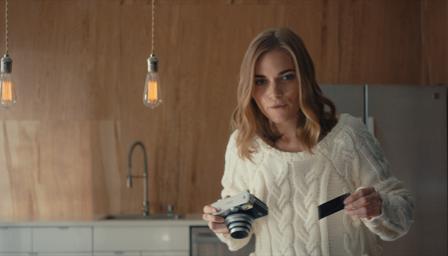}}\hspace{-1pt}
	{\includegraphics[width=0.122\linewidth]{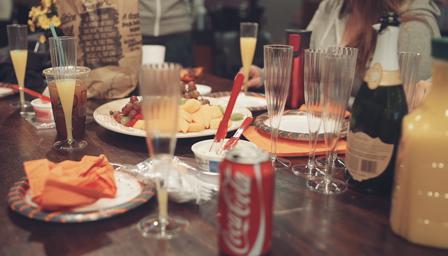}}\hspace{-1pt}
	{\includegraphics[width=0.122\linewidth]{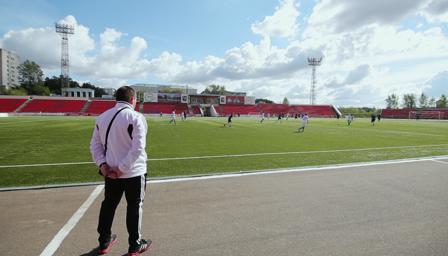}}\newline
	{\includegraphics[width=0.122\linewidth]{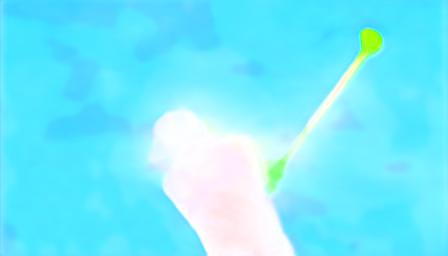}}\hspace{-1pt}
	{\includegraphics[width=0.122\linewidth]{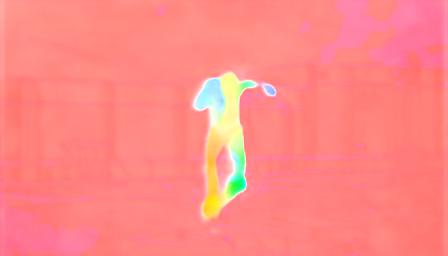}}\hspace{-1pt}
	{\includegraphics[width=0.122\linewidth]{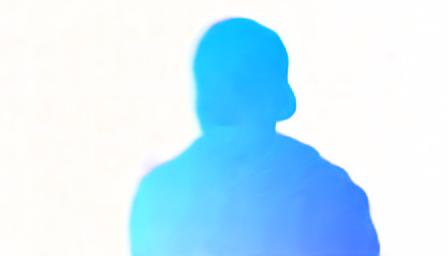}}\hspace{-1pt}
	{\includegraphics[width=0.122\linewidth]{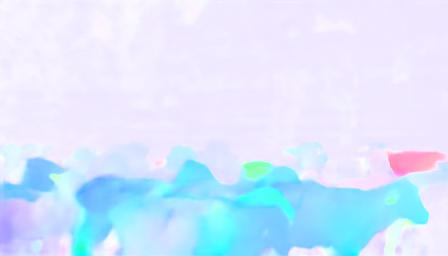}}\hspace{-1pt}
	{\includegraphics[width=0.122\linewidth]{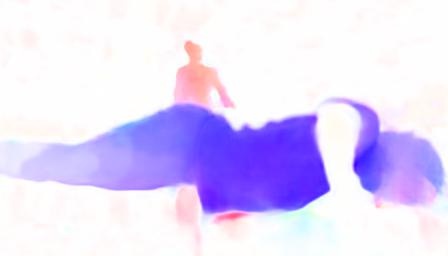}}\hspace{-1pt}
	{\includegraphics[width=0.122\linewidth]{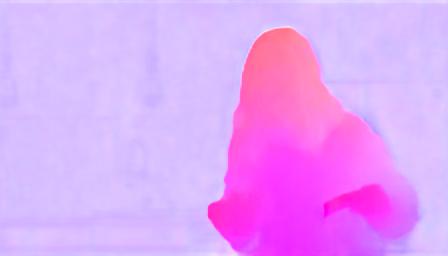}}\hspace{-1pt}
	{\includegraphics[width=0.122\linewidth]{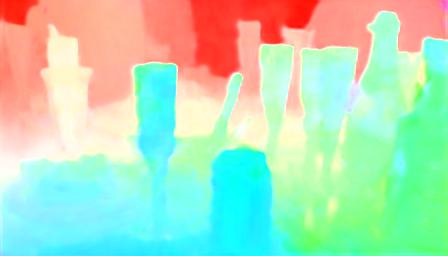}}\hspace{-1pt}
	{\includegraphics[width=0.122\linewidth]{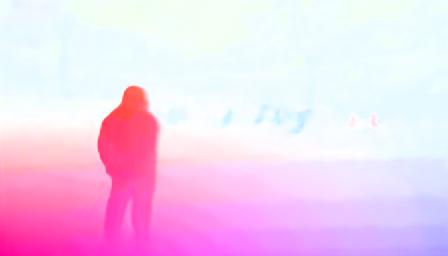}}\newline
	{\includegraphics[width=0.122\linewidth]{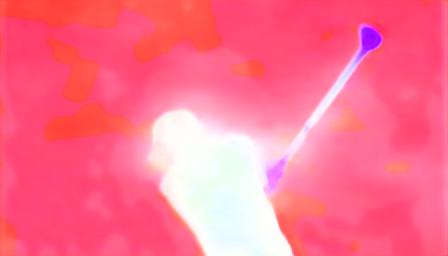}}\hspace{-1pt}
	{\includegraphics[width=0.122\linewidth]{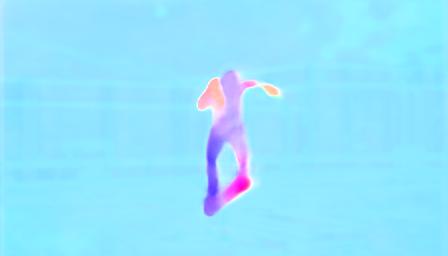}}\hspace{-1pt}
	{\includegraphics[width=0.122\linewidth]{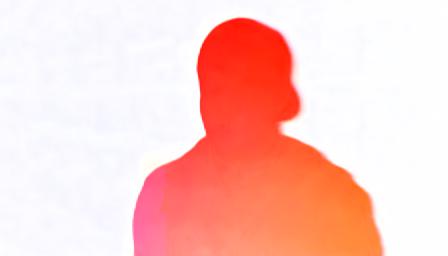}}\hspace{-1pt}
	{\includegraphics[width=0.122\linewidth]{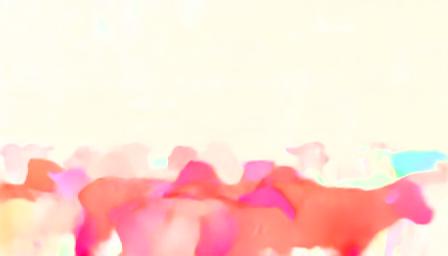}}\hspace{-1pt}
	{\includegraphics[width=0.122\linewidth]{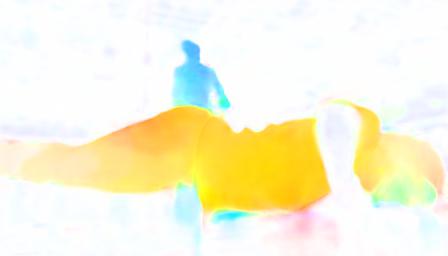}}\hspace{-1pt}
	{\includegraphics[width=0.122\linewidth]{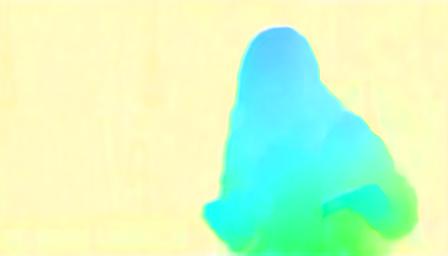}}\hspace{-1pt}
	{\includegraphics[width=0.122\linewidth]{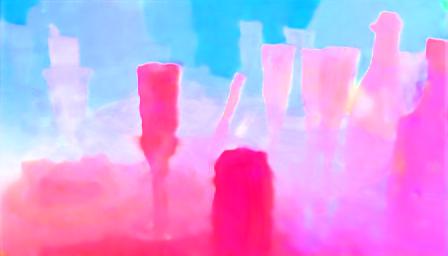}}\hspace{-1pt}
	{\includegraphics[width=0.122\linewidth]{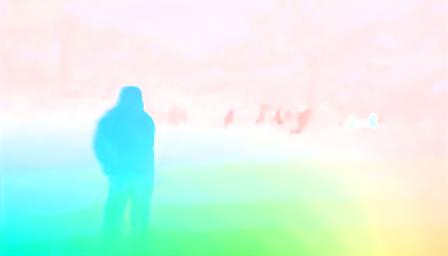}}\newline
	{\includegraphics[width=0.122\linewidth]{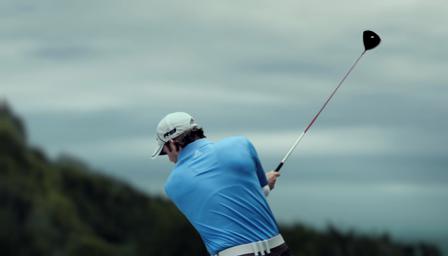}}\hspace{-1pt}
	{\includegraphics[width=0.122\linewidth]{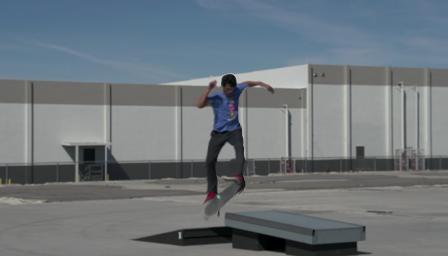}}\hspace{-1pt}
	{\includegraphics[width=0.122\linewidth]{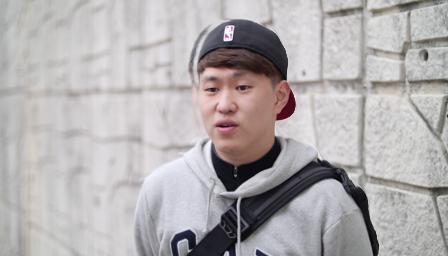}}\hspace{-1pt}
	{\includegraphics[width=0.122\linewidth]{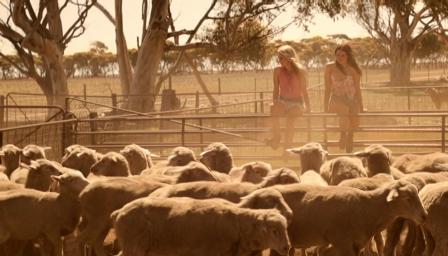}}\hspace{-1pt}
	{\includegraphics[width=0.122\linewidth]{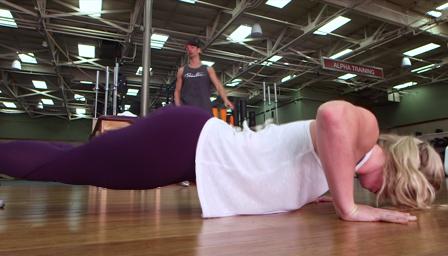}}\hspace{-1pt}
	{\includegraphics[width=0.122\linewidth]{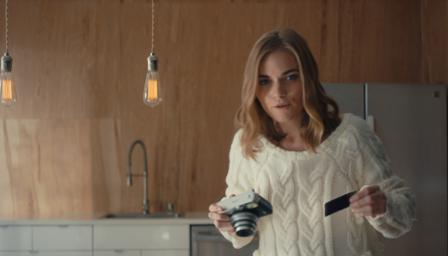}}\hspace{-1pt}
	{\includegraphics[width=0.122\linewidth]{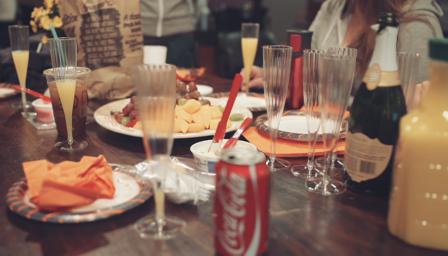}}\hspace{-1pt}
	{\includegraphics[width=0.122\linewidth]{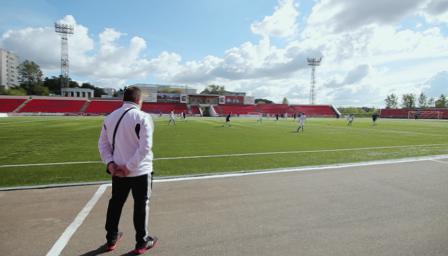}}\newline
	{\includegraphics[width=0.122\linewidth]{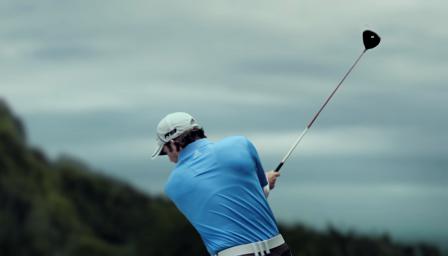}}\hspace{-1pt}
	{\includegraphics[width=0.122\linewidth]{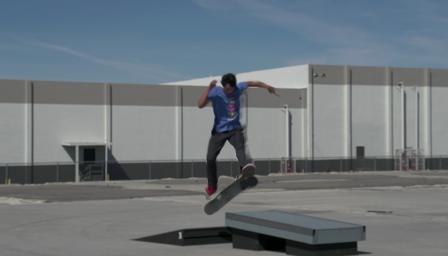}}\hspace{-1pt}
	{\includegraphics[width=0.122\linewidth]{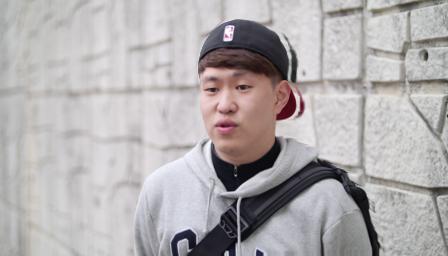}}\hspace{-1pt}
	{\includegraphics[width=0.122\linewidth]{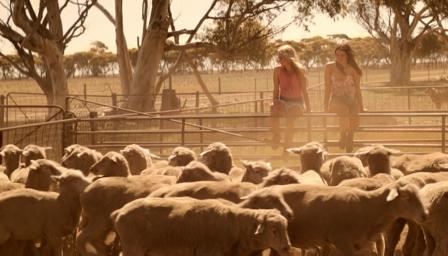}}\hspace{-1pt}
	{\includegraphics[width=0.122\linewidth]{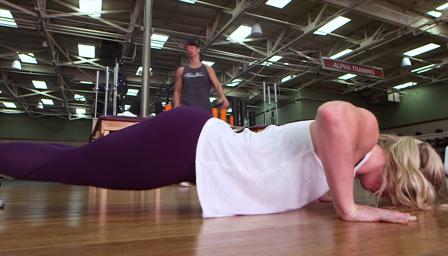}}\hspace{-1pt}
	{\includegraphics[width=0.122\linewidth]{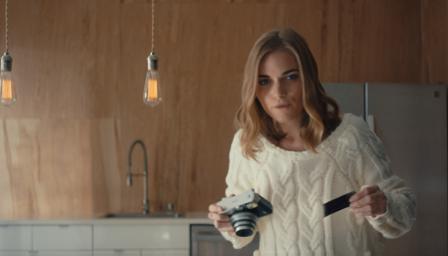}}\hspace{-1pt}
	{\includegraphics[width=0.122\linewidth]{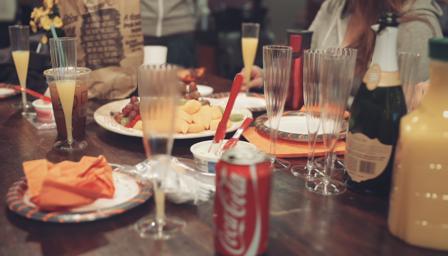}}\hspace{-1pt}
	{\includegraphics[width=0.122\linewidth]{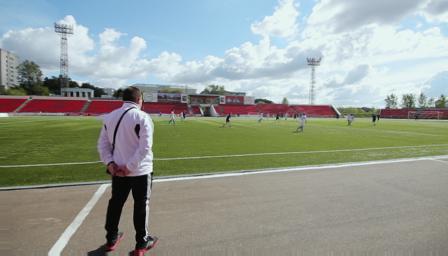}}\newline
	{\includegraphics[width=0.122\linewidth]{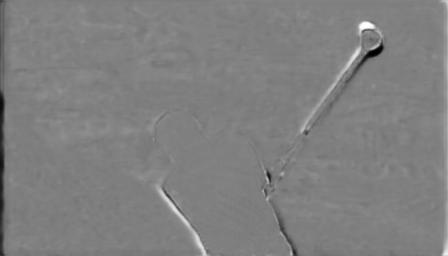}}\hspace{-1pt}
	{\includegraphics[width=0.122\linewidth]{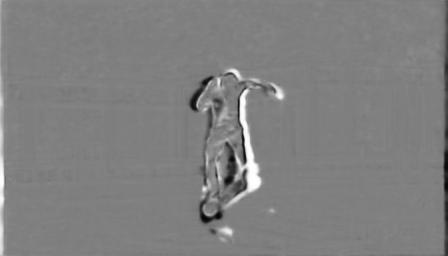}}\hspace{-1pt}
	{\includegraphics[width=0.122\linewidth]{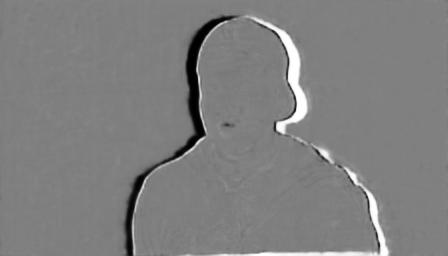}}\hspace{-1pt}
	{\includegraphics[width=0.122\linewidth]{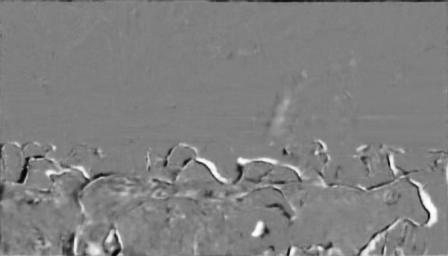}}\hspace{-1pt}
	{\includegraphics[width=0.122\linewidth]{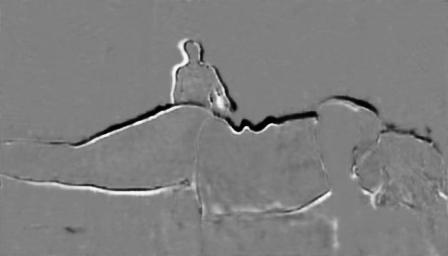}}\hspace{-1pt}
	{\includegraphics[width=0.122\linewidth]{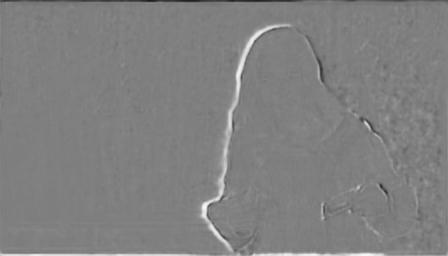}}\hspace{-1pt}
	{\includegraphics[width=0.122\linewidth]{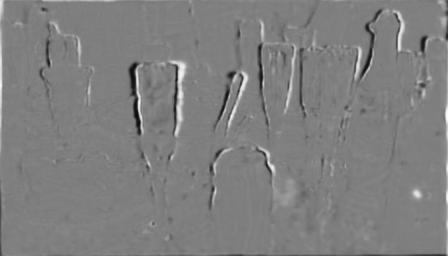}}\hspace{-1pt}
	{\includegraphics[width=0.122\linewidth]{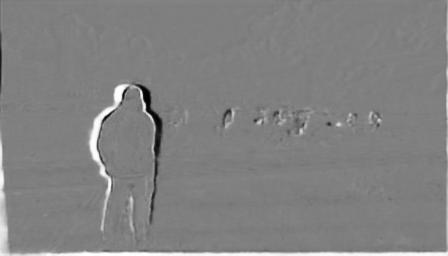}}\newline
	{\includegraphics[width=0.122\linewidth]{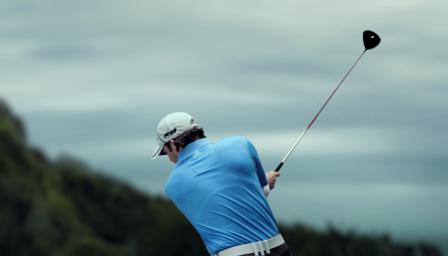}}\hspace{-1pt}
	{\includegraphics[width=0.122\linewidth]{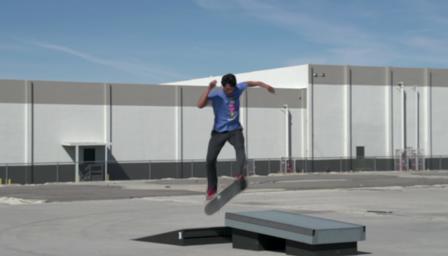}}\hspace{-1pt}
	{\includegraphics[width=0.122\linewidth]{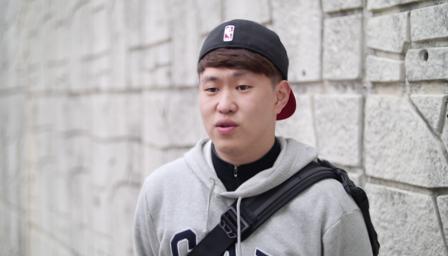}}\hspace{-1pt}
	{\includegraphics[width=0.122\linewidth]{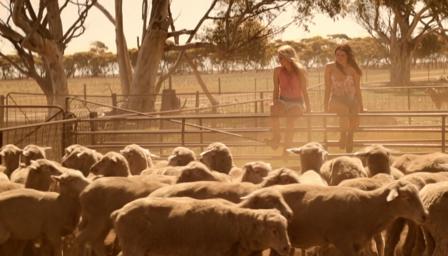}}\hspace{-1pt}
	{\includegraphics[width=0.122\linewidth]{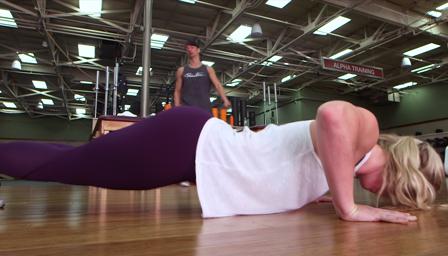}}\hspace{-1pt}
	{\includegraphics[width=0.122\linewidth]{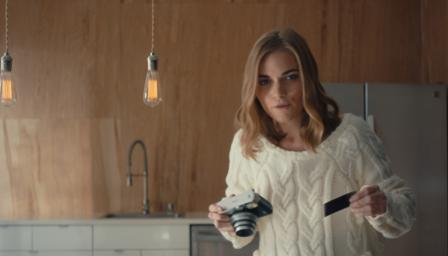}}\hspace{-1pt}
	{\includegraphics[width=0.122\linewidth]{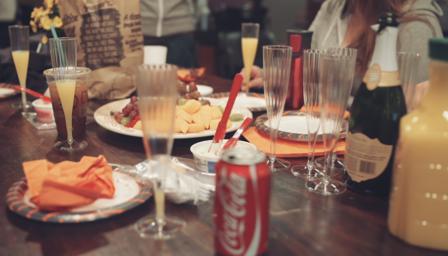}}\hspace{-1pt}
	{\includegraphics[width=0.122\linewidth]{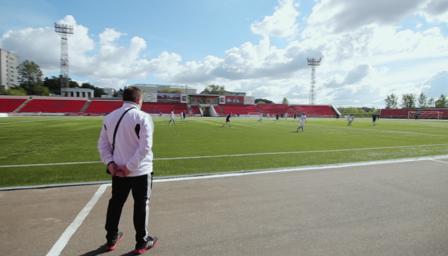}}\newline
	{\includegraphics[width=0.122\linewidth]{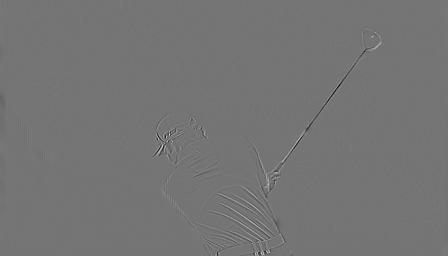}}\hspace{-1pt}
	{\includegraphics[width=0.122\linewidth]{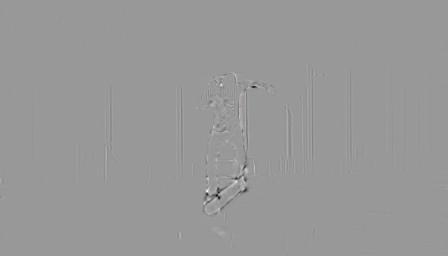}}\hspace{-1pt}
	{\includegraphics[width=0.122\linewidth]{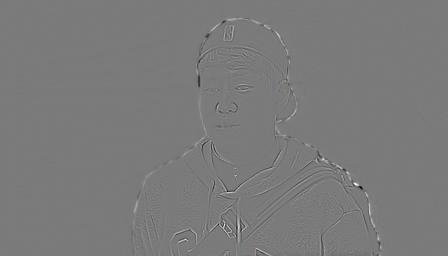}}\hspace{-1pt}
	{\includegraphics[width=0.122\linewidth]{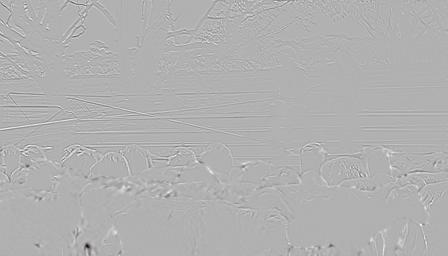}}\hspace{-1pt}
	{\includegraphics[width=0.122\linewidth]{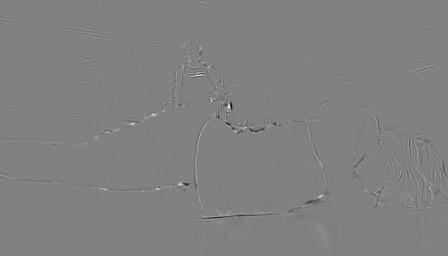}}\hspace{-1pt}
	{\includegraphics[width=0.122\linewidth]{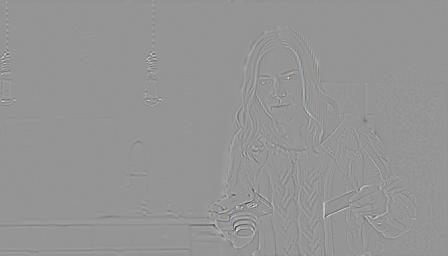}}\hspace{-1pt}
	{\includegraphics[width=0.122\linewidth]{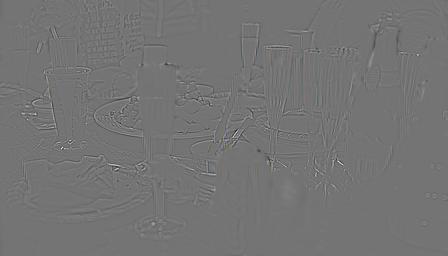}}\hspace{-1pt}
	{\includegraphics[width=0.122\linewidth]{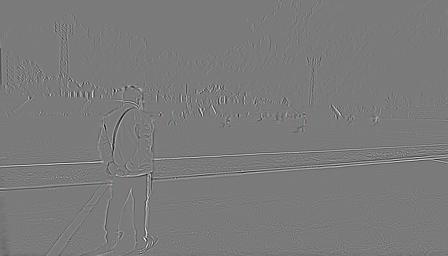}}\newline
	{\includegraphics[width=0.122\linewidth]{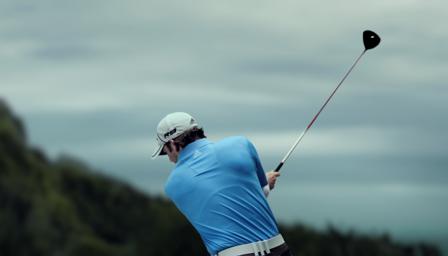}}\hspace{-1pt}
	{\includegraphics[width=0.122\linewidth]{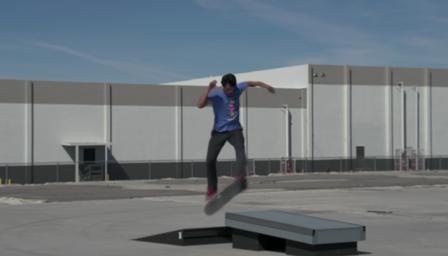}}\hspace{-1pt}
	{\includegraphics[width=0.122\linewidth]{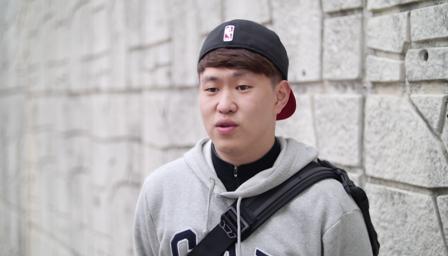}}\hspace{-1pt}
	{\includegraphics[width=0.122\linewidth]{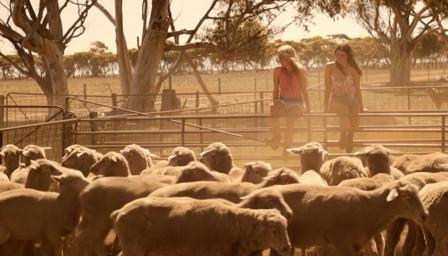}}\hspace{-1pt}
	{\includegraphics[width=0.122\linewidth]{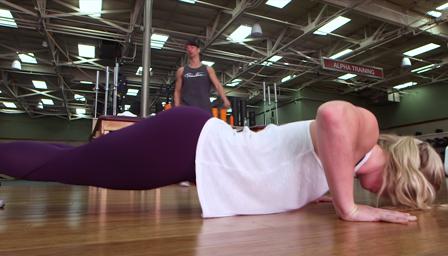}}\hspace{-1pt}
	{\includegraphics[width=0.122\linewidth]{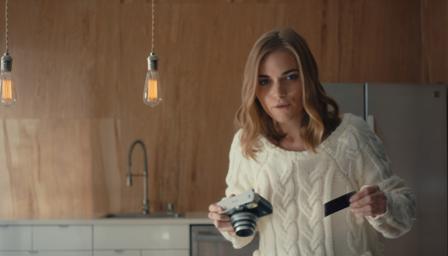}}\hspace{-1pt}
	{\includegraphics[width=0.122\linewidth]{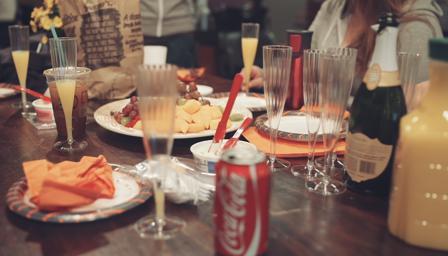}}\hspace{-1pt}
	{\includegraphics[width=0.122\linewidth]{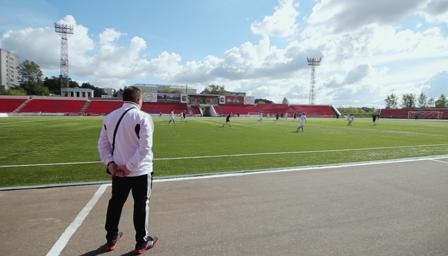}}\newline
	{\includegraphics[width=0.122\linewidth]{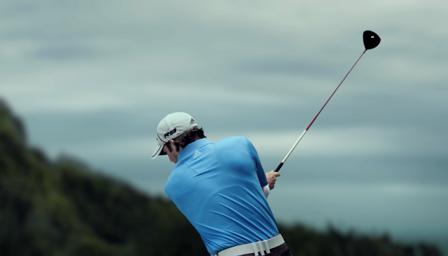}}\hspace{-1pt}
	{\includegraphics[width=0.122\linewidth]{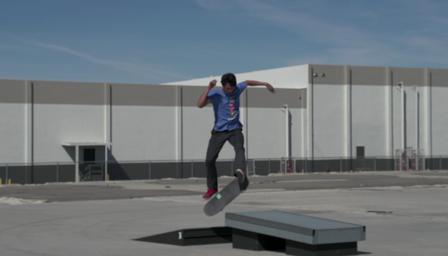}}\hspace{-1pt}
	{\includegraphics[width=0.122\linewidth]{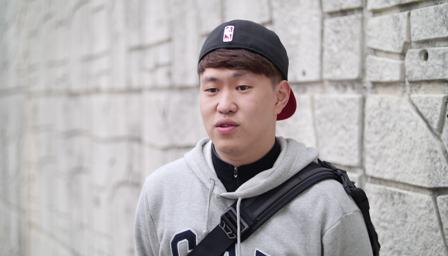}}\hspace{-1pt}
	{\includegraphics[width=0.122\linewidth]{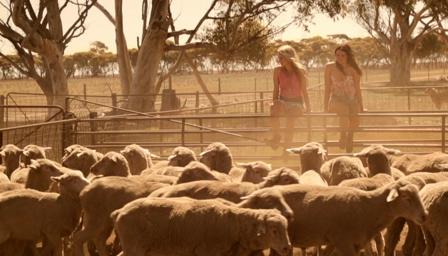}}\hspace{-1pt}
	{\includegraphics[width=0.122\linewidth]{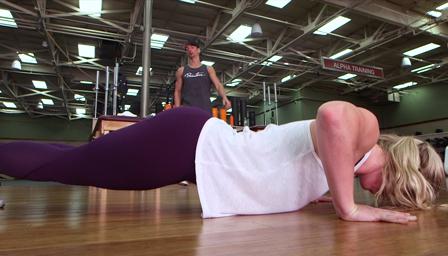}}\hspace{-1pt}
	{\includegraphics[width=0.122\linewidth]{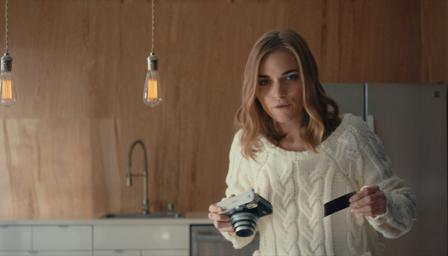}}\hspace{-1pt}
	{\includegraphics[width=0.122\linewidth]{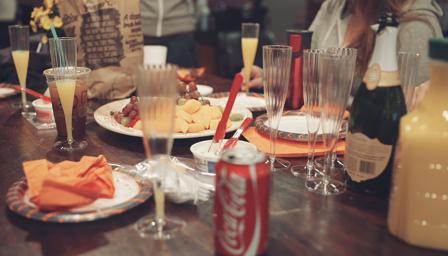}}\hspace{-1pt}
	{\includegraphics[width=0.122\linewidth]{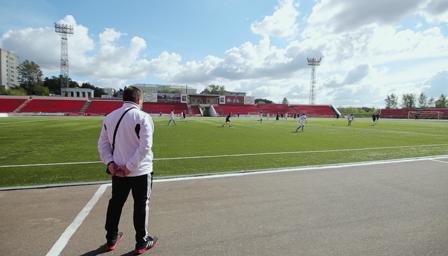}}\newline
	\vspace{-5mm}
	\caption{\textbf{Illustration of intermediate components of IFRNet.} From top to bottom rows are input frames $I_{0}, I_{1}$, predicted intermediate flow fields $F_{t\rightarrow0}, F_{t\rightarrow1}$, warped input frames $\tilde{I}_{0}, \tilde{I}_{1}$, merge mask $M$, merged frame $\hat{I}_{t}^{'}$, residual $R$, final prediction $\hat{I}_{t}$ and ground truth $I_{t}^{gt}$, where merged frame is calculated by $\hat{I}_{t}^{'} = M \odot \tilde{I}_{0} + (1 - M) \odot \tilde{I}_{1}$. For better visualization of residual $R$, we multiply it by 10 and add a bias of 0.5. Each column represents a separate example on Vimeo90K~\cite{xue2019video} dataset. Zoom in for best view.}
	\label{fig:19}
	\vspace{-3mm}
\end{figure*}
Figure~\ref{fig:19} gives visual understanding of frame interpolation process of IFRNet. Thanks to the reference anchor information offered by intermediate feature together with effective supervision provided by geometry consistency loss and task-oriented flow distillation loss, IFRNet can estimate relatively good intermediate flow with clear motion boundary. Further, we can see that merge mask $M$ can identify occluded regions of warped frames by adjusting the mixing weight, where it tends to average the candidate regions when both views are visible. Finally, residual $R$ can compensate for some contextual details, which usually response at motion boundary and image edges. Different from other flow-based VFI methods that take cascaded structure design, merge mask $M$ and residual $R$ in IFRNet share the same encoder-decoder with intermediate optical flow, making proposed architecture achieve better VFI accuracy while being more lightweight and fast.

\begin{figure*}[t]
	\centering
	\includegraphics[width=0.95\linewidth]{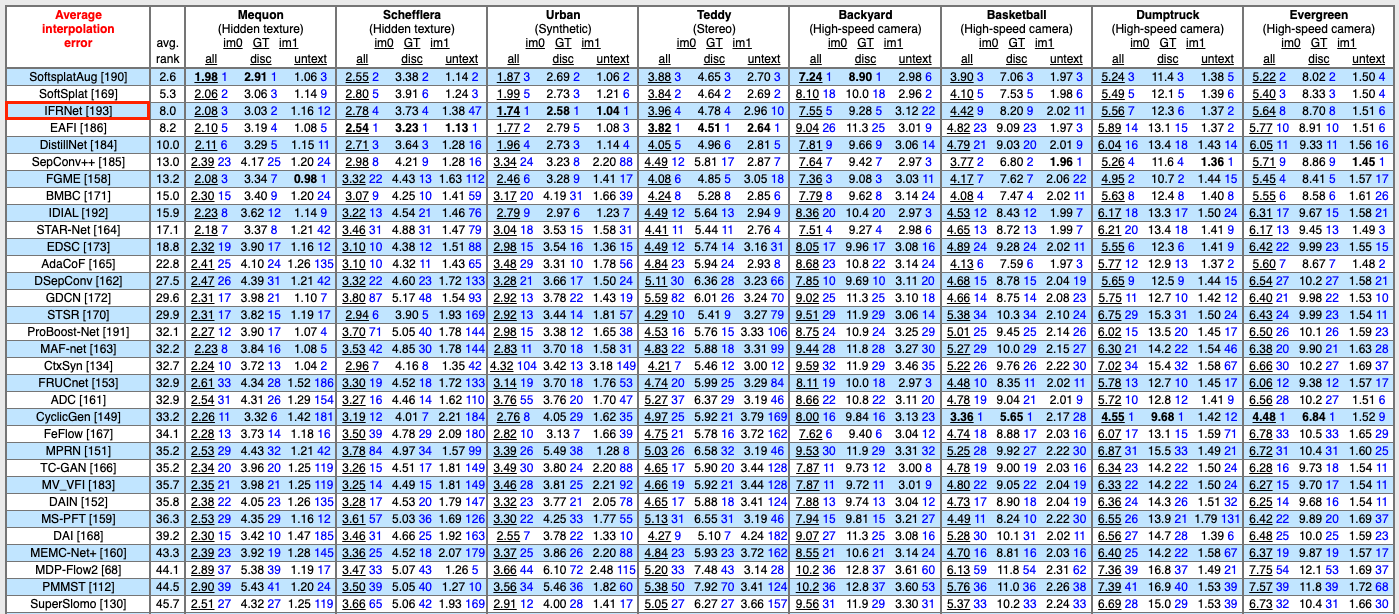}
	\vspace{-2mm}
	\caption{\textbf{Screenshot of our IE-ranking on the Middlebury benchmark (taken on the November 16th, 2021).}}
	\label{fig:20}
	\vspace{-2mm}
\end{figure*}

\begin{figure*}[t]
	\centering
	\includegraphics[width=0.95\linewidth]{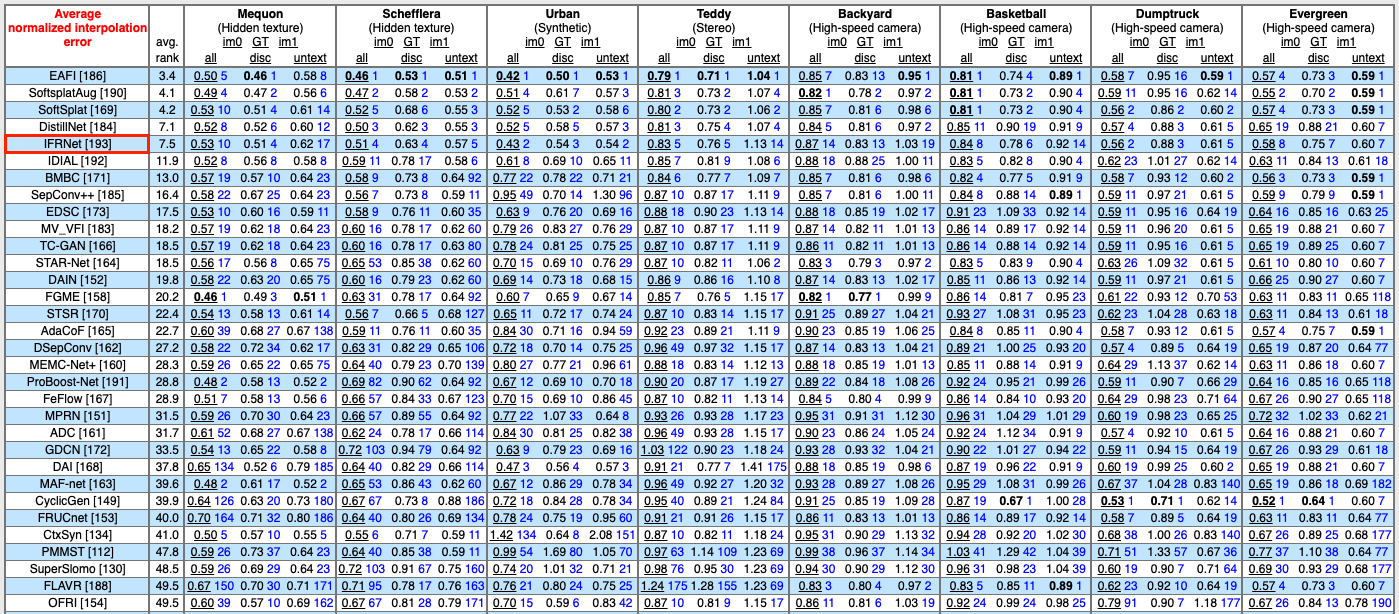}
	\vspace{-2mm}
	\caption{\textbf{Screenshot of our NIE-ranking on the Middlebury benchmark (taken on the November 16th, 2021).}}
	\label{fig:21}
	\vspace{-2mm}
\end{figure*}

Readers may think our IFRNet is similar with PWC-Net~\cite{8579029} which is designed for optical flow. However, It is non-trivial to adapt PWC-Net for frame interpolation, since previous related works employ it as one of many components. We summarize their difference in several aspects: \textbf{1)} Anchor feature in PWC-Net is extracted by the encoder, while in IFRNet, it is reconstructed by the decoder. \textbf{2)} Besides motion information in intermediate feature, there are occlusion, texture and temporal information in it. \textbf{3)} PWC-Net designed for motion estimation, is optimized only by flow regression loss with strong augmentation. However, IFRNet designed for frame synthesizing, is optimized in a multi-target manner with weak data augmentation.

\section{Screenshots of the Middlebury Benchmark}
We take screenshots of the online Middlebury benchmark for VFI on the November 16th, 2021, whose results are shown in Figure~\ref{fig:20} and Figure~\ref{fig:21}. Since the average rank is a relative indicator, previous methods~\cite{8954114,Niklaus_2020_CVPR,Gui_2020_CVPR,BMBC} usually report average IE (interpolation error) and average NIE (normalized interpolation error) for comparison. As summarized in Table 2 in our main paper, proposed IFRNet large model achieves best results on both IE and NIE metrics among all published VFI methods that are trained on Vimeo90K~\cite{xue2019video} dataset. Moreover, IFRNet large runs several times faster than previous state-of-the-art algorithms~\cite{Niklaus_2020_CVPR,park2021asymmetric}, demonstrating the superior VFI accuracy and fast inference speed of proposed approaches.

{\small
	\bibliographystyle{ieee_fullname}
	\bibliography{egbib}
}

\end{document}